\documentclass{article}

\usepackage{PRIMEarxiv}

\usepackage[utf8]{inputenc} % allow utf-8 input
\usepackage[T1]{fontenc}    % use 8-bit T1 fonts
\usepackage{hyperref}       % hyperlinks
\usepackage{url}            % simple URL typesetting
\usepackage{booktabs}       % professional-quality tables
\usepackage{amsfonts}       % blackboard math symbols
\usepackage{nicefrac}       % compact symbols for 1/2, etc.
\usepackage{microtype}      % microtypography
\usepackage{lipsum}
\usepackage{fancyhdr}       % header
\usepackage{graphicx}       % graphics
\graphicspath{{media/}}     % organize your images and other figures under media/ folder

\usepackage{bm}
\usepackage{color}
\usepackage{subcaption}
\usepackage{bm}
\usepackage{amsmath}
\usepackage{amssymb}
\usepackage{algorithm}
\usepackage{algorithmic}
\usepackage{mathrsfs}
\usepackage{comment}
\usepackage{multirow}
\usepackage{lscape}
\newcommand{\argmax}{\mathop{\rm arg~max}\limits}
\newcommand{\argmin}{\mathop{\rm arg~min}\limits}
\newcommand{\indep}{\perp \!\!\! \perp}
\title{Causal rule ensemble approach for multi-arm data}

\author{
    Ke Wan \\
	Department of Medical Data Science\\
	Wakayama Medical University\\
    \texttt{wane19911017@gmail.com} \\
    \And
	Kensuke Tanioka \\
	Department of Biomedical Sciences and Informatics\\
	Doshisha University\\
	\texttt{ktanioka@mail.doshisha.ac.jp} \\
	\And
	Toshio Shimokawa \\
	Department of Biostatistics\\
	Wakayama Medical University\\
	\texttt{toshibow2000@gmail.com} \\
}

\begin{document}
\maketitle

\begin{abstract}
Heterogeneous treatment effect (HTE) estimation is critical in medical research. It provides insights into how treatment effects vary among individuals, which can provide statistical evidence for precision medicine. While most existing methods focus on binary treatment situations, real-world applications often involve multiple interventions. However, current HTE estimation methods are primarily designed for binary comparisons and often rely on black-box models, which limit their applicability and interpretability in multi-arm settings. To address these challenges, we propose an interpretable machine learning framework for HTE estimation in multi-arm trials. Our method employs a rule-based ensemble approach consisting of rule generation, rule ensemble, and HTE estimation, ensuring both predictive accuracy and interpretability. Through extensive simulation studies and real data applications, the performance of our method was evaluated against state-of-the-art multi-arm HTE estimation approaches. The results indicate that our approach achieved lower bias and higher estimation accuracy compared with those of existing methods. Furthermore, the interpretability of our framework allows clearer insights into how covariates influence treatment effects, facilitating clinical decision making. By bridging the gap between accuracy and interpretability, our study contributes a valuable tool for multi-arm HTE estimation, supporting precision medicine.
\end{abstract}

% keywords can be removed

\section{Introduction}\label{sec1}

Treatment effect estimation is a critical aspect of clinical studies because it quantifies the causal effect of interventions and provides statistical evidence to guide clinical decision-making and support evidence-based practice. Many existing clinical studies focus on a single intervention, where subjects are assigned to either a control or treatment group based on their exposure to the intervention of interest, known as a two-arm trial. However, some clinical studies involve multiple interventions, where two or more treatments are compared against a common control group. This design, known as a multi-arm trial, offers several advantages over traditional two-arm trials. For instance, in the field of oncology, where a considerable number of cancer therapeutics are in development\cite{John2012}, evaluating multiple new treatments within a single trial framework increases efficiency by maximizing the information gained from a limited number of participants\cite{Wason2014}. Furthermore, multi-arm trials enhance the efficiency of treatment evaluation by reducing the number of separate trials required, thereby reducing costs, shortening trial duration, and facilitating the simultaneous evaluation of multiple interventions\cite{Parmar2014}. Additionally, as in traditional two-arm settings, treatment effects may vary across individuals owing to differences in patient background. Understanding such heterogeneity is also important in multi-arm studies, as different interventions may show different patterns of effect modification across subpopulations. In this study, we focus on the estimation of heterogeneous treatment effects (HTEs) in multi-arm clinical study; additionally, we consider the application in an observational study.

Existing HTE estimation methods generally focus on two-arm settings. For randomized control trial (RCT) data, Foster et al.\cite{Foster2011} proposed the Virtual Twins. Hill\cite{Hill2011} introduced the use of Bayesian additive regression trees (BART) for HTE estimation, whereas Athey et al.\cite{Athey2016} proposed the causal tree method and generalized it to causal forests in Wager et al.\cite{Wager2018}. For observational study data, Athey and Wager\cite{athey2019} generalized the causal forest to observational study settings based on generalized random forests. Powers et al.\cite{Powers2018} focused on observational studies and proposed pollination random forest, causal boost, and causal mars. Kunzel et al.\cite{kunzel2019} categorized existing HTE estimation methods into several distinct frameworks, including S-, T-, and M-learners. Furthermore, they proposed an innovative framework, referred to as the X-learner. Chernozhukov et al.\cite{Chernozhukov2018} proposed the DR-learner, whereas Nie and Wager\cite{Nie2020} proposed the R-learner for HTE estimation. Although these methods effectively estimate HTE in two-arm data, most are not generalized to multi-arm settings. 

Despite the development of numerous HTE estimation methods, approaches specifically designed for multi-arm HTE estimation remain limited. Linden et al.\cite{Linden2015} compared methods for estimating causal effects in observational studies with multi-arm treatments and highlighted the advantages of doubly robust approaches in reducing bias. Hu et al.\cite{Hu2020} evaluated and extended the use of BART for multi-arm causal effect estimation. However, both studies focused on average treatment effects without considering treatment effect heterogeneity. Acharki et al.\cite{Acharki2023} generalized meta-learners to multi-arm settings and comprehensively compared meta-learners for HTE estimation in multi-arm scenarios. Although these methods have been extended to multi-arm settings, their reliance on black-box models limits their interpretability.

Interpretability is particularly critical in medical applications, where understanding how covariates influence treatment effects is essential for making informed decisions. The lack of interpretability can hinder understanding and trust in the results, particularly among medical researchers and professionals\cite{Price2018, Petch2022}. A common approach to interpreting HTE estimates from black-box models involves the application of an interpretable metric\cite{Lundberg2017} or an interpretable model\cite{Spanbauer2021, Falco2023} to the estimated HTE. However, these post-hoc interpretability techniques, which attempt to explain model outputs after estimation, often provide only incomplete or approximate explanations and fail to clarify how treatment effects are fundamentally estimated\cite{Murdoch2019}. In such situations, it is more natural to create an interpretable model rather than using post-hoc interpretation for a black-boxed model\cite{Rudin2019}. Consequently, interpretable HTE estimation methods specifically tailored for multi-arm settings, which balance predictive accuracy with model transparency, are urgently needed. Although several studies have explored these issues, most focus on binary treatment settings with limited discussion on their application to multi-arm scenarios\cite{Wan2023, Hiraishi2024, Wu2025}. Unlike the two-group scenario, the multi-group scenario should consider pairs of comparisons of treatment effects between different treatment groups. Therefore, it is important to ensure the interpretability of differences in treatment effects between any pair of treatment groups. However, directly applying the existing interpretable HTE estimation method for two-arm scenarios is difficult to ensure such interpretation. 

In this study, we construct a novel framework for developing an interpretable rule ensemble HTE estimation model for multi-arm data, consisting of three-steps including rule generation, rule ensemble, and HTE estimation. First, we use a tree-ensemble model to obtain a set of rules associated with HTE. Second, we allow the treatment and control groups to share the same rules and estimate the coefficients of each rule using the group-wise regularization method. This allows us to estimate the difference in the contribution of these rules to the outcomes between several treatments and one control group, which can also be interpreted as the HTE for these rules. Finally, the HTE can be estimated as the sum of these rules and their corresponding coefficients. Therefore, our framework develops an additive model for HTE and uses rules as base functions. In the proposed approach, we constructed a model that shares rules and linear terms among groups, so that the HTE or the treatment effect differences between any two treatments can be easily interpreted by these rules. Therefore, the proposed framework, which also includes the method proposed by Wan et al.\cite{Wan2023} but is more generalizable, is flexible for building interpretable HTE estimation models without being limited by the number of treatment groups and allows various rule generation models and ensemble in HTE estimation models to be explored.

The remainder of this paper is organized as follows: Section 2 introduces the relative approach; Section 3 describes the proposed method in detail; Section 4 presents several simulation studies comparing the prediction performance of the proposed method with previous HTE estimation methods; Section 5 describes the application of the proposed method to real data and explains how it works; and Section 6 summarizes the study and discusses its results.

\section{Related works}\label{sec2}

Here, we first introduce the definition of HTE for multi-arm trial and then introduce some previous methods.

\subsection{HTE for multi-arm data}
The HTE for multi-arm data is defined within the potential outcome framework. While this framework was originally developed for binary treatment settings\cite{Rubin1974}, it has since been extended to accommodate multiple treatment arms\cite{Imbens2000, Michael2001, Imai2004}. In this study, we focus on the situation of simultaneously comparing multiple treatment groups with a common control group. Therefore, we consider the $N$ samples dataset $\{(Y_i, W_i, \bm{X}_i)\}_{i=1}^N$ in which $W_i\in\{0, 1, \cdots, T\}$ with $T \geq 2$ is the treatment indicator for subject $i$, where $W_i = 0$ if assigned to the control group and $W_i = t \in \{1, \cdots, T\}$ if assigned to the $t$-th treatment group; $\bm{X}_i = (X_{i1},\cdots,X_{ip})^\mathrm{T}\in \mathbb{R}^p$ is the covariates vector with $p$ variables for subject $i$ where the "${}^\mathrm{T}$" on vector means the transpose; and $Y_i$ is the outcome for subject $i$. Let $\{Y_i^{(0)}, Y_i^{(1)}\cdots, Y_i^{(T)}\}$ be the set of potential outcomes for subject $i$, where $Y_i^{(0)}$ and $Y_i^{(t)}, t \in \{1, \cdots, T\}$ are the potential outcomes in the control and $t$-th treatment groups. The HTE of the $t$-th treatment for multi-arm data is defined as
\begin{align}
\Delta^{(t)}(\bm{x}_i) = \mathbb{E}\left(Y_i^{(t)} - Y_i^{(0)} \ \Big|\ \bm{X}_i = \bm{x}_i\right). \label{HTE_def}
\end{align}
However, the potential outcomes for the control and treatment groups for each subject, $i$, cannot be observed simultaneously, as each subject is assigned to only one treatment arm. Consequently, the HTE cannot be directly inferred. To ensure that the HTE for multi-arm data is identifiable, the following three assumptions are made:
\begin{description}
\item{Assumption 1:} (Unconfoundedness)\quad Potential outcomes $\{Y_i^{(0)}, Y_i^{(1)}\cdots, Y_i^{(T)}\}$ are independent of treatment assignment $W_i$, given observed covariates $\bm{X}_i$.
\begin{align*}
\{Y_i^{(0)}, Y_i^{(1)}\cdots, Y_i^{(T)}\}\perp W_i\bigm|\bm{X}_i = \bm{x}_i,\quad \forall W_i\in\{0, 1,\cdots, T\}.
\end{align*}
\item{Assumption 2:} (Overlap)\quad Every subject has a non-zero probability of receiving each treatment level, given their covariates.
\begin{align*}
e(t, \bm{x}_i) = \mathbb{P}\left(W_i = t \bigm| \bm{X}_i = \bm{x}_i\right), \quad \forall t\in\{0, 1,\cdots, T\} \quad \mathrm{satisfies} \quad 0 < e(t, \bm{x}_i) < 1,
\end{align*}
where $e(t, \bm{x}_i)$ is also termed as generalized propensity score (GPS)\cite{Imbens2000}.
\item{Assumption 3:} (Stable Unit Treatment Value Assumption; SUVTA)\quad No interference exists between units, and no other versions or forms of treatment levels are considered.
\end{description}

Under these assumptions, the HTE of the $t$-th treatment for multi-arm data in Eq.(\ref{HTE_def}) can be rewritten as 
\begin{align}
\Delta^{(t)}(\bm{x}_i) &= \mathbb{E}\left(Y_i^{(t)} - Y_i^{(0)} \ \Big|\ \bm{X}_i = \bm{x}_i\right)\notag\\
&= \mathbb{E}\left(Y_i^{(t)} \Big|\ \bm{X}_i = \bm{x}_i\right) - \mathbb{E}\left(Y_i^{(0)} \ \Big|\ \bm{X}_i = \bm{x}_i\right)\notag\\
&= \mathbb{E}\left(Y_i \Big|\ W_i = t, \bm{X}_i = \bm{x}_i\right) - \mathbb{E}\left(Y_i \ \Big|\ W_i = 0, \bm{X}_i = \bm{x}_i\right). \label{HTE_est}
\end{align}

\subsection{Multi-arm HTE estimation approaches}
HTE estimation for multi-arm data is typically based on meta-learners, which provide a flexible framework that enables the use of any regression approaches for estimating treatment effects. The key advantage of meta-learners lies in their ease of application. By allowing any regression model to be incorporated into the estimation process, meta-learners facilitate the seamless extension of advanced and high-accuracy regression techniques to HTE estimation. This flexibility could improve the accuracy of HTE estimation. Here, we present previous studies and divide these meta-learners into conditional mean regression, transform outcome method, and others to demonstrate them in detail. 

\subsubsection{Conditional mean regression} These methods build models for the control and treatment groups. The typical approaches include the T- and S-learner. 

\paragraph{T-learner:} The T-learner builds separate models for each treatment and control group. For each treatment $t\in \{1,\cdots,T\}$ and control group ($t=0$), it estimates the conditional expected outcome for a given covariate vector $\bm{x}_i$. The HTE for the $t$-th treatment group is then estimated by calculating the difference between the estimated outcomes for the treatment and control groups.
\begin{align*}
\Delta^{(t)}(\bm{x}_i) &= \mu^{(t)}(\bm{x}_i) - \mu^{(0)}(\bm{x}_i)
\end{align*}
where $\mu^{(t)}(\bm{x}_i) = \mathbb{E}(Y_i| W_i = t, \bm{X}_i = \bm{x}_i)$ and $\mu^{(0)}(\bm{x}_i) = \mathbb{E}(Y_i| W_i = 0, \bm{X}_i = \bm{x}_i)$ are the estimated outcomes for the $t$-th treatment and control groups, respectively.

\paragraph{S-learner:} The S-learner uses a single model to estimate outcomes for all treatment groups and the control group by including the treatment indicator as a covariate. The HTE for the $t$-th treatment group is then estimated by calculating the difference between the estimated outcomes for the treatment and control groups 
\begin{align*}
\Delta^{(t)}(\bm{x}_i) = \mu(\bm{x}_i, w_i = t) - \mu(\bm{x}_i, w_i = 0)
\end{align*}
where $\mu(\bm{x}_i, w_i = t) = \mathbb{E}(Y_i| W_i = t, \bm{X}_i = \bm{x}_i)$ and $\mu(\bm{x}_i, w_i = 0) = \mathbb{E}(Y_i| W_i = 0, \bm{X}_i = \bm{x}_i)$ are the estimated outcomes for the $t$-th treatment and control group, respectively.

The main advantage of these methods is the ease of extending any regression method to HTE estimation. However, these methods do not consider the bias in treatment assignment. Therefore, they may perform poorly when the treatment assignment is biased. Kunzel et al. (2020)\cite{kunzel2019} proposed X-learners, which extend T-learners and adjust the effect of bias in treatment assignment using propensity scores, demonstrating good performance in HTE estimation. This method was also generalized into multi-arm HTE estimation in Acharki et al. (2023)\cite{Acharki2023}. These approaches build models to estimate the outcome for each group, separately and use the difference between treatment and control groups to estimate the HTE for each treatment group. Therefore, these approaches typically construct models with different base functions across groups, which can hinder the interpretation of the relationships between covariates and the HTE. 

\subsubsection{Transform outcome regression} These methods estimate HTEs by fitting models to transformed outcomes, where the transformed outcome is constructed by weighting the observed outcome using inverse probability weighting. A typical approach is the M-learner. 

\paragraph{M-learner:} The M-learner consists of two steps. First, the transformed outcome for the $t$-th treatment group is calculated as
\begin{align}
Z_i^{(t)} &= \frac{I(W_i = t)Y_i}{e(t,\bm{x}_i)} - \frac{I(W_i = 0)Y_i}{e(0,\bm{x}_i)}.
\label{Trans_outcome}
\end{align}
where $I(\cdot)$ is an indicator function that returns 1 if the statement within parentheses is true and 0 if it is false, $e(t,\bm{x}_i) = \mathbb{P}(W_i = t|\bm{X}_i = \bm{x}_i)$ and $e(0,\bm{x}_i) = P(W_i = 0|\bm{X}_i = \bm{x}_i)$ are the GPS)\cite{Imbens2000} for the $t$-th treatment and control groups, respectively. 

The model is then fitted to the transformed outcomes. The transform outcome satisfies 
\begin{align*}
\mathbb{E}(Z^{(t)}|\bm{X} = \bm{x}) &= \mathbb{E}\left(\frac{I(W = t)Y}{e(t,\bm{x})}\Bigg|\bm{X} =\bm{x}\right) - \mathbb{E}\left(\frac{I(W = 0)Y}{e(0,\bm{x})}\Bigg|\bm{X} =\bm{x}\right)\\
& = \mathbb{P}(W = t|\bm{X} = \bm{x})\frac{\mathbb{E}(Y|W = t, \bm{X} = \bm{x})}{e(t,\bm{x})} - \mathbb{P}(W = 0|\bm{X} = \bm{x})\frac{\mathbb{E}(Y|W = 0, \bm{X} = \bm{x})}{e(0,\bm{x})}\\
& = \mathbb{E}(Y|W = t, \bm{X} = \bm{x}) - \mathbb{E}(Y|W = 0, \bm{X} = \bm{x})\\
& = \Delta^{(t)}(\bm{x}).
\end{align*}
Therefore, the transformed outcome is an unbiased estimator of the HTE, which implies that fitting a model to the transformed outcomes allows for the direct estimation of the HTE. 

Because this method allows us to directly build the HTE estimation model, we can directly obtain an interpretable model for HTE estimation by applying interpretable models such as decision trees and RuleFit. Although this framework brings us closer to our objective, a critical limitation remains. The HTE estimates from the transformed outcome approach are highly sensitive to the estimated GPS, where small GPS values often result in extreme weights. Despite correct GPS estimation, the method can suffer from high variance owing to the presence of extreme weights, as noted by Curth et al. (2021)\cite{Curth2021a}. This limitation may be naturally exacerbated in multi-arm treatment settings, where the GPS for each treatment arm decreases as the number of arms increases, potentially leading to high variability and instability in HTE estimation. The double-robust learner is proposed to address this problem. This approach can be regarded as the extended version of the M-learner, which combines GPS and outcome modeling to achieve robustness in HTE estimation. However, the HTE estimates of this approach mixed the results of outcome modeling, making the result difficult to interpret despite the use of the interpretable model.   

\subsubsection{Other approaches}
In addition to the conditional mean regression framework and the transformed outcome approach, several other methods have been proposed for HTE estimation. Nie and Wager (2020)\cite{Nie2020} introduced the ``R-learner,'' a method designed for binary treatment arms, and discussed its extension to multi-arm treatment HTE estimation. The method is formulated as follows:
\begin{align}
\hat{\Delta}^{(t)}(\bm{x}_i) = \argmin_{\Delta^{(t)}(\bm{x}_i)}\left(\frac{1}{N}\sum_{i = 1}^N\left[\left\{y_i - m(\bm{x}_i)\right\}-\left\langle w_i - \hat{e}(k, \bm{x}_i), \Delta^{(t)}(\bm{x}_i)\right\rangle\right]^2\right)
\end{align}
where $\hat{e}(t, \bm{x}_i)$ denotes the estimated GPS for the $t$-th group, with $k\in \{0,\cdots, T\}$. The function $m:\bm{x} \mapsto y$ represents the estimated outcome given $\bm{x}_i$, and $\langle \cdot, \cdot \rangle$ denotes the inner product. Zhou et al. (2023)\cite{Zhou2023} extended the ``R-learner'' to the ``reference-free R-learner'' for multi-arm treatment estimation, demonstrating strong performance in optimal treatment selection. However, both require the estimates of $m(\bm{x})$ in HTE estimation, which complicates the construction of an interpretable model that explicitly relates covariates to HTEs.

\section{Proposed approaches}\label{sec3}

Meta-learners provide a flexible framework for multi-arm HTE estimation using various regression methods, and with appropriate regression methods, high prediction accuracy can be achieved using these learners. However, most meta-learners use several models in HTE estimation, which limits their interpretability. Using M-learners with an interpretable ensemble model such as RuleFit applied could build an interpretable model for the; however, it often suffers from high variance owing to extreme weighting, which could lead to model instability and poor HTE estimation accuracy. To address these challenges, alternative approaches for multi-arm HTE estimation are needed.

In this study, we focus on a RuleFit-based HTE estimation method proposed by Wan et al. \cite{Wan2023}. This method constructs an interpretable rule-based model for HTE estimation while maintaining strong predictive performance. It can be viewed as a modification of the RuleFit method by integrating the M-learner framework with a specialized version of the T-learner, which is known as shared-base conditional mean regression \cite{Powers2018}. This modification consists of two key components. First, in the rule generation step, transformed outcomes are used to generate a sequence of rules related to HTE. Second, in the rule ensemble step, these rules serve as base functions, which are then combined using adaptive group lasso to construct separate models for the treatment and control groups. Importantly, the approach enforces a constraint ensuring that both groups share the same set of base functions, thereby enhancing comparability across treatment arms. Additionally, this structure allows us to interpret how each base function contributes to HTE estimation. The key steps of the method proposed by Wan et al.\cite{Wan2023} are as follows:

\begin{itemize}
\item \textbf{Rule Generation}
\begin{itemize}
\item Step 1: Calculating the transformed outcome

\item Step 2: Fitting a gradient boosting tree to the transformed outcome calculated in Step 1.

\item Step 3: Decomposing all base functions of the model built in Step 2 into rules. 

\end{itemize}

\item \textbf{Rule Ensemble}

\begin{itemize}

\item Step 4: Preparing data for adaptive group lasso. 

\item Step 5: Estimating the coefficients for the base functions using adaptive group lasso. 

\end{itemize}

\end{itemize}

In this process, Steps 1–-3 correspond to rule generation, whereas Steps 4–-5 correspond to rule ensemble. This framework enables direct comparison of rule coefficients between the treatment and control groups, providing insights into how each rule contributes to the estimated HTE. This method provides an interpretable rule-based model for HTE estimation while maintaining prediction performance. However, its applicability is limited to binary treatment settings and does not naturally extend to multi-arm scenarios. 

Therefore, we propose a novel HTE estimation approach based on Wan et al.\cite{Wan2023} with several modifications. First, as shown in Eq \ref{Trans_outcome}, in the multi-arm HTE estimation setting, each subject has $T$ transformed outcomes. Thus, a crucial modification in the proposed method is adapting the rule generation step to accommodate multiple transformed outcomes simultaneously. Several studies have extended RuleFit to multiple outcome estimation by incorporating rule generation techniques suitable for multi-output data. For instance, Timo et al. (2012)\cite{Timo2012} used multi-target random forests, whereas Fokkema et al. (2020)\cite{Fokkema2020} applied a multi-target boosting algorithm with conditional inference trees as base learners. By incorporating these advancements, we generalize the previous method to a multi-arm setting, creating a flexible framework for interpretable multi-arm HTE estimation that accommodates various rule generation strategies. In this study, we focus on using multi-target boosting for rule generation, following the structure of RuleFit and Wan et al. (2023). In addition, we explore different tree models as base learners within the boosting framework to assess their impact on rule generation. Specifically, we use classification and regression trees (CART) and conditional inference trees. For simplicity, the rule generation process using CART is referred to as the ``gbm'' process, whereas the one using conditional inference trees is referred to as the ``ctree'' process. Furthermore, Wan et al. (2023) employed adaptive group lasso to balance interpretability and prediction accuracy in HTE estimation. However, in settings with more than three treatment arms, the data structure becomes more complex, and adaptive group lasso may remove too many rules, potentially leading to poor predictive accuracy. To address this issue, we explore alternative group-wise regularization methods, such as group lasso, to retain a more informative set of rules while maintaining interpretability. 

This study introduces an interpretable multi-arm HTE estimation framework by extending the RuleFit-based method proposed by Wan et al.\cite{Wan2023}. Specifically, we

\begin{itemize}

\item Modify the rule generation step to accommodate multiple transformed outcomes using multi-target boosting

\item Explore the use of different base tree models (CART and conditional inference trees) for boosting-based rule generation

\item Evaluate alternative regularization techniques, comparing adaptive group lasso with group lasso to balance interpretability and predictive performance

\end{itemize}

\begin{figure}[h]
\centering
\includegraphics[width=\linewidth]{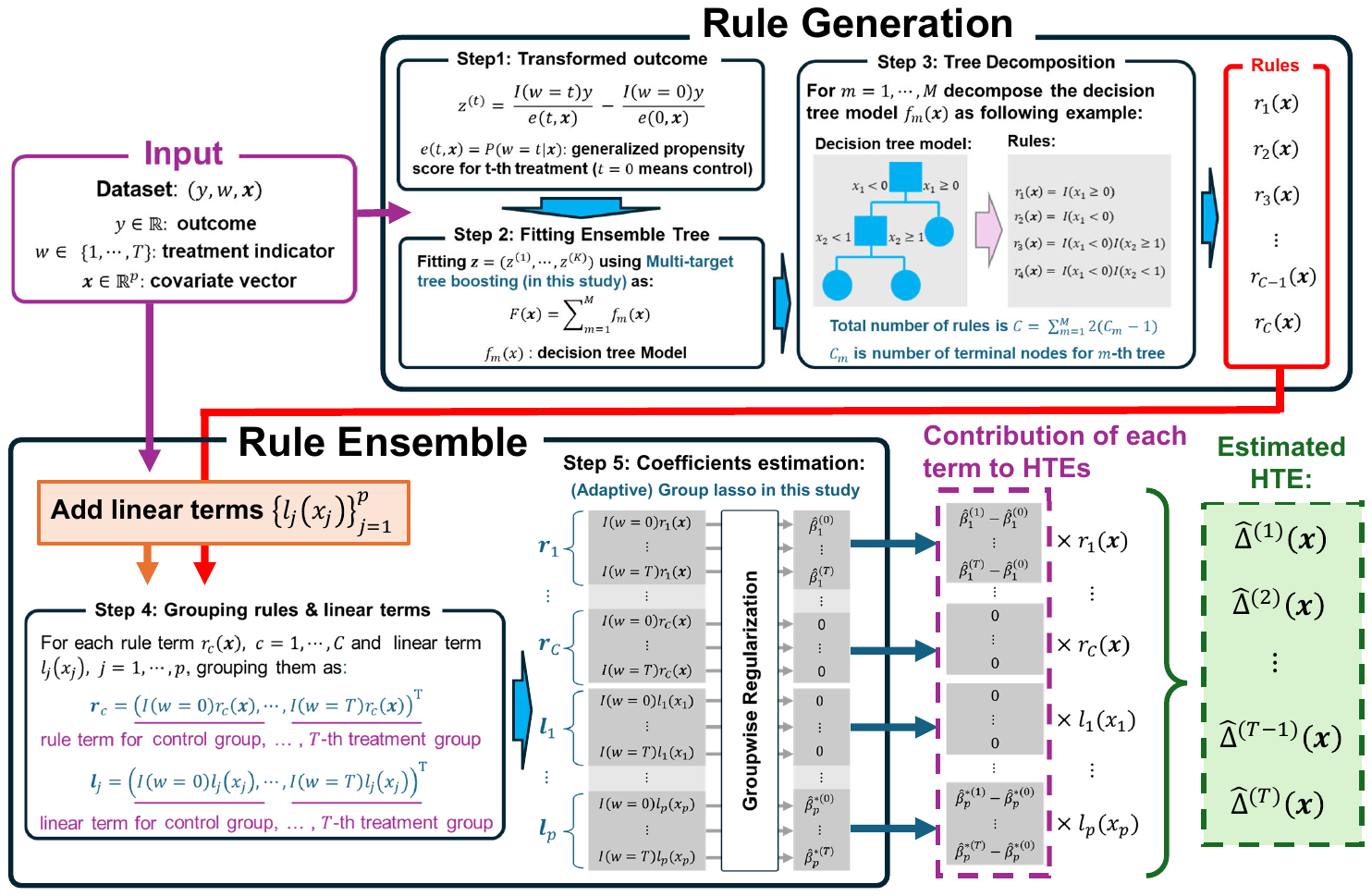}
\caption{Brief procedure of proposed approach.}
\label{Fig1}
\end{figure}

Here we show the brief procedure of proposed approach in Fig\ref{Fig1}. To clearly present our approach, we first introduce the notations and methodology of RuleFit before detailing the proposed modifications and their implementation in multi-arm HTE estimation.

\subsection{RuleFit}
RuleFit is an interpretable machine learning approach proposed by Friedman and Popescu (2008)\cite{Friedman2008}. The model was constructed as a linear combination of rules derived from data. Each rule can be interpreted as a conjunction of several simple if-then statements about covariates. Consequently, RuleFit can easily interpret the relationships between covariates and outcomes. Additionally, RuleFit balances the trade-off between interpretability and prediction accuracy, demonstrating comparable prediction accuracy to several tree-ensemble methods such as MART and random forest. Let the covariates vector be $\bm{x} = (x_1, x_2, \cdots, x_{p})^\mathrm{T} \in \mathbb{R}^p$, the RuleFit model can be defined as
\begin{align*}
F_{RuleFit}(\bm{x}) = \beta_0 + \sum_{c=1}^C\beta_cr_c(\bm{x}) + \sum_{j=1}^p\beta_j^*l_j(x_j), 
\end{align*}
where $r_c(\bm{x})$ is the rule terms of the detected decision rules; $l_j(x_j)$ is the modified version of linear terms; $C$ is the number of rule terms in the model; and $\beta_0, \beta_c$, and $\beta^*$ are the intercept, coefficient of the rule terms, and coefficient of the linear terms, respectively. The details of the rule terms and the modified version of the linear terms are described below.

\paragraph{Rule terms} Rule terms $r_c : \mathbb{R}^p \mapsto\mathbb{R}$ are defined as the conjunctions of indicator functions:
\begin{align*}
r_c(\bm{x}) = \prod_{j=1}^pI(x_{j} \in S_{jc}),
\end{align*}
where $I(\cdot)$ is an indicator function that returns 1 if the statement within parentheses is true and 0 if it is false. $S_j$ is the set of all possible values for $x_j$, and subset $S_{jc}\subseteq S_j$ is defined by interval
\begin{align*}
S_{jk} = [x_{jc}^-, x_{jc}^+),
\end{align*}
where $x_{jc}^-$ and $x_{jc}^+$ represent the lower and upper bounds of $x_j$ defined by the $c$-th rule term, respectively. 

\paragraph{Linear terms} The model built solely on the rule terms could complicate the approximation of the linear functions. Therefore, to increase model accuracy, linear terms are added to it. However, considering the robustness of linear terms to outliers, the linear terms are transformed into a ``Winsorized'' version as follows:  
\begin{align}
l_j(x_{ij}) = \min(\delta^+_j,\max(\delta_j^-,x_{ij})) \label{win_term},
\end{align}
where $\delta_j^+$ and $\delta_j^-$ are the thresholds for determining the outliers defined by the $q$-quantile and $(1-q)$-quantile of $x_j$, respectively, with a recommended value of $q = 0.025$\cite{Friedman2008}. To ensure the rule and linear terms have an equal chance of being selected, the linear term is normalized as follows:
\begin{align*}
l_j(x_{ij}) \leftarrow 0.4\cdot l_j(x_{ij})/std(l_j(x_{ij})),
\end{align*}
where $std(\cdot)$ represents the standard deviation, and 0.4 is the average standard deviation of the rule terms under the assumption that the support of the rule terms $r_c(\bm{x}_i)$ from the training data
\begin{align}
    \varrho_k = \frac{1}{N} \sum_{i = 1}^N r_c(\bm{x}_i) \label{support}
\end{align}
are distributed uniformly from $U(0,1)$\cite{Friedman2008}.

\subsection{Model of the proposed approach}
Given the dataset $\{(y_i, w_i, \bm{x_i})\}_{i=1}^N$, where $w_i \in \{0, \dots, T\}$ represents the treatment assignment ($w_i = 0$ denoting the control group and $w_i = t$ denoting the $t$-th treatment group), the proposed model is defined as
\begin{align}
f(\bm{x}_i, w_i) &= \sum_{t= 0}^TI(w_i = t)\left\{\beta_0^{(t)} + \sum_{c=1}^{C}\beta_c^{(t)}r_c(\bm{x}) + \sum_{j=1}^p\beta_j^{*(t)}l_j(x_j)\right\} 
\label{prop_mod}
\end{align}
where $\beta_0^{(t)}, \beta_c^{(t)}$ and $\beta_j^{*(t)}$ are the intercept, coefficient of the rule terms, and coefficient of the linear terms, respectively, for the $t$-th group. Terms $r_c(\bm{x})$ and $l_j(x_j)$ represent the rule and linear terms shared across the treatment and control groups, and $C$ denotes the number of rule terms in the model. 

The proposed model can also be regarded as the application of RuleFit individually to each treatment and control group while ensuring that the same rule and linear terms are shared across groups. Therefore, the outcome of the $t$-th treatment group can be estimated as
\begin{align*}
\hat{\mu}(\bm{x}_i, w_i = t) = \hat{\beta}_0^{(t)} + \sum_{c=1}^C\hat{\beta}_c^{(t)}r_c(\bm{x}) + \sum_{j=1}^p\hat{\beta}_j^{*(t)}l_j(x_j),
\end{align*}
and the outcome of the control group can be estimated as
\begin{align*}
\hat{\mu}(\bm{x}_i, w_i = 0) = \hat{\beta}_0^{(0)} + \sum_{c=1}^C\hat{\beta}_c^{(0)}r_c(\bm{x}) + \sum_{j=1}^p\hat{\beta}_j^{*(0)}l_j(x_j).
\end{align*}
where $\hat{\beta}_0^{(t)}, \hat{\beta}_c^{(t)}$, and $\hat{\beta}_j^{*(t)}$ are the estimated intercept, coefficient of the rule terms, and coefficient of the linear terms for $t$-th $(t\in \{0,1,\cdots,T\})$ group. Therefore, the contribution of base functions $r_c(\bm{x})$ and $l_j(x_j)$ to the estimated outcome can be easily interpreted based on their coefficients. The proposed approach mainly aims to build an interpretable model for HTE estimation. Therefore, it is necessary to interpret how each base function contributes to the estimated HTE, that is, we need to calculate the difference in the contribution of each base function to the outcome between the treatment and control groups. Therefore, we regard the coefficients of the same base functions as a group and enforce that they are either zero or non-zero simultaneously across all treatment groups to ensure the contributions of each base function between the treatment and control groups are comparable. Under these constraints, the HTE for the $t$-th treatment group can be estimated as 
\begin{align}
\hat{\Delta}^{(t)}(\bm{x}_i) & = \hat{\mu}(\bm{x}_i, w_i = t) - \hat{\mu}(\bm{x}_i, w_i = 0)\notag \\
& = (\beta_0^{(t)} - \beta_0^{(0)}) +  \sum_{c=1}^C(\beta_c^{(t)} - \beta_c^{(0)})r_c(\bm{x}) + \sum_{j=1}^p(\beta_j^{*(t)} - \beta_j^{*(0)})l_j(x_j) \label{prop_hte}.
\end{align}
Let $\gamma_0^{(t)} = \beta_0^{(t)} - \beta_0^{(0)}$, $\gamma_c^{(t)} = \beta_c^{(t)} - \beta_c^{(0)}$, and $\gamma_j^{*(t)} = \beta_j^{*(t)} - \beta_j^{*(0)}$, then Eq.\ref{prop_hte} can be rewritten as
\begin{align}
\hat{\Delta}^{(t)}(\bm{x}_i) = \gamma_0^{(t)} +  \sum_{c=1}^C{\gamma}_c^{(t)}r_c(\bm{x}) + \sum_{j=1}^p{\gamma}_j^{*(t)}l_j(x_j).
\end{align}

Therefore, the relationship between covariates and HTE for the $t$-th treatment group can be interpreted using the base function and their corresponding coefficients. Similarly, we can also directly compare the treatment effects between any two treatment arms while maintaining interpretability. For any $t_1 \neq t_2$, the difference in HTE can be expressed as

\begin{align}
\hat{\Delta}^{(t_1)}(\bm{x}_i) - \hat{\Delta}^{(t_2)}(\bm{x}_i) = (\beta_0^{(t_1)} - \beta_0^{(t_2)}) +  \sum_{c=1}^C(\beta_c^{(t_1)} - \beta_c^{(t_2)})r_c(\bm{x}) + \sum_{j=1}^p(\beta_j^{*(t_1)} - \beta_j^{*(t_2)})l_j(x_j) \label{prop_comp}
\end{align}
Therefore, our proposed method can estimate the HTE for each treatment arm and facilitate the comparison of treatment effects between any two arms without sacrificing interpretability. This characteristic makes our method particularly useful when comparing a treatment of interest with others and interpreting the factors driving such differences.

\subsection{Algorithm of the proposed approach}
The proposed algorithm consists of two main steps: rule generation and rule ensemble. We provide a detailed explanation of the proposed algorithm in this section.

\subsubsection{Rule Generation}
During this stage, the objective is to generate candidate rule terms for the proposed model. Our approach aims to develop interpretable models that elucidate the relationships between covariates and HTEs using base functions. This procedure consists of three steps: calculating the transformed outcome for each subject, fitting the transformed outcome to a multi-target tree boosting model\cite{Appice2007}, and converting the base functions of the model into rules.
\paragraph{Step1: Calculating the transformed outcome} Given the dataset $\{(y_i, w_i, \bm{x}_i)\}_{i=1}^N$, the transformed outcomes are calculated using Eq.\ref{Trans_outcome} as
\begin{align*}
z_i^{(t)} &= \frac{I(w_i = t)y_i}{e(t,\bm{x}_i)} - \frac{I(w_i = 0)y_i}{e(0,\bm{x}_i)},
\end{align*}
where $z_i^{(t)}$ denotes the transformed outcome of $t$-th$ (t =1,\cdots, T)$ treatment for $i$-th subject. For each subject $i$, we can then get a vector of transformed outcomes as $\bm{z}_i = \left(z_i^{(1)},\cdots, z_i^{(T)}\right)^{\mathrm{T}}$. 

\paragraph{Step2: Fitting the multi-target tree boosting model to transformed outcomes} In this step, we fit the multi-target tree boosting model to the transformed outcome data $\{(\bm{z}_i, \bm{x}_i)\}_{i=1}^N$. If the multi-target tree boosting model consists of $M$ trees, the model can be formulated as the function $F_M: \mathbb{R}^p \mapsto \mathbb{R}^T$, which is denoted as follows:
\begin{align*}
F_M(\bm{x}_i) = \sum_{m = 1}^M f_m(\bm{x}_i)
\end{align*}
where $f_m: \mathbb{R}^p \mapsto \mathbb{R}^T$ denotes the $T$ dimension base function of the tree boosting model. In this study, we consider using CART (the ``gbm'' process) and conditional inference trees (the ``ctree'' process) as the base functions. These trees can be built efficiently using R packages \texttt{rpart} and \texttt{partykit}, respectively. Both can be denoted as
\begin{align*}
f_m(\bm{x}_i) = \sum_{c = 1}^{C_m} \bm{\gamma}_c I(\bm{x}_i\in R_c)
\end{align*}
where $R_c$ denotes the $c$-th disjoint partitioned region, $\bm{\gamma}_c = (\gamma^{(1)}_c, \cdots, \gamma^{(T)}_c)^{\mathrm{T}}$ denotes the corresponding weights of $R_c$, and $C_m$ denotes the maximum number of partitioned regions for the base function created in the $m$-th boosting step. Unlike common tree boosting algorithms, we adopt the rule generation approach of RuleFit\cite{Friedman2008}, where the number of partitioned regions for base function $f_m(\bm{x}_i)$ is randomly determined as
\begin{align*}
C_m = 2 + \mathrm{floor}(\omega), \omega \sim \mathrm{exponential}(1/(\bar{L}-2)).
\end{align*}
where $\bar{L}$ is the average depth of base functions, $\mathrm{exponential}(\cdot)$ is the exponential distribution function, and $\mathrm{floor}(\omega)$ represents the largest integer less than or equal to $\omega$. This base function depth settings keeps the depth of most base functions around $\bar{L}$ and allows for the presence of several base functions with large depths. This allows the model to capture higher-order interactions while maintaining the predictive accuracy of the tree boosting model.

\paragraph{Step3: Decompose all base function of the model built in Step 2 into rules} From Step 2, we obtain a set of base functions, $\{f_m(\bm{x}_i)\}_{m=1}^M$. For each $m = 1, \cdots, M$, we decompose the tree-based function $f_m(\bm{x}_i)$ into a set of rules, $\{r_k(\bm{x}_i)\}_{k=1}^{K_m}$, where $K_m$ denotes the number of rules derived from $f_m(\bm{x}_i)$. Here, we show an example of base function decomposition as in Figure \ref{Fig1}. By decomposing all base functions $\{f_m(\bm{x}_i)\}_{m=1}^M$, we obtain a set of rule terms, represented as $\{r_k(\bm{x}_i)\}_{k=1}^K$, where $K = K_1 + \cdots + K_M$.  

\subsubsection{Rule ensemble}
During this stage, our objective is to estimate parameters for each base function, including rule and linear terms, in the proposed method. As mentioned before, to ensure interpretability of the HTE, we require that the coefficients for the same rule and linear terms be consistently zero or non-zero across all treatment groups. To achieve this, we employ group-wise regularization, including both group lasso\cite{Yuan2006} and adaptive group lasso\cite{Wang2008}, which imposes joint sparsity across groups, ensuring the simultaneous inclusion or exclusion of base functions across treatment and control arms. This stage also consists of two steps: preparing the data for group-wise regularization and using group-wise regularization to estimate the coefficients for the base functions.

\paragraph{Step 4: Preparing the data for group-wise regularization}
First, we add the ``winsorized'' version of the linear terms (Eq. \ref{win_term}) and combine them with the rule terms generated from the rule generation process to form the base functions of the model. We then prepare a dataset to apply group-wise regularization, which is used to estimate the coefficients for each base function across treatment and control groups. We group the rule terms as
\begin{align*}
\bm{r}_{ic} = \left(I(w_i = 0)r_c(\bm{x}_i), I(w_i = 1)r_c(\bm{x}_i),\cdots, I(w_i = T)r_c(\bm{x}_i)\right)^\mathrm{T} \quad \mathrm{for \ all \ } c= 1,\cdots,C,    
\end{align*}
and group the linear terms as
\begin{align*}
\bm{l}_{ij} = \left(I(w_i = 0)l_j(x_{ij}), I(w_i = 1)l_j(x_{ij}),\cdots, I(w_i = T)l_j(x_{ij})\right)^\mathrm{T} \quad \mathrm{for \ all \ } j= 1,\cdots,p.
\end{align*}
Finally, we combine the grouping rules and linear terms to create a dataset as
\begin{align*}
\mathscr{B} = \left\{(y, \bm{r}_{i1}, \cdots, \bm{r}_{iC}, \bm{l}_{i1}, \cdots, \bm{l}_{ip})\right\}_{i=1}^N.
\end{align*}

\paragraph{Step 5: Estimate the coefficients for the base functions using group-wise regularization} In this step, we apply the group-wise regularization to dataset $\mathscr{B}$. In this study, we consider both group lasso and adaptive group lasso in coefficients estimation. 

\noindent\textbf{Application of group lasso}
The coefficients for the model are estimated by minimizing the following objective functions:
\begin{align}
\left(\hat{\bm{\beta}_0}, \{\hat{\bm{\beta}}_c\}_{c=1}^C, \{\hat{\bm{\beta}}^*_j\}_{j=1}^p\right) &= \argmin_{\bm{\beta}_0, \{\bm{\beta}_c\}_{c=1}^C, \{\bm{\beta}^*_j\}_{j=1}^p} \frac{1}{2}\sum_{i=1}^N\left(y_i - \bm{\beta}_0 - \sum_{c = 1}^C\bm{\beta}^\mathrm{T}_c\bm{r}_{ic} - \sum_{j = 1}^p \bm{\beta}^{*\mathrm{T}}_j\bm{l}_{ij} \right)^2 \notag \\
& + \lambda \sqrt{T}\left(\sum_{c=1}^C ||\bm{\beta}_c||_2 + \sum_{j=1}^p ||\bm{\beta}_j^*||_2\right)
\label{group_est}
\end{align}
where $||\cdot||_2$ denotes the L2-norm, $\bm{\beta}_0 = \left(\beta_0^{(0)}, \beta_0^{(1)}, \cdots, \beta_0^{(T)}\right)^\mathrm{T}$ is the intercept vector, $\bm{\beta}_c = \left(\beta_c^{(0)}, \beta_c^{(1)}, \cdots, \beta_c^{(T)}\right)^\mathrm{T}$ is the coefficient vector for the $c$-th rule term, and $\bm{\beta}_j^* = \left(\beta_j^{*(0)}, \beta_j^{*(1)}, \cdots, \beta_j^{*(T)}\right)^\mathrm{T}$ is the coefficient vector for the $j$-th linear term. $\lambda$ is the tuning parameter determined using 10-fold cross-validation. The group lasso can be rapidly implemented in the R package $grpreg$.  

\noindent\textbf{Application of adaptive group lasso}
The coefficients for the model are estimated in two steps\cite{Huang2010}. First, adaptive weights $w_c$ and $w^*_j$ are calculated using the estimated coefficients in Eq.\ref{group_est} as
\begin{align*}
w_c = 
\begin{cases}
||\hat{\bm{\beta}}_c||_2^{-1},\quad &\mathrm{if} \ ||\hat{\bm{\beta}}_c||_2 > 0\\
\infty,\quad &\mathrm{if} \ ||\hat{\bm{\beta}}_c||_2 = 0
\end{cases}
\qquad \mathrm{and} \qquad
w^*_j = 
\begin{cases}
||\hat{\bm{\beta}}_j^*||_2^{-1},\quad &\mathrm{if} \ ||\hat{\bm{\beta}}_j^*||_2 > 0\\
\infty,\quad &\mathrm{if} \ ||\hat{\bm{\beta}}_j^*||_2 = 0
\end{cases}
\quad.
\end{align*}
Second, the coefficients for the model are estimated as
\begin{align*}
\left(\hat{\bm{\alpha}_0}, \{\hat{\bm{\alpha}}_c\}_{c=1}^C, \{\hat{\bm{\alpha}}^*_j\}_{j=1}^p\right) &= \argmin_{\bm{\alpha}_0, \{\bm{\alpha}_c\}_{c=1}^C, \{\bm{\alpha}^*_j\}_{j=1}^p} \frac{1}{2}\sum_{i=1}^N\left(y_i - \bm{\alpha}_0 - \sum_{c = 1}^C\bm{\alpha}^\mathrm{T}_c\bm{r}_{ic} - \sum_{j = 1}^p \bm{\alpha}^{*\mathrm{T}}_j\bm{l}_{ij} \right)^2 \notag \\
& + \lambda' \sqrt{T}\left(\sum_{c=1}^C w_c||\bm{\alpha}_c||_2 + \sum_{j=1}^p w^*_j||\bm{\alpha}_j^*||_2\right)
\end{align*}
where $\bm{\alpha}_0 = \left(\alpha_0^{(0)}, \alpha_0^{(1)}, \cdots, \alpha_0^{(T)}\right)^\mathrm{T}$ is the intercept vector, $\bm{\alpha}_c = \left(\alpha_c^{(0)}, \alpha_c^{(1)}, \cdots, \alpha_c^{(T)}\right)^\mathrm{T}$ is the coefficient vector for the $c$-th rule term, and $\bm{\alpha}_j^* = \left(\alpha_j^{*(0)}, \alpha_j^{*(1)}, \cdots, \alpha_j^{*(T)}\right)^\mathrm{T}$ is the coefficient vector for the $j$-th linear term. $\lambda'$ is the tuning parameter determined using 10-fold cross-validation. Therefore, using the adaptive group lasso, the intercept, coefficient of the rule terms, and coefficient of the linear terms of the model in Eq.\ref{prop_mod} are estimated as $\bm{\alpha}_0, \bm{\alpha}_c $, and $\bm{\alpha}_j^*$, respectively. 

\subsection{Interpretation tools}
The proposed approach provides a framework for building rule-based interpretable models for multi-arm HTE estimation. This allows the interpretation of the relationship between covariates and HTE based on base functions and their corresponding coefficients. However, the constructed model includes many base functions. While it is essential to incorporate a diverse set of rules and linear terms to comprehensively capture the heterogeneity of treatment effects, focusing on all outcomes in a real data analysis can lead to complexity and confusion. Therefore, in real-world analyses, the focus is typically on the base function or variables that are contributing more to the HTE. To support this, we introduce interpretability tools: base function importance and variable importance, which rank the contributions of each base function and variable, respectively. We generalize the base function importance and variable importance measures of \cite{Wan2023} and provide a detailed description as follows:

\noindent{\textbf{Base function importance}} The base function importance includes the importance of the rule and linear terms. A high or low base function importance value indicates that the corresponding base function contributes more or little to the HTE, respectively. Here, we modify the base function importance based on the original functions and determine the importance of the rule and linear terms for the $t$-th treatment group as follows: 
\begin{align}
\displaystyle
I^{(t)}_c &= \left|\hat{\beta}^{(t)}_c -\hat{\beta}^{(0)}_c\right|\cdot\sqrt{\varrho_c(1-\varrho_c)} \quad \mathrm{and} \label{rule_imp} \\
I^{(t)}_j &= \left|\hat{\beta}^{*(t)}_j - \hat{\beta}^{*(0)}_j\right|\cdot std\left(l_j(x_j)\right), \label{lin_imp}
\end{align}
respectively, where $\varrho_c$ is the support for the rules, as shown in Eq. \ref{support}, and $std\left(l_j(x_j)\right)$ is the standard deviation of $l_j(x_j)$ for training data.

\noindent{\textbf{Variable importance}} Variable importance is a widely used approach for post-hoc interpretation of black-box machine learning models. It provides insights into the ranking of variables based on their contributions to the outcomes. In this study, our primary interest lies in multi-arm HTE estimation; therefore, variable importance is employed to rank the contribution of each variable to the HTEs. The variable importance for the $t$-th treatment group is computed as follows:
\begin{align}
\displaystyle
I^{*(t)}_{j}(\bm{x}) = I_j^{(t)}(x_j) + \sum_{x_j\in r_c} \frac{I^{(t)}_c(\bm{x})}{m_c}, \label{var_imp}
\end{align}
where the first term $I_j(x_j)$ denotes the importance of the $j$ th linear term, and the second term denotes the sum of the importance of the rules that contain $x_j \ (x_j\in r_c)$, $m_c$ is the total number of variables $x_j$ used to define the rule.

Furthermore, in multi-arm settings, comparisons between any two treatment groups can also be considered. Therefore, we also provide base function importance and variable importance specifically for pairwise treatment comparisons, allowing for a more detailed examination of the factors influencing differences in treatment effects. Accordingly, for any treatment groups $t_1$ and $t_2$ with $t_1\neq t_2$, the importance of the rule and linear terms is computed as follows:
\begin{align}
\displaystyle
I_c &= \left|\hat{\beta}^{(t_1)}_c -\hat{\beta}^{(t_2)}_c\right|\cdot\sqrt{\varrho_c(1-\varrho_c)} \quad \mathrm{and} \label{rule_imp} \\
I_j &= \left|\hat{\beta}^{*(t_1)}_j - \hat{\beta}^{*(t_2)}_j\right|\cdot std\left(l_j(x_j)\right), \label{lin_imp}
\end{align}
respectively, and the variable importance is computed as follows:
\begin{align}
\displaystyle
I^*_{j}(\bm{x}) = I_j(x_j) + \sum_{x_j\in r_c} \frac{I_c(\bm{x})}{m_c}. \label{var_imp}
\end{align}

\section{Simulation studies}\label{sec4}

We generated various synthetic datasets to evaluate the estimation performance of the proposed approach. To ensure a comprehensive assessment, we considered diverse HTE and data generation processes and employed multiple evaluation metrics. As previously discussed, our proposed method serves as a framework for multi-arm HTE estimation, incorporating CATE and conditional inference trees for rule generation, as well as group lasso and adaptive group lasso for the rule ensemble process during model construction. Accordingly, this simulation study compared different model-building strategies under various settings. Additionally, to ensure a fair evaluation, we compared the performance of the proposed approach against meta-learners, a widely used existing framework. For each meta-learner, we employed commonly used models in HTE estimation, including XGBoost, random forest, and BART. This section consists of two parts. First, we explain the details of the simulation design. Second, simulation results for different scenarios are presented.

\subsection{Simulation design}
For each simulation scenario, we created a pair of datasets in the form of $\{(y_i,t_i,\bm{x}_i)\}_{i=1}^N$, one for training and the other for testing. To comprehensively evaluate the proposed approaches, we considered treatment settings with three to five groups (including the control) under both RCT and observational conditions, yielding $3\times2 = 6$ distinct patterns for generating treatment indicators. In addition, we generated outcomes using three different models for the true main effect and three for the true treatment effect, resulting in $3\times3 = 9$ outcome generation. These design choices yielded $6\times9 = 54$ different simulation scenarios. The detailed design of covariates $\bm{x}_i$, treatment indicator $t_i$, outcome $y_i$, and true HTE is as follows:

\noindent\textbf{Covariates}: We considered 10 covariates as $\bm{x}_i = (x_{i1}, x_{i2}, \cdots,x_{i10})^\mathrm{T}$, with five odd-numbered variables $x_{i1}, x_{i3}, \cdots, x_{i9} \overset{iid}{\sim} N(0,1)$ being continuous and five even-numbered variables $x_{i2}, x_{i4}, \cdots, x_{i10} \overset{iid}{\sim} Bernoulli(0.5)$ being binary. 

\noindent\textbf{Treatment indicator:} We considered treatment indicator $w_i \in \{0,1,\cdots, T\}$ with three levels $T \in \{2, 3, 4\}$. This setup allowed for three, four, or five distinct treatment levels, including the control group ($w_i = 0$). The treatment indicator $t_i\in\{0,1,\cdots, T\}$ was generated from the multinomial distribution and to account for the situation of the RCT and observational study, we incorporated two different treatment assignment mechanisms:

\begin{itemize}
\item{\textit{RCT setting}}: Individual $i$ was randomly assigned into the control and treatment groups; therefore, the probabilities to be assigned into each group are
\begin{align*}
\mathbb{P}(w_i = 0) = \cdots = \mathbb{P}(w_i = T) = \frac{1}{T + 1}.
\end{align*}
\item{\textit{Observational study setting}}: In this setting, selection bias was considered among three groups.
We generated the probability of treatment assignment for individual $i$ using multinomial logistic regression. Therefore, the probability to be assigned to the $t$-th treatment group is
\begin{align*}
\mathbb{P}(w_i = t) = \frac{f^{(t)}(\bm{x}_i)}{1 + \sum_{t=1}^{T}f^{(t)}(\bm{x}_i)}
\end{align*}
where $f^{(t)}(\bm{x}_i)$ is denoted as
\begin{align*}
f^{(1)}(\bm{x}_i) &= \exp\left(-0.50 - 0.1x_{i1} - 0.2x_{i2} - 0.3x_{i3} + 0.2x_{i4} - 0.7x_{i5}\right),\\
f^{(2)}(\bm{x}_i) &= \exp\left(-0.75 - 0.2x_{i1} - 0.4x_{i2} - 0.6x_{i3} + 0.4x_{i4} - 0.3x_{i5}\right),\\
f^{(3)}(\bm{x}_i) &= \exp\left(-1.00 - 0.2x_{i1} - 0.5x_{i2} - 0.5x_{i3} + 0.5x_{i4} - 0.3x_{i5}\right),\\
f^{(4)}(\bm{x}_i) &= \exp\left(-1.50 - 0.3x_{i1} - 0.4x_{i2} - 0.2x_{i3} + 0.4x_{i4} - 0.1x_{i5}\right),
\end{align*} 
respectively, and the probability to be assigned to the control group is
\begin{align*}
\mathbb{P}(w_i = 0) = 1 - \sum_{t=1}^T\mathbb{P}(w_i = t).
\end{align*}
\end{itemize}

\noindent\textbf{Outcomes:} The outcome for individual $i$ was obtained from the normal distribution as
\begin{align*}
y_i \sim N\left(f(\bm{x}_i, w_i), 1 \right) 
\end{align*}
where $f(\bm{x}_i, t_i)$ is the true data generation model denoted as 
\begin{align*}
f(\bm{x}_i,w_i) = \mu(\bm{x}_i) + \sum_{t=0}^TI(w_i=t)\delta_{t}(\bm{x}_i),
\end{align*}
where $\mu(\bm{x}_i)$ is the function of the main effect, and $\delta_{w_i}(\bm{x}_i)$ is the function of treatment effect. To create the simulation dataset in different situations, we considered linear (Eq.\ref{M1}), stepwise (Eq.\ref{M2}), and nonlinear (Eq.\ref{M3}) functions for the main effect function $\mu(\bm{x}_i)$ as
\begin{align}
M1: \mu(\bm{x}_i) &= 0.6x_{i1} + 0.9x_{i2} + 0.6x_{i3} - 0.9x_{i4} + 0.6x_{i5} \label{M1},\\
M2: \mu(\bm{x}_i) &= 1.2I(x_{i1} > -1)I(x_{i3} < 1) - 1.2I(x_{i2} < 0.5) - 1.2I(x_{i3} > -1)I(x_{i5} < 1) + 1.2I(x_{i4} > 0.5) \label{M2},\\
M3: \mu(\bm{x}_i) &= 0.6x_{i1}^2 + 0.5x_{i2}x_{i3} - 1.2\cos(\pi x_{i4}x_{i5}) \label{M3},
\end{align}
as well as linear (Eq.\ref{T2}), stepwise (Eq.\ref{T3}), and nonlinear (Eq.\ref{T4}) functions for the treatment effect function $\delta_k(\bm{x}_i)$ as 
\begin{align}
T1: \delta_t(\bm{x}_i) &= \beta_1^{(t)}\left(0.5x_{i1} + x_{i2}\right) + \beta_2^{(t)}\left(0.5x_{i3} + x_{i4}\right) 
+ \beta_3^{(t)}\left(0.5x_{i5} + x_{i2}\right) \label{T2},\\ 
T2: \delta_t(\bm{x}_i) &= \beta_1^{(t)}\left(1.4I(x_{i1} > 0) - 0.3I(x_{i2} > 0.5)\right) + \beta_2^{(t)}\left(1.4I(x_{i3}>0) - 0.3I(x_{i4}>0.5)\right) \notag \\
&+ \beta_3^{(t)}\left(1.4I(x_{i5}>0) - 0.3I(x_{i2}>0.5)\right) \label{T3},\\
T3: \delta_t(\bm{x}_i) &= \beta_1^{(t)}\left(0.75\sin(x_{i1}) + x_{i2}\right) + \beta_2^{(t)}\left(0.75\sin(x_{i3}) + x_{i4}\right) + \beta_3^{(t)}\left(0.75\sin(x_{i5}) + x_{i2}\right)\label{T4},
\end{align}
where parameters vectors $\left(\beta_1^{(t)}, \beta_2^{(t)}, \beta_3^{(t)}\right)$ are defined as 
\begin{align*}
\left(\beta_1^{(t)}, \beta_2^{(t)}, \beta_3^{(t)}\right) = 
\begin{cases}
& (2, 2, 2) \qquad \mathrm{for} \quad t = 0 \\
& (-1, 2, 4) \qquad \mathrm{for} \quad t = 1 \\
& (3, 3, -1) \qquad \mathrm{for} \quad t = 2 \\
& (-3, 3, 1) \qquad \mathrm{for} \quad t = 3 \\
& (-1, 4, 1) \qquad \mathrm{for} \quad t = 4 
\end{cases}
\end{align*} 

\noindent\textbf{True HTE} The true HTE for the $t$-th treatment group is generated as  
\begin{align*}
\Delta_t(\bm{x}_i) = f(\bm{x}_i, w_i = t) - f(\bm{x}_i, w_i = 0)
\end{align*}

\subsection{Evaluation Metrics}

To evaluate the estimation performance of the proposed approach using different model construction methods and compare it with meta-learners employing various models, we consider four different metrics. 

\noindent\textbf{Precision in the estimation of heterogeneous effect (PEHE)} This is a commonly used measure for assessing the prediction accuracy of HTE estimation methods\cite{Hill2011,Acharki2023}. Because we consider multi-arm settings, we computed the overall PEHE by averaging the PEHE values across all treatment groups as follows:
\begin{align*}
\sqrt{\frac{1}{K}\sum_{k=1}^K\left[\frac{1}{N}\sum_{i=1}^N\left(\Delta_k(\bm{x}_i) - \hat{\Delta}_k(\bm{x}_i)\right)^2\right]},
\end{align*}
where $\Delta_k(\bm{x}_i)$ represents the true the, and 
$\hat{\Delta}_k(\bm{x}_i)$ denotes the estimated HTE for the $k$-th treatment arm. A higher mean of PEHE indicates lower prediction accuracy.

\noindent\textbf{Absolute Relative bias} This metric also evaluates the prediction accuracy of HTE estimation by quantifying the deviation of the estimated treatment effects from the true effects. Here, we also computed the overall absolute bias by averaging the absolute relative bias values across all treatment groups as follows:
\begin{align*}
\frac{1}{K}\sum_{k=1}^K\left|\frac{\frac{1}{N}\sum_{i=1}^N\left(\Delta_k(\bm{x}_i) - \hat{\Delta}_k(\bm{x}_i)\right)}{\frac{1}{N}\sum_{i=1}^N\Delta_k(\bm{x}_i)}\right|.
\end{align*}
Therefore, lower absolute relative bias values indicate higher estimation accuracy.

Furthermore, in multi-arm treatment estimation, identifying the optimal treatment and ranking the treatment effects are crucial. To assess these aspects, we introduce two additional metrics. 

\noindent\textbf{Cohen's Kappa} This metric evaluates the agreement in the best treatment decision based on the estimated HTE and true HTE. It takes values between $-1$ and $1$, where higher values indicate greater agreement between the estimated and true best treatment assignments. Additionally, according to the interpretation by \cite{Landis1977}, a Cohen's Kappa value above 0.61 is considered substantial agreement, whereas a value exceeding 0.81 is regarded as almost perfect agreement.

\noindent\textbf{Spearman’s rank correlation} This metric measures the agreement between the estimated and true rankings of treatment effectiveness. We ranked the treatments based on both the estimated and true HTE and used Spearman’s rank correlation to assess their agreement for each subject. It takes values between $0$ and $1$, where higher values indicate greater agreement between the estimated and true treatment effect rankings. To obtain an overall measure, we computed the average Spearman’s rank correlation by averaging the individual Spearman’s rank correlation values across all subjects. A higher average Spearman’s rank correlation indicates that a greater proportion of subjects exhibit strong agreement between the estimated and true treatment rankings.

\noindent\textbf{Number of base functions for proposed approach}: This metric was used to measure the number of base functions with the model created by the proposed approach. The HTE estimation model for our proposed approach built in the proposed framework is an additive of rule and linear terms; as fewer terms are included in the model, the results are easier to interpret. Therefore, in this study, we used the number of terms to evaluate model complexity.

\subsection{Comparison Method}
In this simulation study, we evaluated the performance differences in the proposed approach under different rule generation and ensemble processes and compared it with existing HTE estimation frameworks to assess its usefulness. 

For a comprehensive evaluation of existing HTE estimation frameworks, we considered all commonly used meta-learners, including S-learner, T-learner, X-learner, M-learner, DR-learner, R-learner, and reference-free R-learner. For each meta-learner, we used XGBoost, random forest, and BART as base models to ensure a thorough comparison. $7\times3 = 21$ previous methods were considered. XGBoost, random forest, and BART were implemented using R packages \texttt{xgboost}, \texttt{ranger}, and \texttt{BART}, respectively. 

For our proposed approach, we evaluated different configurations in both the rule generation and ensemble steps. Specifically, in the rule generation step, we considered the implementation of both ``gbm'' and ``ctree'' processes. In the ensemble step, we applied both group lasso and adaptive group lasso to evaluate their impact on model performance. $2\times2 = 4$ different model generation processes were considered in the proposed approach. In implementing our proposed method, we followed the parameter settings recommended by Friedman and Popescu (2008): the number of base learners, mean depth of each base learner, and shrinkage rate for boosting steps were set to 333, 2, and 0.01, respectively. Additionally, because some methods such as X-learner, M-learner, DR-learner, R-learner, and our proposed approach require propensity score adjustments, we ensured consistency across methods by estimating generalized propensity scores using multinomial logistic regression, implemented via R package \texttt{nnet}.

For simplicity, the meta-learners are denoted as S-learner using BART (sbart),T-learner using BART (tbart), X-learner using BART (xbart), M-learner using BART (mbart), DR-learner using BART(drbart), and R-learner using BART (rbart). The proposed approaches are denoted as gbm.gl ("gbm" rule generation with group lasso for rule ensemble), gbm.agl ("gbm" rule generation with adaptive group lasso for rule ensemble), ctree.gl ("ctree" rule generation with group lasso for rule ensemble), and ctree.agl ("ctree" rule generation with adaptive group lasso for rule ensemble). 

\subsection{Simulation results}
Here, we present the prediction performance of the proposed approach and meta-learners. Because meta-learners using BART tend to perform better than those using xgboost or random forest in most metrics, and the M- and reference free R-learner perform much worse than other meta-learners in most scenarios, we focused on comparing the results of S-, T- X-, DR- and R-learners using BART and the proposed approach for simplicity. The full results of the simulations for the proposed approach and the meta-learner using BART are shown in Appendix \ref{apd1}, and the full results of the simulation of the meta-learner using xgboost and random forest are shown in Appendix \ref{apd2}. The simulation results indicate that our proposed approach effectively balanced the trade-off between interpretability and prediction accuracy. The results in terms of mPEHE and absolute relative bias show that the proposed approach has comparable prediction accuracy to the meta-learners using BART and even performed better when the treatment effect was created from linear functions. Furthermore, the proposed approach performed well in selecting the optimal treatment for each subject. Compared with the meta-learners, the proposed approach achieved higher Cohen’s kappa values in most scenarios. In addition, the proposed method performed well in ranking the effectiveness of treatment when the treatment effect was generated from liner or nonlinear functions; however, it performed poorly when the treatment effect was generated from a stepwise function, and the meta-learners were better under this condition. The detailed simulation results are as follows: 

\begin{figure}[tb]
\centering
\includegraphics[width=\linewidth]{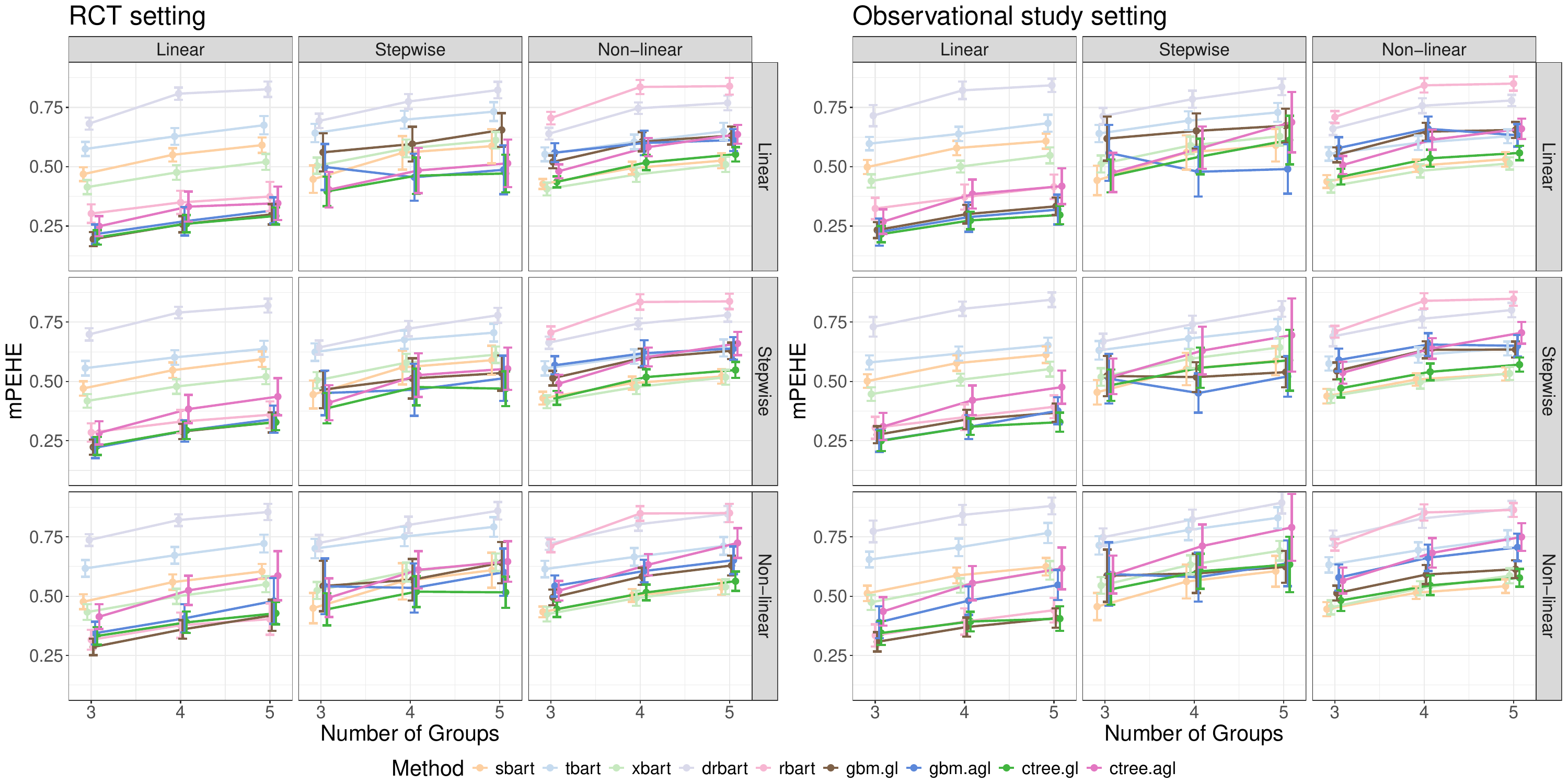}
\caption{Results of mean mPEHE across twelve scenarios for all approaches. The plots in the first column illustrate the performance of meta-learners using BART under both RCT (left) and observational settings (right). In each plot, the $x$-axis represents the number of treatment groups while the $y$-axis denotes the mPEHE. These line charts depict the mean mPEHE along with their standard deviations, represented by error bars.}
\label{mPEHE_fig}
\end{figure}

Figure \ref{mPEHE_fig} shows the mPEHE results comparing the meta-learners using BART and the proposed approaches in both RCT and observational study settings, where lower mPEHE indicates higher prediction accuracy. 
The prediction accuracy of the proposed approaches was comparable to that of the meta-learners when the treatment effect was generated from a stepwise or nonlinear function, and most of them tended to show a higher prediction accuracy than the meta-learners when the treatment effect was generated from a linear function. Focusing on the best performing meta-learners and the proposed approach, the proposed approach achieved higher prediction accuracy than meta-learners in most of the scenarios. Furthermore, the prediction accuracy decreased as the number of groups increased in both the meta-learners and proposed approaches. Particularly, R-learner achieved the highest prediction accuracy among the meta-learners when the treatment effect was generated from linear functions in both RCTs and observational study settings. In this setting, the prediction accuracy of the proposed method using group lasso for rule ensemble (gbm.gl and ctree.gl) was higher than that achieved using adaptive group lasso for rule ensemble (gbm.agl and ctree.agl). The prediction accuracies of most of the proposed approaches (gbm.gl, ctree.gl, and gbm.agl) were higher than those of the R-learners when the main effect was generated from a linear or stepwise function. Even when the main effects were generated from nonlinear functions, the proposed approach with group lasso achieved better results than the R-learner. When the treatment effect was generated from stepwise functions, the S- and X-learners consistently outperformed other meta-learners in both RCT and observational study settings. In these settings, the prediction accuracy of the proposed approach using the ``ctree'' rule generation and group lasso for rule ensemble (ctree.gl) was the best in the RCT settings. The proposed approach using the ``gbm'' rule generation and adaptive group lasso (gbm.agl) also showed comparable results when the main effect was generated from a linear or stepwise function, specifically when the number of groups was 4 or 5. In observational study settings, the prediction accuracy of the proposed approach using ``gbm'' rule generation and adaptive group lasso for rule ensemble (gbm.agl) was the best. S- and X-learners outperformed the best-performing proposed approaches (ctree.gl and gbm.agl) only when the number of groups was 3, and the best-performing proposed approaches tended to achieve higher prediction accuracy when the number of groups was 4 and 5. When the treatment effect was generated from the nonlinear function, the S- and X-learners still achieved the best performance among meta-learners. In this setting, the proposed approach using the ``ctree'' rule generation and group lasso for rule ensemble (ctree.gl) performed best in both RCT and observational study settings. The best proposed approach (ctree.gl) did not outperform the S- or X-learners in terms of prediction accuracy. This is because, in order to preserve model interpretability, the proposed approach employed an additive model structure comprising rule and linear terms as base functions. Such a structure may limit the model’s capacity to capture complex nonlinear HTE. However the X- and S-learners can built more complex models than the proposed approach to fit the complex nonlinear HTE. Therefore, the X- and S-learners are naturally more suitable for complex nonlinear HTE estimation than the proposed approach. However, the best performing proposed approach still achieved comparable results to the X- or S-learner. Therefore, we consider that the proposed approach balances interpretability and prediction accuracy effectively.

\begin{figure}[tb]
\centering
\includegraphics[width=\linewidth]{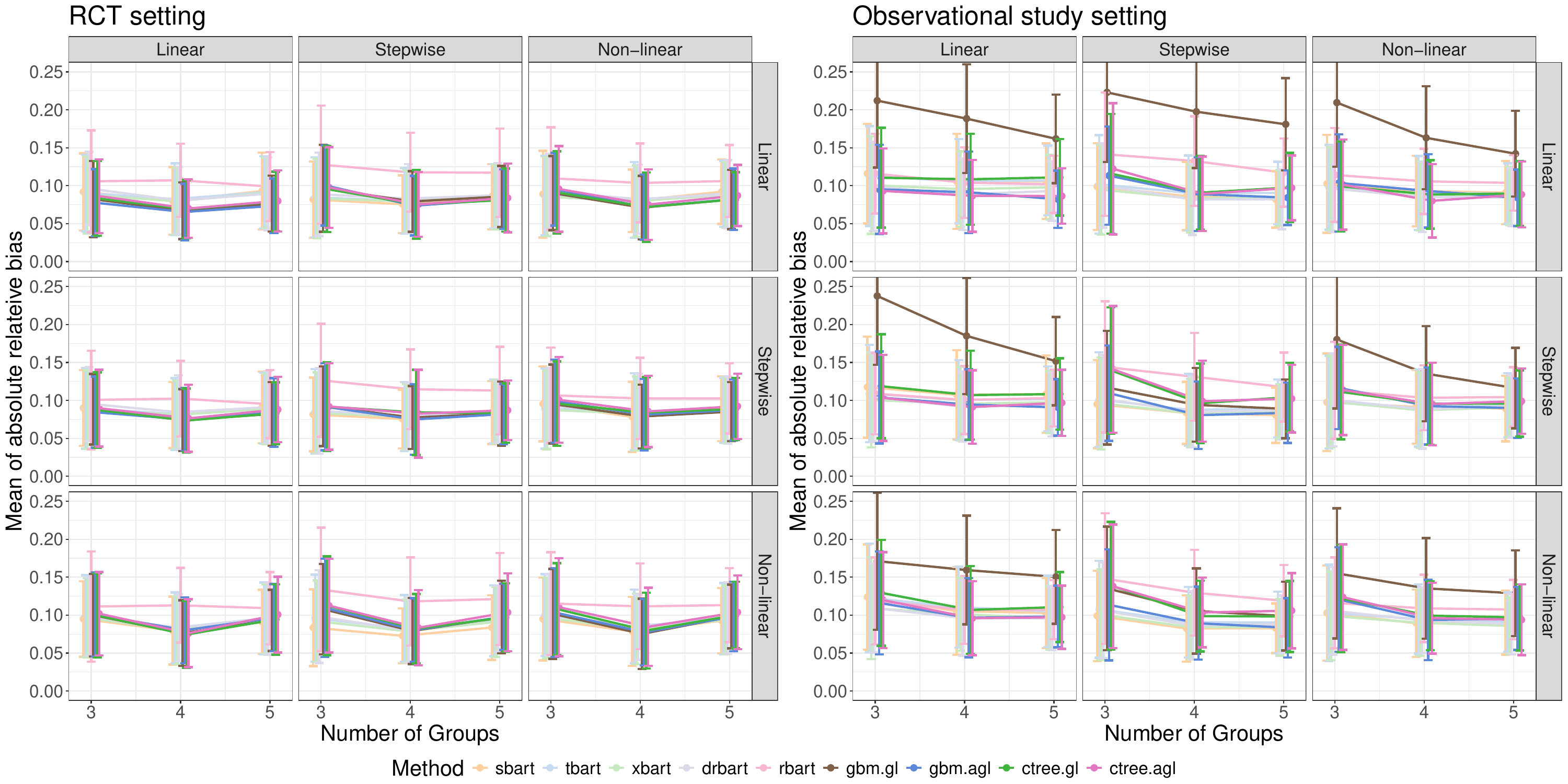}
\caption{Results in terms of the mean absolute relative bias across twelve scenarios for all approaches. In each plot, the x-axis represents the number of treatment groups, while the $y$-axis denotes the mean absolute relative bias across. These line charts depict the average of mean absolute relative bias along with their standard deviations, represented by error bars.}
\label{mabias_fig}
\end{figure}

Figure \ref{mabias_fig} compares the meta-learners using BART and the proposed approaches in both RCT and observational study settings in terms of absolute relative bias. 
\begin{comment}
In the RCT setting, both the meta-learners and the proposed methods exhibited similarly low bias. In the observational study setting, the proposed approaches using adaptive group lasso (gbm.agl and ctree.agl) tended to achieve lower bias than the meta-learners using BART when treatment effects were generated from linear functions. However, they did not outperform the meta-learners using BART when the treatment effects were generated from stepwise or nonlinear functions. Focusing on the proposed approach across different model building processes, all methods exhibited similar levels of bias in the RCT setting. However, in the observational study setting, the bias was generally lower when using adaptive group lasso with "gbm" based rule generation compared to using group lasso. In contrast, for "ctree" rule generation, the bias was comparable between adaptive group lasso and group lasso. These results suggest that, overall, the use of adaptive group lasso in the proposed approach tends to yield lower bias than group lasso, particularly when combined with "gbm" rule generation. 
\end{comment}

Most proposed approaches (gbm.agl, ctree.gl, and ctree.agl) consistently showed comparable low bias to the best performing meta-learners: S-, T-, X-, and DR-learner, for all the scenarios in both the RCT and observational study settings. For most of the meta-learners and proposed approaches, their biases did no change as the number of groups increased. Particularly, in the RCT settings, S-learner, T-learner, X-learner, DR-learner, and all proposed approaches had lower bias in all scenarios. In the observational study setting, the bias of the proposed approachwith the ``gbm'' rule generation and group lasso (gbm.gl) was higher than that of the other proposed methods in most scenarios, whereas the other proposed methods still showed low bias. In particular, the proposed approachwith the ``gbm'' rule generation and adaptive group lasso for rule ensemble (gbm.agl) performed better than the other proposed approaches in most scenarios. Therefore, the proposed approachwith the ``gbm'' rule generation and adaptive group lasso rule ensemble (gbm.agl) could be preferred in observational study settings.

\begin{figure}[tb]
\centering
\includegraphics[width=\linewidth]{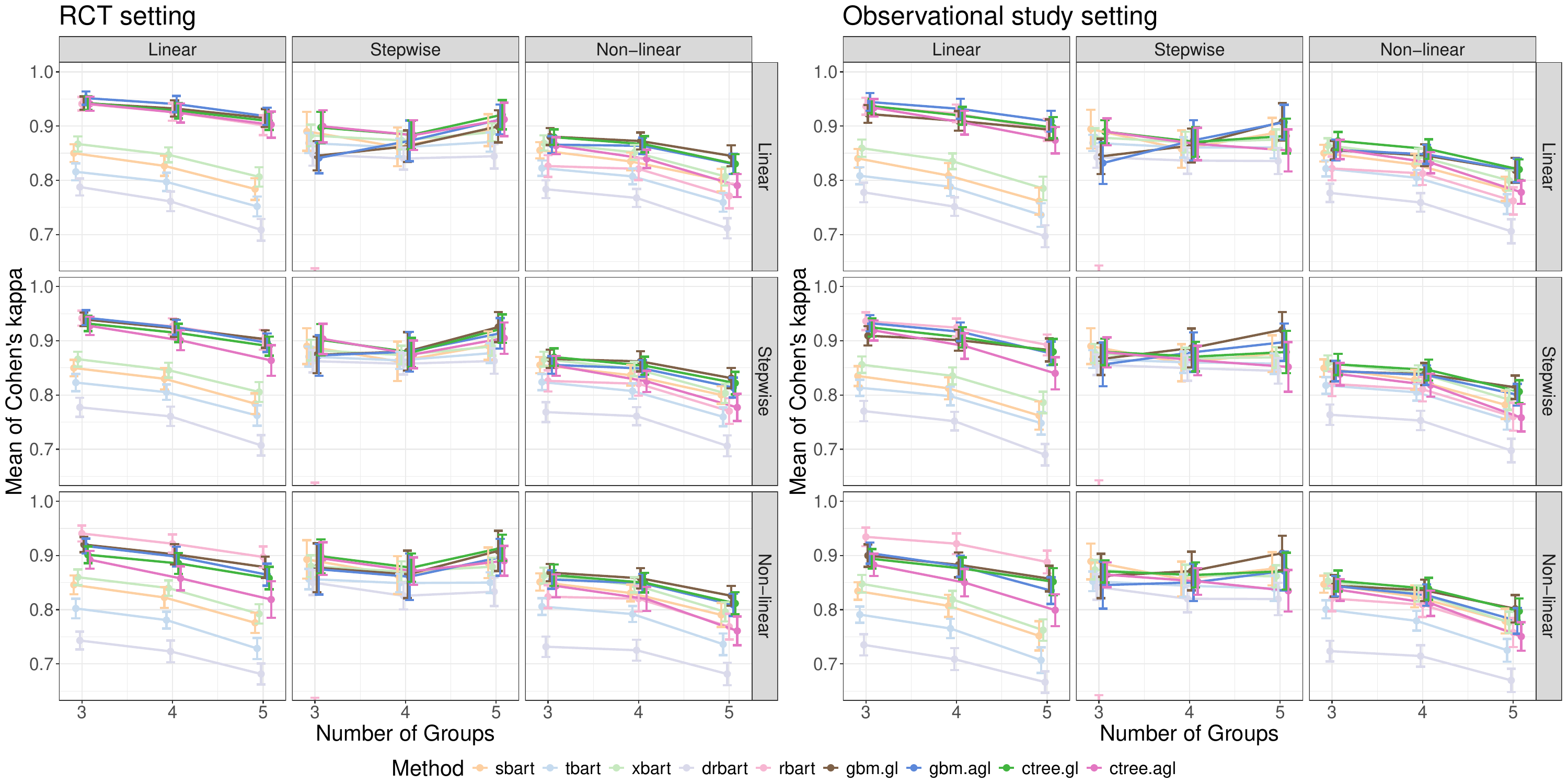}
\caption{Results in terms of mean Cohen's kappa across twelve scenarios for all approaches. The plots in the first column illustrate the performance of meta-learners using BART under both RCT (rct) and observational (obv) settings, whereas the plots in the second column display the results of the proposed approach in the same settings. In each plot, the x-axis represents the number of treatment groups, while the y-axis denotes the Spearman’s rank correlation. These line charts depict the mean Cohen's kappa along with their standard deviations, represented by error bars.}
\label{mkappa_fig}
\end{figure}

Figure \ref{mkappa_fig} compares meta-learners using BART and the proposed approaches in both RCT and observational settings in terms of average Cohen's kappa, where higher values indicate more correct optimized treatment selection. The average Cohen's Kappa values for all proposed approaches were above 0.8 in most scenarios, both in the RCT and observational study settings. Therefore, all the proposed approaches were able to select the most optimal treatment for each subject. The S-, T-, and X-learners showed mean Cohen's Kappa values above 0.8 when the number of groups was 3. Furthermore, when the treatment effect was generated from linear or nonlinear functions, the average Cohen's kappa values of both the proposed approach and the meta-learners tended to decrease, with most of the proposed approaches able to maintain a value above 0.8, whereas the meta-learners tended to fall below 0.8 when the number of groups reached 5. Thus, in these simulation studies, the proposed approach was able to select the optimal treatment for each subject more accurately than the meta-learners. Specifically, when the treatment effect was generated from a linear function, the Cohen's kappa values of R-learners were above 0.9 and much higher than those of other meta-learners in both RCT and observational settings. In these settings, the proposed approaches also showed comparably high Cohen's kappa when the main effect was generated from linear and stepwise function. Although the Cohen's kappa values of the proposed approaches were lower than those of R-learners when the main effect was created from nonlinear function, they were still better than those of other meta-learners. When the treatment effect was generated from the stepwise functions, the average Cohen's kappa did not considerably decrease with an increase in the number of groups for both meta-learners and the proposed approach. In these settings, the S-learner, T-learner, X-learner, DR-learners, and proposed approach showed high Cohen's kappa values in both RCT and observational study settings. When the treatment effect was generated from nonlinear functions, the average Cohen's kappa values for the S-learners, T-learners, X-learners, R-learners, and proposed approaches were similar and above 0.8 when the number of groups was 3. However, when the number of groups was increased to 5, the average Choden’s kappa values of the proposed approach and the meta-learners, except X-learners, tend to fell below 0.8.       

\begin{figure}[tb]
\centering
\includegraphics[width=\linewidth]{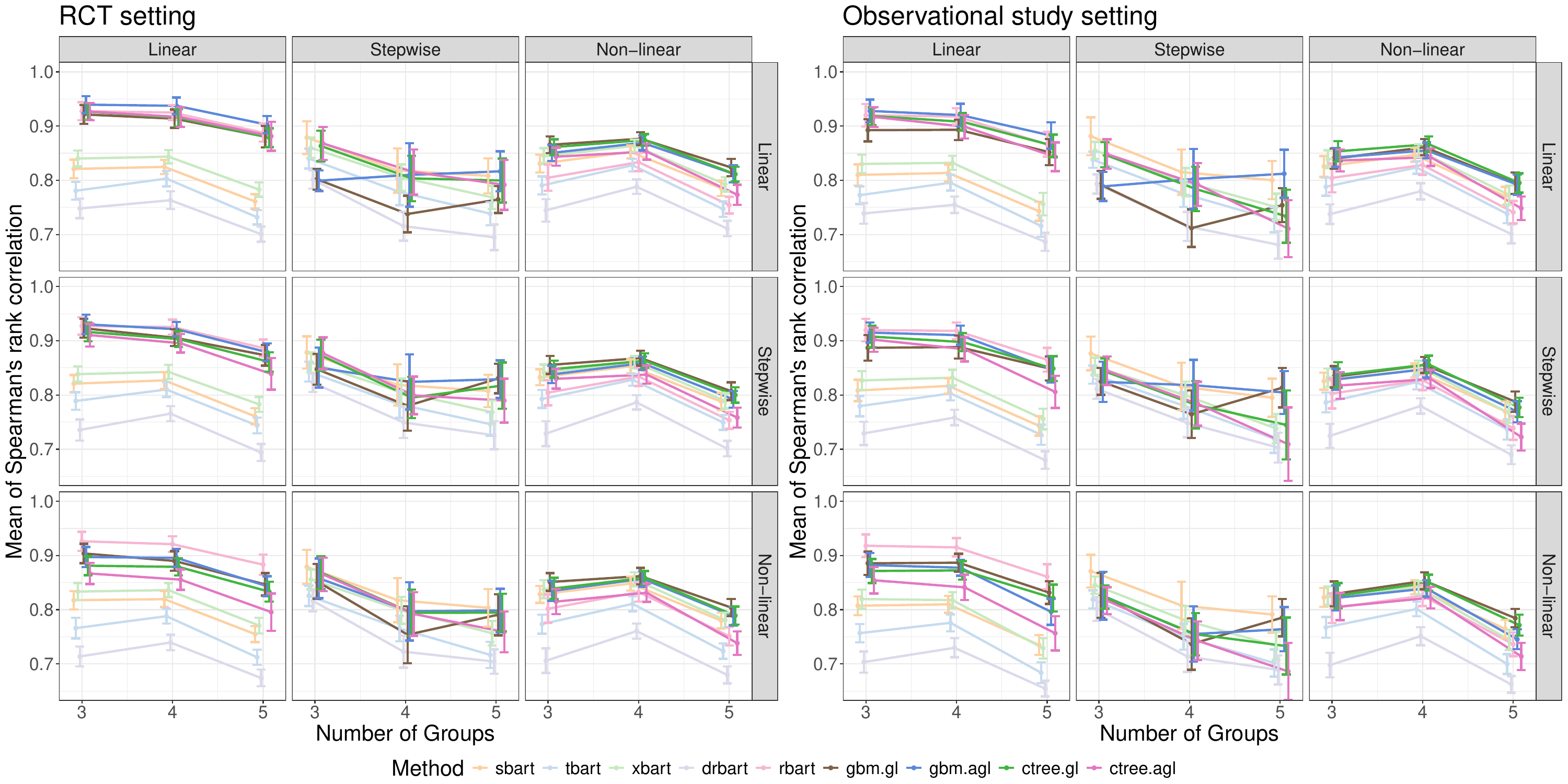}
\caption{Results in terms of mean Spearman’s rank correlation across twelve scenarios for all approaches are presented. The plots in the first column illustrate the performance of meta-learners using BART under both RCT (rct) and observational (obv) settings, whereas the plots in the second column display the results of the proposed approach in the same settings. In each plot, the x-axis represents the number of treatment groups while the y-axis denotes the Spearman’s rank correlation. These line charts depict the mean Spearman’s rank correlation along with their standard deviations, represented by error bars.}
\label{mord_fig}
\end{figure}

Figure \ref{mord_fig} compares meta-learners using BART and the proposed approaches in both RCT and observational settings in terms of the average Spearman’s rank correlation, with higher values indicating a more correct ranking of treatment effects for each subject. Overall, the average Spearman's rank correlation results showed a similar trend to the average Cohen's kappa values, with most of the proposed approaches outperforming the meta-learners in most scenarios. However, when the treatment effect was generated from a stepwise function, the average Spearman's rank correlation of the meta-learners decreased as the number of groups increased. In this setting, the results of the proposed approaches tended to be unstable, and the correlation of the proposed approach using the ``gbm'' rule generation and group lasso (gbm.gl) was lower than those of the S-, T-, and X-learners; however, the proposed approach using the ``gbm'' rule generation and adaptive group lasso (gbm.agl) outperformed these meta-learners. In addition, the correlation values of the proposed approach using the ``ctree'' rule generation and adaptive group lasso decreased rapidly as the number of groups increased.

\begin{figure}[h]
\centering
\includegraphics[width=\linewidth]{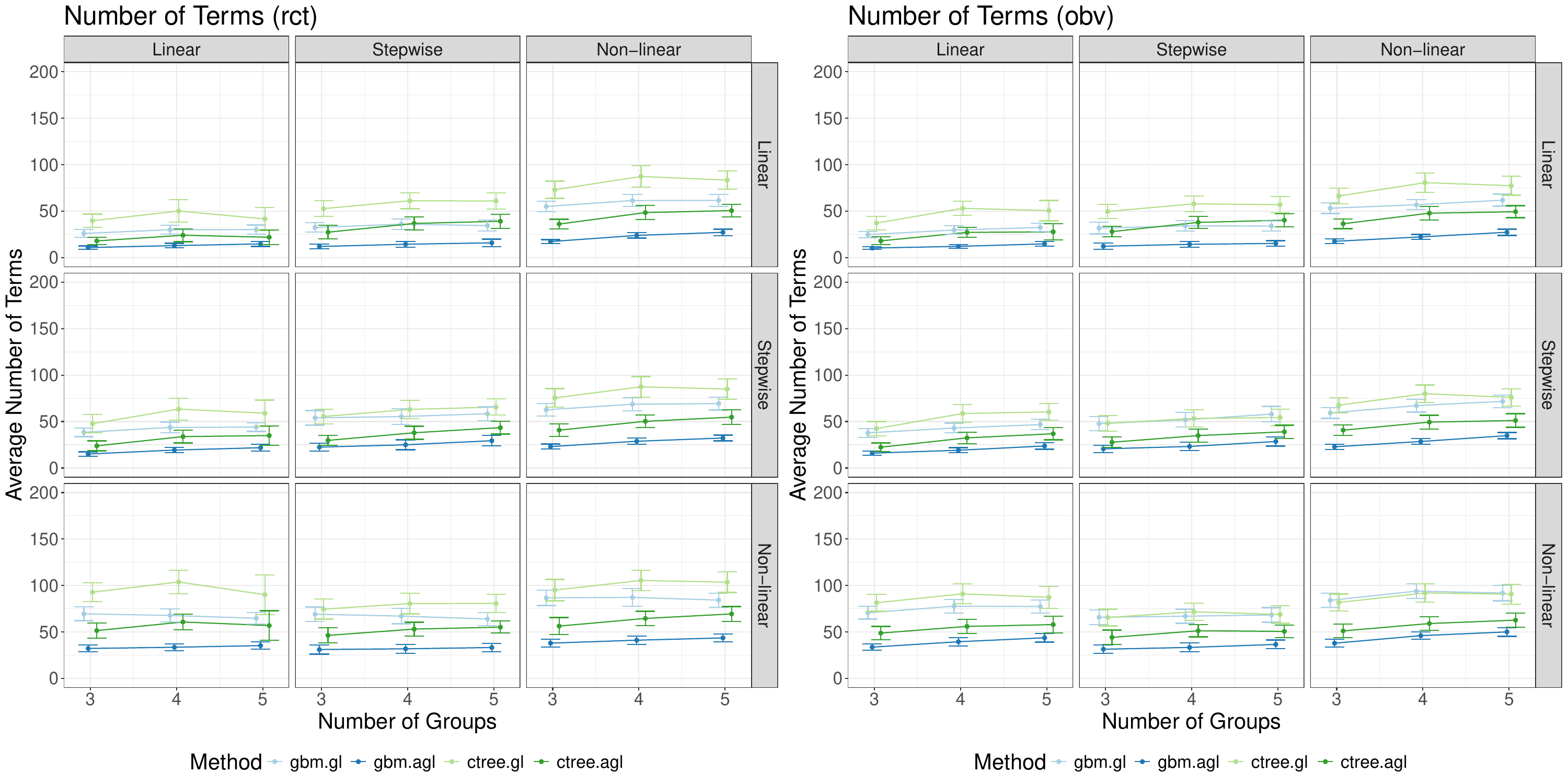}
\caption{Number of terms generated for different model generation process for the proposed approach. The plots in the first column illustrate the performance of meta-learners using BART under both RCT (rct) and observational (obv) settings, whereas the plots in the second column display the results of the proposed approach in the same settings. In each plot, the x-axis represents the number of treatment groups, while the y-axis denotes the number of terms within model. These line charts depict the mean the Spearman’s rank correlation values along with their standard deviations, represented by error bars.}
\label{rule_fig}
\end{figure}

Figure \ref{rule_fig} shows the average number of terms generated by the proposed approach in both RCT and observational settings. The results show that the complexity of the rule-based model, measured by the number of rules, varied across scenarios. In general, more basis functions were generated when the treatment effect was generated from a nonlinear function than when it was generated from a linear or stepwise function. Focusing on the rule generation process, the ``gbm'' approach resulted in fewer rules in the final model than the ``ctree'' approach. In terms of rule ensembles, the adaptive group lasso removed more basis functions from the model than the group lasso. Thus, considering the number of terms in the model, the ``gbm'' rule generation and the adaptive group lasso resulted in the fewest number of base functions in the model, while the ``ctree'' rule generation and the group lasso resulted in the largest number of base functions in the model.

\section{Real data application}\label{sec5}

We applied the proposed approachto a dataset from the AIDS Clinical Trials Group Protocol 175 (ACTG175), a randomized clinical trial involving HIV-1-infected adults with CD4 cell counts between 200 and 500 cells/$\mathrm{mm^3}$. This analysis demonstrates the performance of the proposed approachon real-world data. We first briefly present the ACTG175 dataset and then introduce the selection of the rule generation method, the rule ensemble method, and the tuning of the hyperparameters for the proposed approach on real data. Finally, we demonstrate how the proposed approach could be used to interpret the estimated HTE.

\subsection{Summary of the dataset}
The ACTG175 dataset comprises 1762 subjects across four treatment groups: zidovudine monotherapy (ZDV), zidovudine plus didanosine combination therapy (ZDV + DID), zidovudine plus zalcitabine combination therapy (ZDV + ZAL), and didanosine monotherapy (DID). For this study, we utilized 500 subjects treated with ZDV as the control group and 500 subjects each treated with ZDV + DID, ZDV + ZAL, and DID as the three respective treatment groups. The outcome variable was defined as changes in CD4 cell count at 20 weeks from baseline. CD4 cell count is a key metric for evaluating HIV progression, with a declining CD4 count indicating disease advancement. Therefore, a positive outcome value indicates that the patient has improved after treatment at 20 weeks, whereas a negative outcome value indicates that the patient's condition has worsened after treatment at 20 weeks. We selected 12 patient background covariates for our analysis, similar to a previous study\cite{Tsiatis2008}. These covariates included five continuous variables: baseline CD4 cell counts (cd40; cells/$\mathrm{mm^3}$), baseline CD8 cell count (cd80; cells/$\mathrm{mm^3}$), age (years), weight (wtkg;kg), and Karnofsky score (karnof; on a scale of 0-100). We also included seven binary variables: hemophilia (hemo; 0 = no, 1 = yes), homosexual activity (homo; 0 = no, 1 = yes), race (0 = white, 1 = other), sex (0 = female, 1 = male), history of intravenous drug use (drugs: 0 = no, 1 = yes), history of antiretroviral therapy (str2; 0 = naive, 1 = experienced), and symptomatic indicators (symptoms: 0 = asymptomatic, 1 = symptomatic).

\subsection{Model evaluation and parameter tuning}
As introduced in Section 3, the proposed approach allows the use of different rule generation and rule ensemble methods. In the simulation studies, we have discussed the prediction performance achieved using ``gbm'' and ``ctree'' rule generation and using group lasso and adaptive group lasso for rule ensemble. The results indicate that the HTE estimation performance of the proposed approaches is based on a combination of rule generation and rule ensemble methods. In addition, the proposed approach has several hyperparameters that need to be tuned. Therefore, before applying our proposed approach to the data, we must first determine the rule generation and rule ensemble methods and the hyperparameters. For this purpose, we need to evaluate the prediction performance for each setting. However, unlike in the simulation studies, we did not know the true HTEs for each subject and group in the real data. Therefore, it was impossible to directly compare the estimated HTE with the true HTE. We present a graphical evaluation method to evaluate the estimated HTE. We then describe the process of conducting such graphical evaluations in detail and outline the results that indicate a well-performed model. We then consider a metric based on these graphical evaluations to numerically evaluate the performance of the built model.

\begin{figure}[h]
    \centering
    \includegraphics[width=0.75\linewidth]{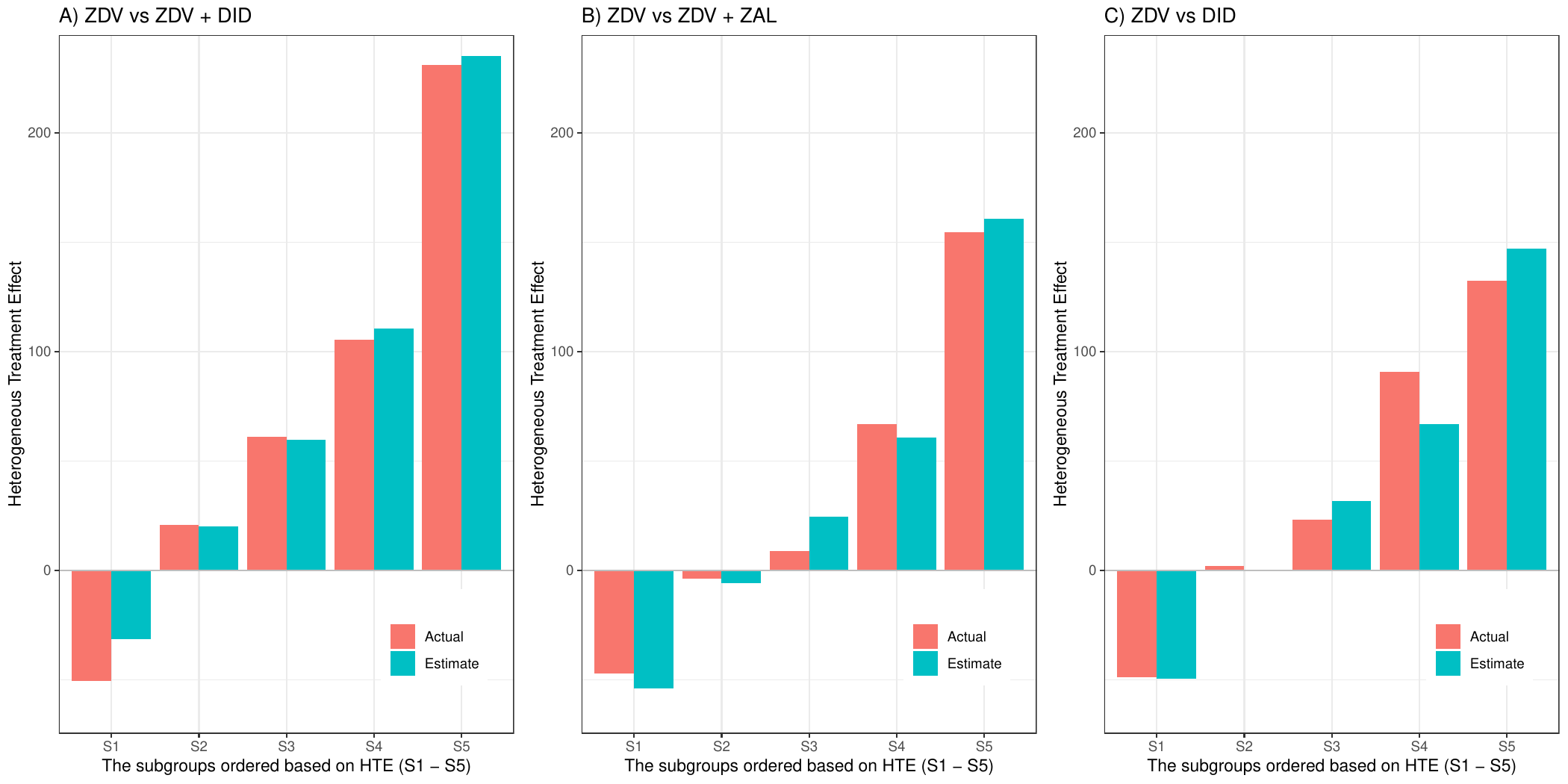}
    \caption{Graphical evaluation of the estimated HTE. The y-axis represents the HTE of each subgroup. The x-axis represents the name of the subgroups. The red bar denotes the actual HTE calculated as the mean difference between treatment and control within subgroups; the blue bar denotes the estimated HTE calculated as the mean of the estimated HTE within subgroups.}
    \label{agment}
\end{figure}

The detailed steps of graphical evaluations are as follows. First, the results of the subjects were ordered based on the estimated HTEs for each treatment group. Second, for each treatment group, the ordered subjects were divided into several subgroups, and the actual and estimated HTEs were calculated for each subgroup. Here, the actual HTE for a subgroup was calculated as the average treatment effect within the subgroup, whereas the estimated HTE for the subgroup was calculated as the average estimated HTE for the subject within the subgroup. If HTE is estimated properly, the actual and estimated HTEs will show the same trend, the estimated HTE will be similar to actual HTE, and the actual HTE and estimated HTE will both be positive or negative for each subgroup. Figure \ref{agment} shows the graphical evaluation of the proposed approach with tuned parameters used in this real data application, and we also use it as an example to explain how the estimated HTE was evaluated. 
The results of the graphical evaluation indicate similar trends between the actual and estimated HTEs. Moreover, the difference in values between the actual and estimated HTEs was also small. Therefore, we confirm that the HTE estimation model is well built.  

%良い結果を出すために必要な条件を追加, **ワッサーシュタイン距離（Wasserstein Distance）
However, model selection and parameter tuning require a numerical evaluation metric; therefore, we used the Spearman's rank correlation to numerically evaluate the trend between the estimated and actual HTEs and mean absolute error to numerically evaluate the difference between the estimated and actual HTEs, and calculated the percentage of agreement between the signs of estimated and actual HTEs. We then combined these three metrics and considered a novel metric for parameter tuning. Here, we define the actual HTE for $t$-th group and $s$-th subgroups (red bar in Figure\ref{agment}) as $\Delta_s^{(t)\mathrm{actual}}$ and the corresponding estimated HTE (blue bar in Figure\ref{agment}) as $\hat{\Delta}_s^{(t)}$, where $t\in\{$ ZDV vs ZDV + DID, ZDV vs ZDV + ZAL, ZDV vs DID$\}$ and $s\in S =\{$S1, S2, S3, S4, S5$\}$. The metric for the $t$-th group is defined as 
\begin{align}
\frac{1}{|Cor(\Delta^{(t) \mathrm{actural}}, \hat{\Delta}^{(t)})|}\left\{\frac{1}{|S|}\sum_{s \in S}\frac{|\Delta_s^{(t) \mathrm{actural}} - \hat{\Delta}_s^{(t)}|}{I(sign(\Delta_s^{(t) \mathrm{actural}}) = sign(\hat{\Delta}_s^{(t)}))}\right\}\label{tune_metric}
\end{align}
where $Cor()$ is the Spearman's rank correlation function. Therefore, when HTEs are perfectly estimated, this metric is equal to 0. With increasing estimation error and mismatch between the effectiveness of the actual and estimated HTE, the value of the metric tends to be larger. For model selection and parameter tuning, we randomly divided the data into two, one for training the model and the other for validating the model using the metric. We considered the ``gbm'' and ``ctree'' processes in rule generation, group lasso, and adaptive group lasso in the rule ensemble. Regarding the hyperparameters, we considered the number of trees $=\{333, 666, 1000\}$, depth of each tree $=\{2, 3, 4\}$, and shrinkage rate in boosting $=\{0.1, 0.01, 0.001\}$. Therefore, we had a total of $108$ patterns and used grid search to find the best combination of these patterns. The results are shown in Appendix\ref{apd2}. Finally, we established that using the ``gbm'' process for rule generation and adaptive group lasso for rule ensemble with hyperparameters---number of trees $= 666$, depth of each tree $= 3$, and shrinkage rate $= 0.001$---are the best combinations. We used these settings to apply the proposed approach to analyze the real dataset.

\subsection{Application results of the proposed approach}
Here, we present the results of applying the proposed method in three parts. First, we show how the estimated HTEs are interpreted; second, we explain how the HTEs based on the constructed model are interpreted; and third, we interpret the results using interpretation tools such as variable importance and make a brief comparison with conventional approaches to interpretation. In this application, the treatment effect of ZDV + DID, ZDV + ZAL, and DID is defined as the difference between their outcomes and that of ZDV. Furthermore, in multi-arm settings, the comparison of treatment effects between treatment groups is also important. Therefore, we also considered a treatment comparison between ZDV + ZAL and DID in this application. For simplicity, we denote these four groups as ``ZDV vs ZDV + DID,'' ``ZDV vs ZDV + ZAL,'' ``ZDV vs DID,'' and ``ZDV + ZAL vs DID.''  

\subsubsection{Interpretation based on estimated HTE}
\begin{figure}[h]
    \centering
    \includegraphics[width=0.75\linewidth]{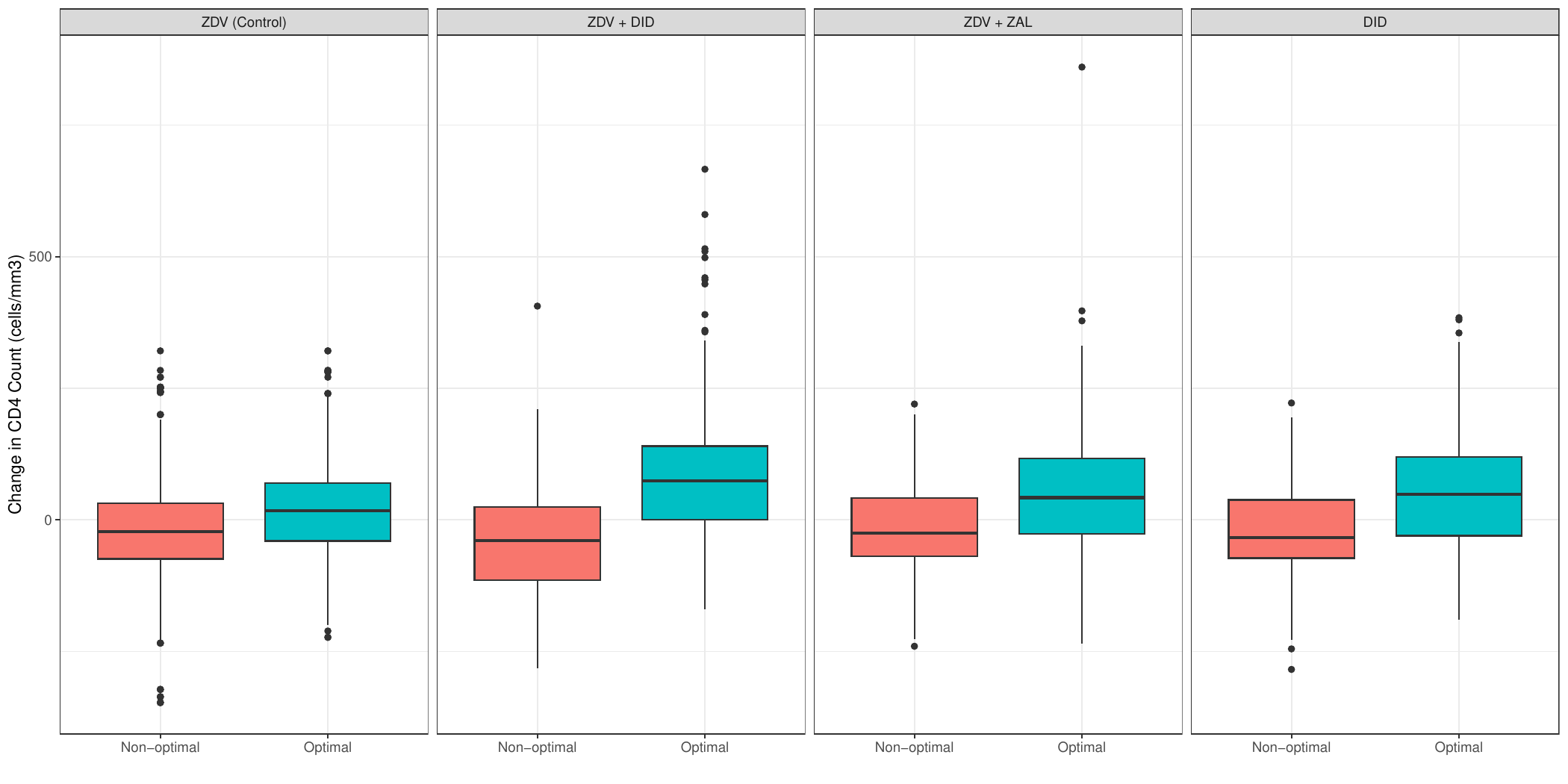}
    \caption{Comparison of differences in outcomes between those who received optimal treatment and those who did not. The y-axis represents change in CD4 counts (outcomes), and the x-axis shows whether the subjects received the optimal treatment.}
     \label{comp_opt}
\end{figure}

First, we demonstrate how to interpret the estimated HTE. For clarity, we selected three subjects as examples to present the estimated HTE for ZDV + DID, ZDV + ZAL, DID, and the results of the treatment effect comparison between ZDV + ZAL and DID are presented in Table\ref{Table1}.  
\begin{table}[b]
\centering
\caption{Estimated HTEs for five selected subjects (ZDV is control)}
\label{Table1}
\begin{tabular}{c|cccc}
\hline
ID    & ZDV vs ZDV + DID & ZDV vs ZDV + ZAL & ZDV vs DID  & ZDV + ZAL vs DID \\ \hline
10198 & -181.2      & -68.9      & -2.6  & 66.3\\
10368 & 403.7     & 151.1      & 0.2  & -150.9\\
50663 & 34.5     & -66.6     & -35.6 & 30.9\\ \hline
\end{tabular}
\end{table}
In the table, the first column shows subject ID; the second through fourth columns show the estimated HTE for ZDV + DID, ZDV + ZAL, and DID, respectively; and the fifth column shows the comparison of treatment effects for ZDV + ZAL vs. DID. For the first 10198 subjects, the estimated HTEs for ZDV + DID, ZDV + ZAL, and DID were all negative, indicating that ZDV was preferable for these subjects. In particular, the value for ZDV + DID was the smallest, indicating that this treatment was the least recommended. For the second 10368 subjects, the estimated HTE for ZDV + DID was the largest, indicating that this treatment was the most recommended. For the third 50663 subjects, the estimated HTE for ZDV + DID was the largest, indicating that this treatment was the most recommended; however, the estimated HTE for ZDV + DID was 0.2, which is almost equal to 0 compared to the other treatments. Therefore, we can conclude that ZDV and DID were equally effective for this subject. The estimated HTE for ZDV + ZAL was the smallest, indicating that this treatment was the least recommended. To compare ZDV + ZAL and DID, the difference between the estimated HTE of ZDV + ZAL and DID was calculated, with positive values indicating that DID was better and negative values indicating that ZDV + ZAL was better. Therefore, we can use estimated HTE to select the optimal treatment for each subject. In this way, we can also divide the data into two groups: subjects who received the recommended treatment and subjects who did not receive the recommended treatment. Here, we compare the outcome based on whether the treatment received was the optimal, as shown in Figure\ref{comp_opt}.  
The results indicate that for both treatments, subjects who received the optimal treatment tended to have positive CD4 count changes, whereas subjects who did not receive the optimal treatment tended to have negative CD4 count changes. Therefore, the subjects who received the recommended treatment using the estimated HTE of the proposed approach tended to do better than those who did not. Therefore, we empirically confirm that the estimated HTEs of the proposed approaches can be used to select the optimal treatment for a subject. 

\subsubsection{Interpretation based on estimated Model}
Second, we demonstrate ways to interpret the HTEs solely based on the developed HTE estimation model. In this application, our proposed approach generated 70 rules. For each rule, we provide its importance, support, and corresponding coefficient vector. Specifically, the importance of a rule reflects its contribution to the HTEs for different treatments, with larger importance values indicating that the rule plays a more important role in explaining the heterogeneity of treatment effects. The support represents the proportion of subjects that satisfy the rule, whereas the coefficient vector shows the direction and magnitude of the rule’s effect on the estimated HTEs. This detailed rule-level analysis allows us to identify and interpret the key covariates and interactions that drive treatment heterogeneity. Although all rules generated by the model are interpretable, not all of them may be useful in real-world practice. For example, some rules may have very low support, meaning they apply to only a small number of subjects, making them less generalizable or reliable for interpretation. Additionally, certain rules may have low importance, indicating that they contribute minimally to the heterogeneity of treatment effects and may not provide representative or actionable insights. To get a generalizable and representative interpretation, we selected the top five most important rules with support greater than 0.1 for each treatment comparison, as illustrated in Table \ref{rule_tab}. The details of the interpretations are as follows:

\paragraph{\textbf{ZDV vs ZDV + DID}} The top five importance rules with support greater than 0.1 are important. The first rule indicates that subjects weighing between 81.76 and 86.93 kg show a strong positive treatment effect, indicating that ZDV + DID is advantageous. The second rule suggests that subjects weighing 56.02 kg or more, with no antiretroviral history and no hemophilia, would benefit more from ZDV. The third rule indicates that treatment is also effective in Caucasian subjects over 24.5 years of age with a baseline CD8 cell count of less than 1238, favoring ZDV + DID. The fourth rule indicates that subjects with no antiretroviral history, baseline CD4 cell counts below 412, and CD8 cell counts above 834.5 show favorable treatment response in favor of ZDV + DID. Finally, the fifth rule reveals that subjects with baseline CD4 cell counts below 392.5 and CD8 cell counts below 1142 tend to have a negative treatment effect, suggesting that ZDV + DID may be less effective in this subgroup. 

\paragraph{\textbf{ZDV vs ZDV + ZAL}} The first and most important rule is that subjects weighing 81.76–86.93 kg show a strong treatment response, indicating that ZDV + ZAL is advantageous. The second rule indicates that the ZDV may be superior to ZDV + ZAL for subjects with baseline CD4 cell counts lower than 412, baseline CD8 cell counts higher than 507, and no antiretroviral history. Furthermore, the third rule indicates that ZDV + ZAL may be less effective than ZDV if weight is greater than 52.7 and no history of homosexual activity. The fourth rule indicates that ZDV + ZAL is preferred for subjects who are between 30 and 40 years of age, have a baseline CD4 cell count less than 335, and a CD8 cell count less than 2409. The last rule indicates that ZDV tends to be superior to ZDV + ZAL in subjects weighing 56.02 or more, with no antiretroviral history, and no hemophilia.

\paragraph{\textbf{ZDV vs DID}} The first rule indicates that ZDV tends to be superior to DID if the subject weighs 52.7 or more and has no history of homosexuality. The second rule indicates that DID is better than ZDV for subjects with baseline CD8 cell counts of 1070 or higher. The third rule indicates that subjects with a weight of 52.46 or greater, a baseline CD8 cell count of less than 1216, and a baseline CD4 cell count of less than 443 tend to do better with DID than with ZDV. However, the fourth rule shows that subjects with baseline CD4 cell counts less than 392 and baseline CD8 cell counts less than 1142 tend to do better with ZDV than with DID. The third and fourth rules have similar baseline CD4 and CD8 ranges; however, the third rule has a weight interaction; thus, weight may be important for the effect of ZDV and DID. The last rule indicates that DID is preferred if the subject is older than 24 years, the baseline CD4 cell count is less than 306, and the karnofsky score is greater than 95. 

\paragraph{\textbf{ZDV + ZAL vs DID}} The first rule indicates that DID is superior to ZDV + ZAL for subjects who are at least 24 years of age, have a baseline CD4 cell count less than 479, and a baseline CD8 cell count less than 848. The second rule indicates that ZDV + ZAL is superior to DID in subjects with baseline CD4 cell counts less than 412, CD8 cell counts greater than 507, and no antiretroviral history. The third rule indicates that DID tends to be superior to ZDV + ZAL if the subject weighs 56.02 or more, has no antiretroviral history, and has no hemophilia. The fourth rule indicates that subjects with baseline CD4 cell counts less than 392 and CD8 cell counts less than 1142 tend to have ZDV + ZAL superior to DID. The fifth rule indicates that ZDV + ZAL tends to be superior to DID in subjects with baseline CD8 cell counts greater than 1075.  

According to these results, baseline CD4 cell count, baseline CD8 cell count, and body weight are the most important variables affecting the priority of the therapeutic efficacy of these treatments. The importance of these variables will be further confirmed based on the importance of the variables in the next section. 

\begin{table}[tb]
\centering
\caption{Top five most important rules with support greater than 0.1 for each treatment comparison}
\label{rule_tab}
\vspace{0.2cm}
\begin{tabular}{lccc}
\multicolumn{4}{l}{\textbf{A) ZDV vs ZDV + DID}}                                                                                                                                    \\ \hline
\multicolumn{1}{l|}{\textbf{Rules}}                                                                           & \textbf{Importance}  & \textbf{Coefficients} & \textbf{Support}     \\ \hline
\multicolumn{1}{l|}{wtkg\textgreater{}=81.76 \& wtkg\textless 86.93}                                          & 47                   & 50.7                  & 0.12                 \\
\multicolumn{1}{l|}{hemo\textless 0.5 \& wtkg\textgreater{}=56.02 \& str2\textless 0.5}                       & 47                   & -33.4                 & 0.37                 \\
\multicolumn{1}{l|}{cd80\textless 1238 \& race\textless 0.5 \& age\textgreater{}=24.5}                        & 46                   & 32.1                  & 0.51                 \\
\multicolumn{1}{l|}{cd40\textless 412.5 \&cd80\textgreater{}=834.5 \& str2\textless 0.5}                      & 46                   & 43.1                  & 0.16                 \\
\multicolumn{1}{l|}{cd40\textless 392.5 \& cd80\textless 1142}                                                & 46                   & -31.8                 & 0.53                 \\ \hline
                                                                                                              &                      &                       &                      \\
\multicolumn{4}{l}{\textbf{B) ZDV vs ZDV + ZAL}}                                                                                                                                    \\ \hline
\multicolumn{1}{l|}{\textbf{Rules}}                                                                           & \textbf{Importance}  & \textbf{Coefficients} & \textbf{Support}     \\ \hline
\multicolumn{1}{l|}{wtkg\textgreater{}=81.76 \&   wtkg\textless 86.93}                                        & 52                   & 59.3                  & 0.12                 \\
\multicolumn{1}{l|}{cd40\textless 412.5 \& cd80\textgreater{}=507 \& str2\textless 0.5}                       & 48                   & 47                    & 0.16                 \\
\multicolumn{1}{l|}{wtkg\textgreater{}=52.73 \& homo\textless 0.5}                                            & 42                   & -33.2                 & 0.31                 \\
\multicolumn{1}{l|}{cd80\textless 2409 \& age\textless 40.5 \&   age\textgreater{}=29.5 \& cd40\textless 335} & 39                   & 33.3                  & 0.24                 \\
\multicolumn{1}{l|}{hemo\textless 0.5 \& wtkg\textgreater{}=56.02 \&   str2\textless 0.5}                     & 37                   & -27.6                 & 0.37                 \\ \hline
                                                                                                              & \multicolumn{1}{l}{} & \multicolumn{1}{l}{}  & \multicolumn{1}{l}{} \\
\multicolumn{4}{l}{\textbf{C) ZDV vs DID}}                                                                                                                                    \\ \hline
\multicolumn{1}{l|}{\textbf{Rules}}                                                                           & \textbf{Importance}  & \textbf{Coefficients} & \textbf{Support}     \\ \hline
\multicolumn{1}{l|}{wtkg\textgreater{}=52.73 \&   homo\textless 0.5}                                          & 57                   & -36.6                 & 0.31                 \\
\multicolumn{1}{l|}{cd80\textgreater{}=1070}                                                                  & 55                   & 35.1                  & 0.34                 \\
\multicolumn{1}{l|}{wtkg\textgreater{}=52.46 \& cd80\textless 1216 \& cd40\textless 443}                      & 48                   & 30.5                  & 0.66                 \\
\multicolumn{1}{l|}{cd40\textless 392.5 \&   cd80\textless 1142}                                              & 44                   & -26.6                 & 0.53                 \\
\multicolumn{1}{l|}{age\textgreater{}=23.5 \& cd40\textless 306.5 \& karnof\textgreater{}=95}                 & 41                   & 31.2                  & 0.20                 \\ \hline
                                                                                                              & \multicolumn{1}{l}{} & \multicolumn{1}{l}{}  & \multicolumn{1}{l}{} \\
\multicolumn{4}{l}{\textbf{D) ZDV + ZAL vs DID}}                                                                                                                                    \\ \hline
\multicolumn{1}{l|}{\textbf{Rules}}                                                                           & \textbf{Importance}  & \textbf{Coefficients} & \textbf{Support}     \\ \hline
\multicolumn{1}{l|}{age\textgreater{}=24.5 \&   cd40\textless 479 \& cd80\textless 848}                       & 75                   & 42.8                  & 0.40                 \\
\multicolumn{1}{l|}{cd40\textless 412.5 \& cd80\textgreater{}=507 \& str2\textless 0.5}                       & 61                   & -46.4                 & 0.16                 \\
\multicolumn{1}{l|}{hemo\textless 0.5 \& wtkg\textgreater{}=56.02 \&   str2\textless 0.5}                     & 53                   & 30.7                  & 0.37                 \\
\multicolumn{1}{l|}{cd40\textless 392.5 \&   cd80\textless 1142}                                              & 50                   & -28.1                 & 0.53                 \\
\multicolumn{1}{l|}{cd80\textgreater{}=1075}                                                                  & 46                   & -27.1                 & 0.33                 \\ \hline
\end{tabular}
\end{table}

\subsubsection{Interpretation based on variable importance}
\begin{figure*}[tb]
\centering
\subfloat[Variable importance of the proposed approach]{\includegraphics[clip, width=\linewidth]{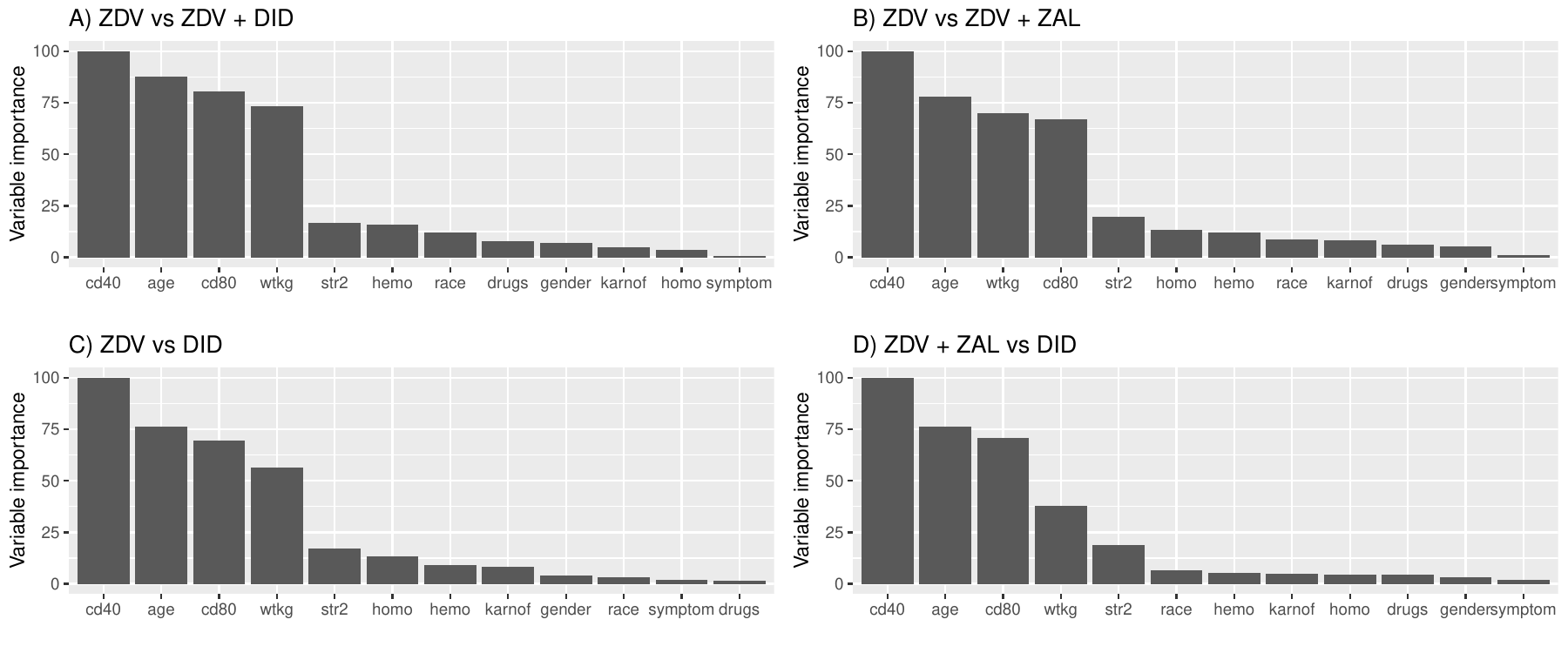}
\label{fig:label-A}}
\\
\subfloat[Variable importance of causal forest]{\includegraphics[clip, width=\linewidth]{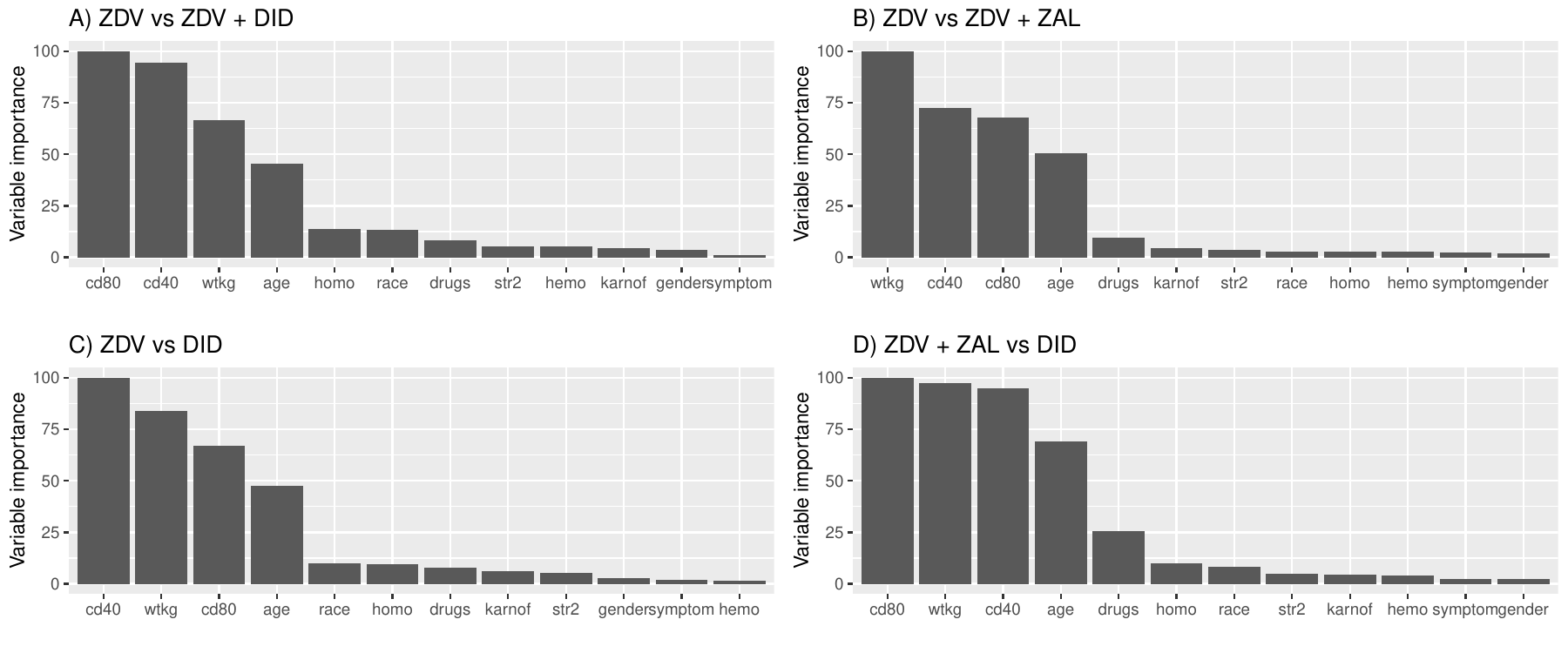}
\label{fig:label-B}}
\caption{Variable importance of the proposed approach and causal forest.}
\label{vip_res}
\end{figure*}

Finally, the interpretation of the results of the proposed approach is presented based on variable importance, a commonly used interpretation method. Here, we compared the interpretation results of the causal forest with the interpretation of the proposed approach. Causal forests cannot simultaneously provide variable importance for all four groups. Therefore, we calculated variable importance based on causal forest for ZDV vs. ZDV + DID, ZDV vs. ZDV + ZAL, ZDV vs. DID, and ZDV + ZAL vs. ZDV + DID. The variable importance of the proposed approach and the causal forest are shown in Figure \ref{vip_res}. Although there are some differences between the proposed approach and the causal forest, baseline CD4 cell count, baseline CD8 cell count, weight, and age are the most important variables in both the proposed approach and the causal forest. These variables are shown to be considerably important than the others, which implies that these variables contribute most to HTE. In other words, these background variables have the most influence on the selection of the optimal treatment. Here, it is also confirmed that the proposed method and the causal forests yield the same most important variables; therefore, the proposed method provides reasonable results.

\section{Conclusion and discussion}
In this study, we proposed a novel approach for creating interpretable rule-based HTE estimation models. The proposed approach exhibits a key advantage over conventional HTE estimation approaches for multi-arm data, such as meta-learners, in that it is interpretable while maintaining comparable prediction accuracy. The model constructed using the proposed approach takes the form of an additive model with rules and linear terms as base functions, facilitating the interpretation of causal relationships between covariates and HTE. Furthermore, in the proposed framework, we enforce each group to share the same basis function. Consequently, we can compare the treatment effect between any pair of treatment groups and interpret the differences in treatment effect between them based on the base functions. Thus, the proposed approach can be used to explore the treatment that is optimal for subjects with respective background characteristics. Additionally, because our approach builds an additional model that follows a structure similar to linear regression, it can be interpreted in a familiar manner by medical researchers and practitioners. 

We evaluated the estimation performance of the proposed approach through numerical simulations, comparing it with several commonly used meta-learners. In a previous study, Acharki et al.\cite{Acharki2023} employed mPEHE to assess the prediction accuracy of meta-learners using XGBoost and random forest for multi-arm HTE estimation. Their results indicated that the S- and X-learners achieved high prediction accuracy, whereas the M-learners performed poorly. However, their simulations were limited to linear models in RCT settings and hazard rate models in observational study settings. In contrast, our simulation study considers a broader range of simulation scenarios, including linear, stepwise, and nonlinear treatment and main effect functions, in both RCT and observational settings. Furthermore, we extend the evaluation by including meta-learners using BART, which has demonstrated strong predictive performance in recent literature. To provide a more comprehensive assessment, we evaluate each method not only in terms of mPEHE, but also based on estimation bias, the accuracy of optimal treatment selection, and the correctness of treatment effect rankings. Consistent with the findings of Acharki et al.\cite{Acharki2023}, the S- and X-learners demonstrated the highest prediction accuracy among the meta-learners in most settings. Notably, the best-performing variant of the proposed approach tend to outperformed the S- and X-learners when the treatment effect was generated from linear or stepwise functions, but showed slightly lower accuracy when the treatment effect was nonlinear. In terms of bias, both the S- and X-learners and the best-performing proposed method exhibited very low levels. Therefore, the slightly inferior prediction accuracy of the proposed approach in nonlinear scenarios may be attributed to higher variance in the estimates. This can be explained by the fact that the proposed approach is designed to prioritize interpretability through an additive model structure composed of rule-based and linear terms. While this structure enhances transparency, it may constrain the model’s ability to capture complex nonlinear treatment effect heterogeneity. In contrast, the S- and X-learners are not subject to such structural constraints and can leverage more flexible ensemble methods, such as BART, to better accommodate nonlinear relationships. Focusing on the correctness of optimal treatment selection, the proposed approach achieved Cohen's kappa values exceeding 0.8 in most scenarios. Similarly, in evaluating the correctness of treatment effect ranking, the proposed method attained Spearman’s rank correlations above 0.75 across nearly all settings, outperforming most of the meta-learners in both metrics. Therefore, we thought our proposed approaches well balanced the trade-off between the prediction accuracy and interpretability. Moreover, different combinations of rule generation and ensemble methods exhibited distinct performance characteristics. The use of "ctree" rule generation with group lasso demonstrated superior performance in scenarios where the treatment effect was generated from nonlinear functions, highlighting its strength in capturing complex interaction structures. In contrast, the combination of "gbm" rule generation with adaptive group lasso tended to perform best when the treatment effects followed linear or stepwise patterns, thus, the choice of rule generation and ensemble methods should be considered in practical use. 

The proposed approach was also applied to real data from the HIV study ACTG 175. To use the proposed approach in real data applications, we must first determine the rule generation method, rule ensemble method, and hyperparameters. However, on real datasets, it is difficult to directly evaluate the performance of a model because the true HTE is unknown. Therefore, we considered a metric for evaluating the performance of the proposed approach and used it to determine the rule generation method, rule ensemble method, and hyperparameters. Thereafter, we focused on the interpretation of the proposed approach. First, we provided an interpretation based on the estimated HTE. Furthermore, based on the estimated HTE of each group, we divided the subjects into those who received the optimal treatment and those who did not, confirming that the subjects who received the optimal treatment tended to show better outcomes than those who did not. Additionally, we confirmed that the estimated HTE of the proposed approach can be used to select the optimal treatment for the subjects. We then elaborated ways to interpret the relationship between the covariates and estimated HTE using the constructed model. Finally, we interpreted the results using an interpretation tool, variable importance, and compared the interpretation of the results of the proposed approach to that of an existing method, causal forest, and confirmed that the interpretation based on the proposed approach is reasonable.

%Bibliography
\bibliographystyle{unsrt}  
\bibliography{references}  

\clearpage
\appendix % 付録部分の始まり
\section{Appendix 1: Simulation Results of meta-learners using BART and proposed approaches}\label{apd1}

\begin{landscape}
% \usepackage{multirow}
\begin{table}[]
\caption{Results of mPEHE across twelve scenarios for meta-learners using BART and proposed approaches in RCT setting}\label{mPEHE.rct.bart}
\vspace{1cm}
\small
\begin{tabular}{c|c|cc|cc|cc|cc|cc|cc|cc|cc|cc}
\multirow{2}{*}{Group} & \multirow{2}{*}{Methods} & \multicolumn{2}{c|}{L-L}      & \multicolumn{2}{c|}{L-S}      & \multicolumn{2}{c|}{L-N}      & \multicolumn{2}{c|}{S-L}      & \multicolumn{2}{c|}{S-S}      & \multicolumn{2}{c|}{S-N}      & \multicolumn{2}{c|}{N-L}      & \multicolumn{2}{c|}{N-S}      & \multicolumn{2}{c}{N-N}       \\ \cline{3-20} 
                       &                          & mean          & sd            & mean          & sd            & mean          & sd            & mean          & sd            & mean          & sd            & mean          & sd            & mean          & sd            & mean          & sd            & mean          & sd            \\ \hline
                       & sbart                    & 0.47          & 0.03          & \textbf{0.45} & \textbf{0.06} & \textbf{0.43} & \textbf{0.02} & 0.47          & 0.03          & \textbf{0.44} & \textbf{0.06} & \textbf{0.43} & \textbf{0.03} & 0.48          & 0.03          & \textbf{0.45} & \textbf{0.06} & \textbf{0.43} & \textbf{0.02} \\
                       & tbart                    & 0.57          & 0.03          & 0.64          & 0.04          & 0.55          & 0.03          & 0.56          & 0.03          & 0.62          & 0.04          & 0.56          & 0.03          & 0.62          & 0.04          & 0.70          & 0.04          & 0.61          & 0.03          \\
                       & xbart                    & 0.41          & 0.03          & 0.50          & 0.03          & \textbf{0.41} & \textbf{0.03} & 0.42          & 0.03          & 0.50          & 0.03          & \textbf{0.41} & \textbf{0.03} & 0.43          & 0.03          & 0.52          & 0.04          & \textbf{0.42} & \textbf{0.03} \\
                       & mbart                    & 1.76          & 0.18          & 1.70          & 0.16          & 1.64          & 0.17          & 1.61          & 0.15          & 1.46          & 0.14          & 1.56          & 0.15          & 1.78          & 0.19          & 1.65          & 0.16          & 1.67          & 0.17          \\
                       & drbart                   & 0.68          & 0.03          & 0.69          & 0.03          & 0.64          & 0.02          & 0.70          & 0.03          & 0.64          & 0.03          & 0.66          & 0.03          & 0.74          & 0.02          & 0.72          & 0.03          & 0.72          & 0.03          \\
3                      & rfbart                   & 0.34          & 0.04          & 1.50          & 0.02          & 0.72          & 0.03          & 0.33          & 0.04          & 1.50          & 0.02          & 0.72          & 0.03          & 0.36          & 0.04          & 1.50          & 0.02          & 0.73          & 0.03          \\
                       & rbart                    & 0.30          & 0.04          & 1.49          & 0.02          & 0.71          & 0.03          & 0.28          & 0.04          & 1.49          & 0.02          & 0.71          & 0.03          & \textbf{0.32} & \textbf{0.04} & 1.49          & 0.02          & 0.71          & 0.03          \\
                       & gbm.gl                   & \textbf{0.19} & \textbf{0.03} & 0.56          & 0.08          & 0.52          & 0.03          & \textbf{0.22} & \textbf{0.03} & 0.47          & 0.08          & 0.51          & 0.03          & \textbf{0.29} & \textbf{0.03} & 0.54          & 0.11          & 0.49          & 0.03          \\
                       & gbm.agl                  & \textbf{0.22} & \textbf{0.04} & 0.50          & 0.10          & 0.56          & 0.04          & \textbf{0.22} & \textbf{0.04} & 0.45          & 0.09          & 0.57          & 0.04          & 0.34          & 0.05          & 0.54          & 0.12          & 0.54          & 0.04          \\
                       & ctree.gl                 & \textbf{0.20} & \textbf{0.03} & \textbf{0.40} & \textbf{0.06} & \textbf{0.44} & \textbf{0.02} & \textbf{0.23} & \textbf{0.04} & \textbf{0.39} & \textbf{0.06} & \textbf{0.43} & \textbf{0.03} & \textbf{0.33} & \textbf{0.04} & \textbf{0.44} & \textbf{0.07} & \textbf{0.45} & \textbf{0.03} \\
                       & ctree.agl                & 0.25          & 0.04          & \textbf{0.40} & \textbf{0.08} & 0.48          & 0.03          & 0.28          & 0.05          & \textbf{0.41} & \textbf{0.07} & 0.49          & 0.04          & 0.41          & 0.05          & \textbf{0.49} & \textbf{0.08} & 0.52          & 0.04          \\ \hline
                       & sbart                    & 0.55          & 0.03          & 0.56          & 0.07          & \textbf{0.49} & \textbf{0.03} & 0.55          & 0.03          & 0.56          & 0.07          & \textbf{0.49} & \textbf{0.03} & 0.56          & 0.03          & \textbf{0.56} & \textbf{0.08} & \textbf{0.50} & \textbf{0.03} \\
                       & tbart                    & 0.63          & 0.04          & 0.70          & 0.04          & 0.60          & 0.03          & 0.60          & 0.03          & 0.68          & 0.04          & 0.60          & 0.03          & 0.67          & 0.03          & 0.75          & 0.04          & 0.67          & 0.04          \\
                       & xbart                    & 0.48          & 0.03          & 0.58          & 0.04          & \textbf{0.47} & \textbf{0.03} & 0.48          & 0.03          & 0.58          & 0.04          & \textbf{0.48} & \textbf{0.03} & 0.50          & 0.03          & 0.60          & 0.04          & \textbf{0.49} & \textbf{0.03} \\
                       & mbart                    & 2.06          & 0.20          & 2.01          & 0.17          & 1.91          & 0.20          & 1.85          & 0.21          & 1.72          & 0.15          & 1.79          & 0.20          & 2.13          & 0.23          & 1.96          & 0.17          & 1.99          & 0.21          \\
                       & drbart                   & 0.81          & 0.03          & 0.77          & 0.03          & 0.75          & 0.02          & 0.79          & 0.02          & 0.72          & 0.03          & 0.74          & 0.02          & 0.82          & 0.02          & 0.80          & 0.03          & 0.80          & 0.03          \\
4                      & rfbart                   & 0.48          & 0.05          & 1.77          & 0.02          & 0.90          & 0.04          & 0.47          & 0.05          & 1.77          & 0.02          & 0.89          & 0.04          & 0.50          & 0.05          & 1.77          & 0.02          & 0.90          & 0.04          \\
                       & rbart                    & 0.35          & 0.05          & 1.74          & 0.02          & 0.84          & 0.03          & 0.33          & 0.05          & 1.74          & 0.02          & 0.83          & 0.03          & \textbf{0.38} & \textbf{0.05} & 1.75          & 0.02          & 0.85          & 0.03          \\
                       & gbm.gl                   & \textbf{0.26} & \textbf{0.03} & 0.60          & 0.07          & 0.61          & 0.04          & \textbf{0.29} & \textbf{0.03} & \textbf{0.51} & \textbf{0.09} & 0.60          & 0.04          & \textbf{0.36} & \textbf{0.04} & 0.57          & 0.09          & 0.58          & 0.04          \\
                       & gbm.agl                  & \textbf{0.27} & \textbf{0.06} & \textbf{0.46} & \textbf{0.10} & 0.60          & 0.05          & \textbf{0.29} & \textbf{0.04} & \textbf{0.46} & \textbf{0.11} & 0.62          & 0.05          & \textbf{0.41} & \textbf{0.06} & \textbf{0.53} & \textbf{0.10} & 0.61          & 0.05          \\
                       & ctree.gl                 & \textbf{0.26} & \textbf{0.04} & \textbf{0.46} & \textbf{0.08} & \textbf{0.52} & \textbf{0.03} & \textbf{0.29} & \textbf{0.04} & \textbf{0.48} & \textbf{0.08} & \textbf{0.52} & \textbf{0.03} & 0.39          & 0.05          & \textbf{0.52} & \textbf{0.07} & \textbf{0.52} & \textbf{0.03} \\
                       & ctree.agl                & 0.33          & 0.06          & \textbf{0.48} & \textbf{0.10} & 0.58          & 0.04          & 0.38          & 0.06          & 0.53          & 0.09          & 0.60          & 0.04          & 0.52          & 0.06          & 0.61          & 0.08          & 0.63          & 0.05          \\ \hline
                       & sbart                    & 0.59          & 0.03          & 0.59          & 0.07          & \textbf{0.53} & \textbf{0.03} & 0.59          & 0.03          & 0.59          & 0.06          & \textbf{0.52} & \textbf{0.03} & 0.60          & 0.03          & \textbf{0.61} & \textbf{0.07} & \textbf{0.54} & \textbf{0.03} \\
                       & tbart                    & 0.68          & 0.04          & 0.73          & 0.04          & 0.65          & 0.04          & 0.64          & 0.03          & 0.71          & 0.04          & 0.64          & 0.04          & 0.72          & 0.04          & 0.79          & 0.04          & 0.71          & 0.04          \\
                       & xbart                    & 0.52          & 0.03          & 0.61          & 0.03          & \textbf{0.51} & \textbf{0.03} & 0.52          & 0.03          & 0.61          & 0.04          & \textbf{0.52} & \textbf{0.03} & 0.55          & 0.03          & 0.64          & 0.04          & \textbf{0.54} & \textbf{0.03} \\
                       & mbart                    & 2.45          & 0.26          & 2.38          & 0.19          & 2.24          & 0.20          & 2.16          & 0.19          & 2.01          & 0.15          & 2.10          & 0.16          & 2.52          & 0.26          & 2.31          & 0.21          & 2.35          & 0.22          \\
                       & drbart                   & 0.83          & 0.03          & 0.82          & 0.03          & 0.77          & 0.03          & 0.82          & 0.03          & 0.78          & 0.03          & 0.78          & 0.03          & 0.85          & 0.03          & 0.86          & 0.04          & 0.85          & 0.03          \\
5                      & rfbart                   & 0.54          & 0.06          & 1.76          & 0.02          & 0.92          & 0.04          & 0.53          & 0.06          & 1.76          & 0.02          & 0.92          & 0.04          & 0.55          & 0.06          & 1.76          & 0.02          & 0.93          & 0.04          \\
                       & rbart                    & 0.37          & 0.06          & 1.73          & 0.02          & 0.84          & 0.04          & 0.36          & 0.06          & 1.72          & 0.02          & 0.84          & 0.03          & 0.40          & 0.07          & 1.73          & 0.02          & 0.85          & 0.04          \\
                       & gbm.gl                   & \textbf{0.30} & \textbf{0.04} & 0.65          & 0.07          & 0.63          & 0.04          & \textbf{0.33} & \textbf{0.04} & \textbf{0.54} & \textbf{0.07} & 0.63          & 0.03          & \textbf{0.42} & \textbf{0.07} & 0.64          & 0.09          & 0.63          & 0.04          \\
                       & gbm.agl                  & \textbf{0.32} & \textbf{0.06} & \textbf{0.49} & \textbf{0.10} & 0.61          & 0.05          & \textbf{0.34} & \textbf{0.06} & \textbf{0.51} & \textbf{0.10} & 0.64          & 0.05          & \textbf{0.48} & \textbf{0.10} & \textbf{0.60} & \textbf{0.10} & 0.65          & 0.06          \\
                       & ctree.gl                 & \textbf{0.29} & \textbf{0.04} & \textbf{0.47} & \textbf{0.08} & \textbf{0.55} & \textbf{0.03} & \textbf{0.33} & \textbf{0.03} & \textbf{0.47} & \textbf{0.07} & \textbf{0.55} & \textbf{0.03} & \textbf{0.43} & \textbf{0.05} & \textbf{0.52} & \textbf{0.07} & \textbf{0.56} & \textbf{0.04} \\
                       & ctree.agl                & 0.35          & 0.07          & \textbf{0.51} & \textbf{0.10} & 0.64          & 0.04          & 0.44          & 0.08          & 0.55          & 0.09          & 0.66          & 0.05          & 0.59          & 0.10          & 0.65          & 0.09          & 0.72          & 0.06          \\ \hline
\end{tabular}
\vspace{0.1cm}
\\\textbf{note}: Bolded text means the best three results for each scenario. The first column indicates the number of groups, the second column indicates the evaluation method, and the third to last columns are the results of the different simulation scenarios. Each scenario is named “main effect function - treatment effect function” based on the corresponding data generating function, where “L” is a linear function, “S” is a step-wise function and “N” is a nonlinear function.
\end{table}
\end{landscape}

\begin{landscape}
\begin{table}[]
\caption{Results of mPEHE across twelve scenarios for meta-learners using BART and proposed approaches in observational study setting}\label{mPEHE.obv.bart}
\vspace{1cm}
\small
\begin{tabular}{c|c|ll|ll|ll|ll|ll|ll|ll|ll|ll}
\multirow{2}{*}{Group} & \multirow{2}{*}{Methods} & \multicolumn{2}{c|}{L-L}                           & \multicolumn{2}{c|}{L-S}                           & \multicolumn{2}{c|}{L-N}                           & \multicolumn{2}{c|}{S-L}                           & \multicolumn{2}{c|}{S-S}                           & \multicolumn{2}{c|}{S-N}                           & \multicolumn{2}{c|}{N-L}                           & \multicolumn{2}{c|}{N-S}                           & \multicolumn{2}{c}{N-N}                           \\ \cline{3-20} 
                       &                          & \multicolumn{1}{c}{mean} & \multicolumn{1}{c|}{sd} & \multicolumn{1}{c}{mean} & \multicolumn{1}{c|}{sd} & \multicolumn{1}{c}{mean} & \multicolumn{1}{c|}{sd} & \multicolumn{1}{c}{mean} & \multicolumn{1}{c|}{sd} & \multicolumn{1}{c}{mean} & \multicolumn{1}{c|}{sd} & \multicolumn{1}{c}{mean} & \multicolumn{1}{c|}{sd} & \multicolumn{1}{c}{mean} & \multicolumn{1}{c|}{sd} & \multicolumn{1}{c}{mean} & \multicolumn{1}{c|}{sd} & \multicolumn{1}{c}{mean} & \multicolumn{1}{c}{sd} \\ \hline
                       & sbart                    & 0.50                     & 0.03                    & \textbf{0.44}            & \textbf{0.06}           & \textbf{0.44}            & \textbf{0.03}           & 0.50                     & 0.03                    & \textbf{0.45}            & \textbf{0.05}           & \textbf{0.44}            & \textbf{0.03}           & 0.51                     & 0.03                    & \textbf{0.46}            & \textbf{0.06}           & \textbf{0.45}            & \textbf{0.03}          \\
                       & tbart                    & 0.60                     & 0.03                    & 0.64                     & 0.04                    & 0.55                     & 0.03                    & 0.58                     & 0.03                    & 0.63                     & 0.04                    & 0.58                     & 0.03                    & 0.65                     & 0.03                    & 0.71                     & 0.04                    & 0.63                     & 0.03                   \\
                       & xbart                    & 0.44                     & 0.03                    & 0.51                     & 0.04                    & \textbf{0.42}            & \textbf{0.03}           & 0.45                     & 0.03                    & 0.52                     & 0.04                    & \textbf{0.44}            & \textbf{0.03}           & 0.47                     & 0.03                    & \textbf{0.54}            & \textbf{0.04}           & \textbf{0.45}            & \textbf{0.03}          \\
                       & mbart                    & 3.95                     & 1.30                    & 3.28                     & 0.92                    & 3.10                     & 0.74                    & 3.19                     & 1.07                    & 2.55                     & 0.59                    & 2.56                     & 0.53                    & 3.56                     & 1.05                    & 2.92                     & 0.68                    & 2.90                     & 0.63                   \\
                       & drbart                   & 0.72                     & 0.04                    & 0.71                     & 0.03                    & 0.66                     & 0.03                    & 0.73                     & 0.04                    & 0.67                     & 0.03                    & 0.69                     & 0.04                    & 0.77                     & 0.05                    & 0.75                     & 0.04                    & 0.74                     & 0.03                   \\
3                      & rfbart                   & 0.98                     & 0.10                    & 1.84                     & 0.08                    & 1.16                     & 0.09                    & 0.99                     & 0.10                    & 1.84                     & 0.08                    & 1.16                     & 0.09                    & 0.98                     & 0.10                    & 1.83                     & 0.07                    & 1.15                     & 0.09                   \\
                       & rbart                    & 0.32                     & 0.05                    & 1.49                     & 0.02                    & 0.71                     & 0.03                    & 0.30                     & 0.05                    & 1.49                     & 0.02                    & 0.71                     & 0.02                    & \textbf{0.33}            & \textbf{0.05}           & 1.50                     & 0.02                    & 0.72                     & 0.02                   \\
                       & gbm.gl                   & \textbf{0.23}            & \textbf{0.03}           & 0.62                     & 0.09                    & 0.55                     & 0.03                    & \textbf{0.28}            & \textbf{0.04}           & 0.52                     & 0.08                    & 0.54                     & 0.04                    & \textbf{0.31}            & \textbf{0.04}           & 0.59                     & 0.11                    & 0.51                     & 0.03                   \\
                       & gbm.agl                  & \textbf{0.22}            & \textbf{0.06}           & 0.56                     & 0.12                    & 0.58                     & 0.05                    & \textbf{0.25}            & \textbf{0.05}           & 0.51                     & 0.10                    & 0.59                     & 0.05                    & 0.39                     & 0.07                    & 0.59                     & 0.13                    & 0.58                     & 0.06                   \\
                       & ctree.gl                 & \textbf{0.22}            & \textbf{0.03}           & \textbf{0.46}            & \textbf{0.07}           & \textbf{0.46}            & \textbf{0.03}           & \textbf{0.25}            & \textbf{0.04}           & \textbf{0.48}            & \textbf{0.07}           & \textbf{0.47}            & \textbf{0.04}           & \textbf{0.34}            & \textbf{0.05}           & \textbf{0.53}            & \textbf{0.07}           & \textbf{0.48}            & \textbf{0.04}          \\
                       & ctree.agl                & 0.27                     & 0.05                    & \textbf{0.48}            & \textbf{0.08}           & 0.51                     & 0.04                    & 0.31                     & 0.06                    & \textbf{0.52}            & \textbf{0.07}           & 0.54                     & 0.05                    & 0.44                     & 0.06                    & 0.59                     & 0.08                    & 0.57                     & 0.05                   \\ \hline
                       & sbart                    & 0.58                     & 0.03                    & \textbf{0.56}            & \textbf{0.07}           & \textbf{0.50}            & \textbf{0.03}           & 0.58                     & 0.03                    & \textbf{0.55}            & \textbf{0.07}           & \textbf{0.51}            & \textbf{0.03}           & 0.59                     & 0.03                    & \textbf{0.56}            & \textbf{0.07}           & \textbf{0.51}            & \textbf{0.03}          \\
                       & tbart                    & 0.64                     & 0.03                    & 0.69                     & 0.04                    & 0.60                     & 0.03                    & 0.62                     & 0.03                    & 0.68                     & 0.04                    & 0.61                     & 0.03                    & 0.71                     & 0.04                    & 0.78                     & 0.04                    & 0.69                     & 0.03                   \\
                       & xbart                    & 0.50                     & 0.02                    & 0.59                     & 0.04                    & \textbf{0.48}            & \textbf{0.03}           & 0.51                     & 0.03                    & 0.60                     & 0.04                    & \textbf{0.50}            & \textbf{0.03}           & 0.54                     & 0.03                    & 0.63                     & 0.04                    & \textbf{0.53}            & \textbf{0.03}          \\
                       & mbart                    & 3.63                     & 0.94                    & 3.25                     & 0.70                    & 2.90                     & 0.59                    & 2.95                     & 0.69                    & 2.53                     & 0.42                    & 2.54                     & 0.36                    & 3.38                     & 0.92                    & 2.89                     & 0.57                    & 2.88                     & 0.62                   \\
                       & drbart                   & 0.82                     & 0.04                    & 0.78                     & 0.04                    & 0.76                     & 0.03                    & 0.81                     & 0.03                    & 0.74                     & 0.04                    & 0.76                     & 0.04                    & 0.84                     & 0.04                    & 0.82                     & 0.04                    & 0.83                     & 0.04                   \\
4                      & rfbart                   & 1.26                     & 0.11                    & 2.14                     & 0.08                    & 1.45                     & 0.10                    & 1.27                     & 0.12                    & 2.15                     & 0.08                    & 1.45                     & 0.10                    & 1.26                     & 0.11                    & 2.14                     & 0.08                    & 1.45                     & 0.10                   \\
                       & rbart                    & 0.37                     & 0.05                    & 1.75                     & 0.02                    & 0.84                     & 0.03                    & 0.35                     & 0.05                    & 1.75                     & 0.02                    & 0.84                     & 0.03                    & \textbf{0.39}            & \textbf{0.06}           & 1.75                     & 0.02                    & 0.85                     & 0.03                   \\
                       & gbm.gl                   & \textbf{0.30}            & \textbf{0.04}           & 0.65                     & 0.07                    & 0.65                     & 0.04                    & \textbf{0.34}            & \textbf{0.04}           & \textbf{0.52}            & \textbf{0.06}           & 0.63                     & 0.04                    & \textbf{0.37}            & \textbf{0.04}           & 0.60                     & 0.08                    & 0.59                     & 0.04                   \\
                       & gbm.agl                  & \textbf{0.29}            & \textbf{0.06}           & \textbf{0.48}            & \textbf{0.10}           & 0.66                     & 0.05                    & \textbf{0.31}            & \textbf{0.05}           & \textbf{0.45}            & \textbf{0.08}           & 0.66                     & 0.05                    & 0.48                     & 0.06                    & \textbf{0.58}            & \textbf{0.10}           & 0.66                     & 0.06                   \\
                       & ctree.gl                 & \textbf{0.27}            & \textbf{0.04}           & \textbf{0.54}            & \textbf{0.07}           & \textbf{0.54}            & \textbf{0.04}           & \textbf{0.31}            & \textbf{0.03}           & 0.56                     & 0.08                    & \textbf{0.54}            & \textbf{0.04}           & \textbf{0.39}            & \textbf{0.04}           & \textbf{0.60}            & \textbf{0.07}           & \textbf{0.55}            & \textbf{0.04}          \\
                       & ctree.agl                & 0.38                     & 0.06                    & 0.58                     & 0.09                    & 0.61                     & 0.04                    & 0.42                     & 0.06                    & 0.63                     & 0.10                    & 0.64                     & 0.04                    & 0.55                     & 0.07                    & 0.71                     & 0.09                    & 0.68                     & 0.06                   \\ \hline
                       & sbart                    & 0.61                     & 0.03                    & 0.59                     & 0.07                    & \textbf{0.53}            & \textbf{0.03}           & 0.61                     & 0.03                    & 0.59                     & 0.06                    & \textbf{0.53}            & \textbf{0.03}           & 0.62                     & 0.04                    & \textbf{0.61}            & \textbf{0.06}           & \textbf{0.54}            & \textbf{0.03}          \\
                       & tbart                    & 0.68                     & 0.04                    & 0.73                     & 0.04                    & 0.63                     & 0.03                    & 0.65                     & 0.03                    & 0.72                     & 0.04                    & 0.64                     & 0.03                    & 0.77                     & 0.04                    & 0.83                     & 0.05                    & 0.74                     & 0.04                   \\
                       & xbart                    & 0.55                     & 0.03                    & 0.63                     & 0.04                    & \textbf{0.51}            & \textbf{0.03}           & 0.55                     & 0.03                    & 0.64                     & 0.04                    & \textbf{0.53}            & \textbf{0.03}           & 0.61                     & 0.04                    & 0.69                     & 0.04                    & \textbf{0.59}            & \textbf{0.03}          \\
                       & mbart                    & 3.92                     & 0.76                    & 3.52                     & 0.53                    & 3.15                     & 0.47                    & 3.11                     & 0.55                    & 2.66                     & 0.33                    & 2.71                     & 0.35                    & 3.72                     & 0.62                    & 3.23                     & 0.46                    & 3.24                     & 0.48                   \\
                       & drbart                   & 0.84                     & 0.03                    & 0.84                     & 0.03                    & 0.78                     & 0.02                    & 0.84                     & 0.03                    & 0.81                     & 0.03                    & 0.80                     & 0.03                    & 0.88                     & 0.03                    & 0.89                     & 0.05                    & 0.87                     & 0.03                   \\
5                      & rfbart                   & 1.28                     & 0.12                    & 2.10                     & 0.08                    & 1.46                     & 0.10                    & 1.29                     & 0.12                    & 2.11                     & 0.08                    & 1.46                     & 0.11                    & 1.28                     & 0.12                    & 2.10                     & 0.08                    & 1.46                     & 0.10                   \\
                       & rbart                    & 0.42                     & 0.05                    & 1.73                     & 0.02                    & 0.85                     & 0.03                    & 0.39                     & 0.05                    & 1.73                     & 0.02                    & 0.85                     & 0.03                    & 0.44                     & 0.05                    & 1.73                     & 0.02                    & 0.86                     & 0.03                   \\
                       & gbm.gl                   & \textbf{0.33}            & \textbf{0.04}           & 0.67                     & 0.07                    & 0.65                     & 0.03                    & \textbf{0.37}            & \textbf{0.04}           & \textbf{0.54}            & \textbf{0.06}           & 0.63                     & 0.03                    & \textbf{0.41}            & \textbf{0.04}           & 0.62                     & 0.07                    & 0.61                     & 0.04                   \\
                       & gbm.agl                  & \textbf{0.32}            & \textbf{0.06}           & \textbf{0.49}            & \textbf{0.10}           & 0.63                     & 0.05                    & \textbf{0.38}            & \textbf{0.06}           & \textbf{0.52}            & \textbf{0.09}           & 0.65                     & 0.05                    & \textbf{0.55}            & \textbf{0.06}           & \textbf{0.64}            & \textbf{0.10}           & 0.71                     & 0.06                   \\
                       & ctree.gl                 & \textbf{0.30}            & \textbf{0.04}           & \textbf{0.61}            & \textbf{0.10}           & \textbf{0.56}            & \textbf{0.03}           & \textbf{0.33}            & \textbf{0.04}           & \textbf{0.59}            & \textbf{0.13}           & \textbf{0.57}            & \textbf{0.03}           & \textbf{0.41}            & \textbf{0.05}           & \textbf{0.63}            & \textbf{0.12}           & \textbf{0.58}            & \textbf{0.04}          \\
                       & ctree.agl                & 0.42                     & 0.08                    & \textbf{0.69}            & \textbf{0.13}           & 0.66                     & 0.04                    & 0.48                     & 0.07                    & 0.70                     & 0.15                    & 0.71                     & 0.05                    & 0.62                     & 0.09                    & 0.79                     & 0.14                    & 0.75                     & 0.06                   \\ \hline
\end{tabular}
\vspace{0.1cm}
\\\textbf{note}: Bolded text means the best three results for each scenario. The first column indicates the number of groups, the second column indicates the evaluation method, and the third to last columns are the results of the different simulation scenarios. Each scenario is named “main effect function - treatment effect function” based on the corresponding data generating function, where “L” is a linear function, “S” is a step-wise function and “N” is a nonlinear function.
\end{table}
\end{landscape}

\begin{landscape}
\begin{table}[]
\caption{Results of mean absolute relative bias across twelve scenarios for meta-learners using BART and proposed approaches in RCT setting}\label{mbias.rct.bart}
\vspace{1cm}
\small
\begin{tabular}{c|c|ll|ll|ll|ll|ll|ll|ll|ll|ll}
\multirow{2}{*}{Group} & \multirow{2}{*}{Methods} & \multicolumn{2}{c|}{L-L}                           & \multicolumn{2}{c|}{L-S}                           & \multicolumn{2}{c|}{L-N}                           & \multicolumn{2}{c|}{S-L}                           & \multicolumn{2}{c|}{S-S}                           & \multicolumn{2}{c|}{S-N}                           & \multicolumn{2}{c|}{N-L}                           & \multicolumn{2}{c|}{N-S}                           & \multicolumn{2}{c}{N-N}                           \\ \cline{3-20} 
                       &                          & \multicolumn{1}{c}{mean} & \multicolumn{1}{c|}{sd} & \multicolumn{1}{c}{mean} & \multicolumn{1}{c|}{sd} & \multicolumn{1}{c}{mean} & \multicolumn{1}{c|}{sd} & \multicolumn{1}{c}{mean} & \multicolumn{1}{c|}{sd} & \multicolumn{1}{c}{mean} & \multicolumn{1}{c|}{sd} & \multicolumn{1}{c}{mean} & \multicolumn{1}{c|}{sd} & \multicolumn{1}{c}{mean} & \multicolumn{1}{c|}{sd} & \multicolumn{1}{c}{mean} & \multicolumn{1}{c|}{sd} & \multicolumn{1}{c}{mean} & \multicolumn{1}{c}{sd} \\ \hline
                       & sbart                    & 0.092                    & 0.051                   & \textbf{0.082}           & \textbf{0.050}          & \textbf{0.089}           & \textbf{0.057}          & 0.090                    & 0.050                   & \textbf{0.081}           & \textbf{0.048}          & 0.096                    & 0.049                   & \textbf{0.095}           & \textbf{0.050}          & \textbf{0.084}           & \textbf{0.051}          & \textbf{0.095}           & \textbf{0.055}         \\
                       & tbart                    & 0.090                    & 0.050                   & \textbf{0.085}           & \textbf{0.053}          & \textbf{0.087}           & \textbf{0.052}          & 0.090                    & 0.053                   & \textbf{0.086}           & \textbf{0.056}          & \textbf{0.089}           & \textbf{0.053}          & 0.099                    & 0.054                   & \textbf{0.096}           & \textbf{0.057}          & \textbf{0.100}           & \textbf{0.054}         \\
                       & xbart                    & 0.088                    & 0.051                   & \textbf{0.083}           & \textbf{0.053}          & \textbf{0.085}           & \textbf{0.052}          & \textbf{0.088}           & \textbf{0.052}          & \textbf{0.083}           & \textbf{0.054}          & \textbf{0.088}           & \textbf{0.052}          & \textbf{0.097}           & \textbf{0.052}          & \textbf{0.092}           & \textbf{0.054}          & \textbf{0.096}           & \textbf{0.053}         \\
                       & mbart                    & 0.126                    & 0.070                   & 0.152                    & 0.094                   & 0.128                    & 0.074                   & 0.124                    & 0.062                   & 0.156                    & 0.092                   & 0.134                    & 0.069                   & 0.158                    & 0.095                   & 0.174                    & 0.098                   & 0.164                    & 0.093                  \\
                       & drbart                   & 0.096                    & 0.049                   & 0.089                    & 0.055                   & 0.092                    & 0.050                   & 0.095                    & 0.049                   & 0.086                    & 0.056                   & \textbf{0.093}           & \textbf{0.050}          & \textbf{0.098}           & \textbf{0.052}          & 0.098                    & 0.061                   & 0.101                    & 0.053                  \\
3                      & rfbart                   & 0.125                    & 0.067                   & 0.149                    & 0.086                   & 0.132                    & 0.066                   & 0.122                    & 0.065                   & 0.149                    & 0.083                   & 0.128                    & 0.066                   & 0.128                    & 0.073                   & 0.154                    & 0.088                   & 0.132                    & 0.069                  \\
                       & rbart                    & 0.106                    & 0.067                   & 0.128                    & 0.078                   & 0.110                    & 0.067                   & 0.101                    & 0.065                   & 0.127                    & 0.074                   & 0.107                    & 0.063                   & 0.111                    & 0.073                   & 0.134                    & 0.081                   & 0.115                    & 0.068                  \\
                       & gbm.gl                   & \textbf{0.082}           & \textbf{0.050}          & 0.097                    & 0.057                   & 0.090                    & 0.049                   & 0.088                    & 0.046                   & 0.092                    & 0.052                   & 0.095                    & 0.052                   & 0.100                    & 0.054                   & 0.108                    & 0.060                   & 0.102                    & 0.059                  \\
                       & gbm.agl                  & 0.078                    & 0.044                   & 0.100                    & 0.053                   & 0.095                    & 0.048                   & \textbf{0.084}           & \textbf{0.047}          & 0.092                    & 0.057                   & 0.099                    & 0.055                   & 0.101                    & 0.055                   & 0.110                    & 0.065                   & 0.104                    & 0.059                  \\
                       & ctree.gl                 & \textbf{0.083}           & \textbf{0.049}          & 0.096                    & 0.057                   & 0.091                    & 0.054                   & \textbf{0.087}           & \textbf{0.049}          & 0.092                    & 0.059                   & 0.096                    & 0.056                   & 0.100                    & 0.055                   & 0.112                    & 0.065                   & 0.108                    & 0.060                  \\
                       & ctree.agl                & \textbf{0.086}           & \textbf{0.049}          & 0.098                    & 0.053                   & 0.096                    & 0.056                   & 0.090                    & 0.051                   & 0.092                    & 0.057                   & 0.101                    & 0.056                   & 0.102                    & 0.055                   & 0.113                    & 0.062                   & 0.110                    & 0.065                  \\ \hline
                       & sbart                    & 0.080                    & 0.045                   & 0.076                    & 0.038                   & 0.080                    & 0.041                   & 0.081                    & 0.044                   & \textbf{0.076}           & \textbf{0.039}          & \textbf{0.078}           & \textbf{0.046}          & 0.080                    & 0.045                   & \textbf{0.073}           & \textbf{0.036}          & \textbf{0.079}           & \textbf{0.046}         \\
                       & tbart                    & 0.082                    & 0.047                   & 0.080                    & 0.043                   & 0.081                    & 0.050                   & 0.082                    & 0.047                   & 0.079                    & 0.045                   & 0.085                    & 0.051                   & 0.083                    & 0.047                   & 0.082                    & 0.044                   & 0.085                    & 0.049                  \\
                       & xbart                    & 0.080                    & 0.044                   & 0.079                    & 0.041                   & 0.080                    & 0.047                   & 0.081                    & 0.045                   & 0.079                    & 0.043                   & 0.083                    & 0.049                   & 0.080                    & 0.045                   & \textbf{0.079}           & \textbf{0.042}          & 0.082                    & 0.046                  \\
                       & mbart                    & 0.115                    & 0.064                   & 0.136                    & 0.080                   & 0.120                    & 0.069                   & 0.119                    & 0.061                   & 0.140                    & 0.084                   & 0.127                    & 0.074                   & 0.143                    & 0.074                   & 0.160                    & 0.084                   & 0.143                    & 0.078                  \\
                       & drbart                   & 0.081                    & 0.045                   & 0.083                    & 0.045                   & 0.082                    & 0.046                   & 0.085                    & 0.049                   & 0.079                    & 0.043                   & 0.086                    & 0.049                   & 0.084                    & 0.044                   & 0.080                    & 0.043                   & 0.084                    & 0.043                  \\
4                      & rfbart                   & 0.149                    & 0.072                   & 0.150                    & 0.081                   & 0.147                    & 0.073                   & 0.151                    & 0.073                   & 0.150                    & 0.080                   & 0.146                    & 0.077                   & 0.154                    & 0.076                   & 0.153                    & 0.083                   & 0.151                    & 0.079                  \\
                       & rbart                    & 0.107                    & 0.048                   & 0.118                    & 0.052                   & 0.104                    & 0.052                   & 0.102                    & 0.050                   & 0.115                    & 0.052                   & 0.103                    & 0.054                   & 0.113                    & 0.049                   & 0.118                    & 0.058                   & 0.112                    & 0.056                  \\
                       & gbm.gl                   & \textbf{0.067}           & \textbf{0.037}          & 0.079                    & 0.040                   & \textbf{0.071}           & \textbf{0.042}          & \textbf{0.074}           & \textbf{0.041}          & \textbf{0.077}           & \textbf{0.041}          & \textbf{0.079}           & \textbf{0.042}          & \textbf{0.076}           & \textbf{0.043}          & \textbf{0.079}           & \textbf{0.044}          & \textbf{0.075}           & \textbf{0.046}         \\
                       & gbm.agl                  & \textbf{0.065}           & \textbf{0.037}          & \textbf{0.074}           & \textbf{0.039}          & \textbf{0.072}           & \textbf{0.043}          & \textbf{0.075}           & \textbf{0.042}          & \textbf{0.075}           & \textbf{0.046}          & \textbf{0.081}           & \textbf{0.047}          & 0.080                    & 0.043                   & 0.080                    & 0.042                   & \textbf{0.077}           & \textbf{0.044}         \\
                       & ctree.gl                 & 0.069                    & 0.038                   & \textbf{0.076}           & \textbf{0.045}          & \textbf{0.072}           & \textbf{0.046}          & \textbf{0.074}           & \textbf{0.042}          & 0.084                    & 0.056                   & 0.084                    & 0.047                   & \textbf{0.075}           & \textbf{0.044}          & 0.082                    & 0.046                   & 0.080                    & 0.050                  \\
                       & ctree.agl                & \textbf{0.070}           & \textbf{0.039}          & \textbf{0.075}           & \textbf{0.043}          & 0.075                    & 0.046                   & 0.076                    & 0.044                   & 0.083                    & 0.058                   & 0.086                    & 0.047                   & \textbf{0.076}           & \textbf{0.045}          & 0.083                    & 0.050                   & 0.085                    & 0.051                  \\ \hline
                       & sbart                    & 0.093                    & 0.051                   & 0.085                    & 0.043                   & 0.092                    & 0.043                   & 0.091                    & 0.047                   & 0.084                    & 0.042                   & 0.089                    & 0.043                   & \textbf{0.091}           & \textbf{0.042}          & \textbf{0.084}           & \textbf{0.042}          & \textbf{0.093}           & \textbf{0.043}         \\
                       & tbart                    & 0.091                    & 0.047                   & 0.086                    & 0.042                   & 0.089                    & 0.044                   & 0.090                    & 0.047                   & 0.085                    & 0.042                   & \textbf{0.087}           & \textbf{0.043}          & 0.096                    & 0.047                   & \textbf{0.092}           & \textbf{0.047}          & \textbf{0.095}           & \textbf{0.047}         \\
                       & xbart                    & 0.090                    & 0.047                   & 0.085                    & 0.043                   & 0.088                    & 0.043                   & 0.089                    & 0.047                   & 0.085                    & 0.042                   & \textbf{0.087}           & \textbf{0.044}          & 0.095                    & 0.046                   & \textbf{0.091}           & \textbf{0.046}          & \textbf{0.094}           & \textbf{0.045}         \\
                       & mbart                    & 0.122                    & 0.069                   & 0.154                    & 0.082                   & 0.132                    & 0.068                   & 0.122                    & 0.066                   & 0.154                    & 0.078                   & 0.131                    & 0.068                   & 0.148                    & 0.089                   & 0.176                    & 0.091                   & 0.158                    & 0.087                  \\
                       & drbart                   & 0.089                    & 0.048                   & 0.087                    & 0.042                   & \textbf{0.089}           & \textbf{0.045}          & 0.091                    & 0.047                   & 0.085                    & 0.041                   & 0.087                    & 0.044                   & 0.095                    & 0.047                   & 0.095                    & 0.044                   & 0.096                    & 0.047                  \\
5                      & rfbart                   & 0.143                    & 0.067                   & 0.142                    & 0.074                   & 0.147                    & 0.067                   & 0.143                    & 0.066                   & 0.138                    & 0.073                   & 0.148                    & 0.066                   & 0.149                    & 0.068                   & 0.147                    & 0.075                   & 0.154                    & 0.066                  \\
                       & rbart                    & 0.099                    & 0.045                   & 0.117                    & 0.058                   & 0.106                    & 0.047                   & 0.095                    & 0.045                   & 0.113                    & 0.058                   & 0.103                    & 0.046                   & 0.109                    & 0.048                   & 0.121                    & 0.061                   & 0.113                    & 0.049                  \\
                       & gbm.gl                   & \textbf{0.076}           & \textbf{0.037}          & 0.086                    & 0.040                   & \textbf{0.082}           & \textbf{0.039}          & \textbf{0.082}           & \textbf{0.042}          & \textbf{0.083}           & \textbf{0.042}          & \textbf{0.085}           & \textbf{0.039}          & \textbf{0.093}           & \textbf{0.040}          & 0.097                    & 0.048                   & 0.098                    & 0.042                  \\
                       & gbm.agl                  & \textbf{0.074}           & \textbf{0.036}          & \textbf{0.082}           & \textbf{0.039}          & \textbf{0.082}           & \textbf{0.040}          & \textbf{0.084}           & \textbf{0.045}          & \textbf{0.082}           & \textbf{0.040}          & 0.087                    & 0.040                   & 0.097                    & 0.044                   & 0.098                    & 0.044                   & 0.098                    & 0.046                  \\
                       & ctree.gl                 & \textbf{0.079}           & \textbf{0.039}          & \textbf{0.081}           & \textbf{0.041}          & \textbf{0.082}           & \textbf{0.036}          & \textbf{0.083}           & \textbf{0.041}          & \textbf{0.083}           & \textbf{0.039}          & 0.089                    & 0.041                   & \textbf{0.094}           & \textbf{0.046}          & 0.097                    & 0.045                   & 0.100                    & 0.044                  \\
                       & ctree.agl                & 0.080                    & 0.040                   & \textbf{0.084}           & \textbf{0.045}          & 0.087                    & 0.040                   & 0.088                    & 0.043                   & 0.087                    & 0.039                   & 0.092                    & 0.043                   & 0.101                    & 0.050                   & 0.104                    & 0.051                   & 0.104                    & 0.049                  \\ \hline
\end{tabular}
\vspace{0.1cm}
\\\textbf{note}: Bolded text means the best three results for each scenario. The first column indicates the number of groups, the second column indicates the evaluation method, and the third to last columns are the results of the different simulation scenarios. Each scenario is named “main effect function - treatment effect function” based on the corresponding data generating function, where “L” is a linear function, “S” is a step-wise function and “N” is a nonlinear function.
\end{table}
\end{landscape}

\begin{landscape}
\begin{table}[]
\caption{Results of mean absolute relative bias across twelve scenarios for meta-learners using BART and proposed approaches in observational study setting}\label{mbias.obv.bart}
\vspace{1cm}
\small
\begin{tabular}{c|c|ll|ll|ll|ll|ll|ll|ll|ll|ll}
\multirow{2}{*}{Group} & \multirow{2}{*}{Methods} & \multicolumn{2}{c|}{L-L}                           & \multicolumn{2}{c|}{L-S}                           & \multicolumn{2}{c|}{L-N}                           & \multicolumn{2}{c|}{S-L}                           & \multicolumn{2}{c|}{S-S}                           & \multicolumn{2}{c|}{S-N}                           & \multicolumn{2}{c|}{N-L}                           & \multicolumn{2}{c|}{N-S}                           & \multicolumn{2}{c}{N-N}                           \\ \cline{3-20} 
                       &                          & \multicolumn{1}{c}{mean} & \multicolumn{1}{c|}{sd} & \multicolumn{1}{c}{mean} & \multicolumn{1}{c|}{sd} & \multicolumn{1}{c}{mean} & \multicolumn{1}{c|}{sd} & \multicolumn{1}{c}{mean} & \multicolumn{1}{c|}{sd} & \multicolumn{1}{c}{mean} & \multicolumn{1}{c|}{sd} & \multicolumn{1}{c}{mean} & \multicolumn{1}{c|}{sd} & \multicolumn{1}{c}{mean} & \multicolumn{1}{c|}{sd} & \multicolumn{1}{c}{mean} & \multicolumn{1}{c|}{sd} & \multicolumn{1}{c}{mean} & \multicolumn{1}{c}{sd} \\ \hline
                       & sbart                    & 0.116                    & 0.066                   & \textbf{0.099}           & \textbf{0.057}          & 0.103                    & 0.065                   & 0.118                    & 0.067                   & \textbf{0.095}           & \textbf{0.058}          & \textbf{0.098}           & \textbf{0.064}          & 0.124                    & 0.069                   & \textbf{0.099}           & \textbf{0.060}          & \textbf{0.103}           & \textbf{0.063}         \\
                       & tbart                    & 0.113                    & 0.066                   & 0.102                    & 0.065                   & \textbf{0.098}           & \textbf{0.056}          & 0.109                    & 0.064                   & \textbf{0.101}           & \textbf{0.062}          & \textbf{0.098}           & \textbf{0.061}          & 0.123                    & 0.071                   & 0.108                    & 0.064                   & \textbf{0.103}           & \textbf{0.064}         \\
                       & xbart                    & 0.101                    & 0.064                   & \textbf{0.095}           & \textbf{0.059}          & \textbf{0.096}           & \textbf{0.055}          & \textbf{0.101}           & \textbf{0.062}          & \textbf{0.096}           & \textbf{0.061}          & \textbf{0.098}           & \textbf{0.060}          & \textbf{0.109}           & \textbf{0.067}          & \textbf{0.101}           & \textbf{0.060}          & \textbf{0.099}           & \textbf{0.058}         \\
                       & mbart                    & 0.270                    & 0.163                   & 0.277                    & 0.163                   & 0.248                    & 0.140                   & 0.251                    & 0.153                   & 0.265                    & 0.141                   & 0.238                    & 0.112                   & 0.277                    & 0.166                   & 0.266                    & 0.138                   & 0.251                    & 0.140                  \\
                       & drbart                   & \textbf{0.099}           & \textbf{0.059}          & \textbf{0.100}           & \textbf{0.055}          & \textbf{0.096}           & \textbf{0.057}          & 0.104                    & 0.059                   & 0.102                    & 0.058                   & 0.101                    & 0.061                   & \textbf{0.109}           & \textbf{0.059}          & \textbf{0.106}           & \textbf{0.063}          & 0.106                    & 0.058                  \\
3                      & rfbart                   & 0.468                    & 0.143                   & 0.569                    & 0.183                   & 0.459                    & 0.150                   & 0.489                    & 0.143                   & 0.579                    & 0.186                   & 0.467                    & 0.147                   & 0.459                    & 0.142                   & 0.554                    & 0.186                   & 0.447                    & 0.146                  \\
                       & rbart                    & 0.116                    & 0.052                   & 0.141                    & 0.081                   & 0.114                    & 0.062                   & 0.109                    & 0.054                   & 0.144                    & 0.087                   & 0.113                    & 0.064                   & 0.119                    & 0.057                   & 0.148                    & 0.086                   & 0.116                    & 0.060                  \\
                       & gbm.gl                   & 0.212                    & 0.088                   & 0.223                    & 0.092                   & 0.209                    & 0.085                   & 0.238                    & 0.091                   & 0.117                    & 0.075                   & 0.180                    & 0.091                   & 0.171                    & 0.090                   & 0.135                    & 0.081                   & 0.155                    & 0.086                  \\
                       & gbm.agl                  & \textbf{0.095}           & \textbf{0.059}          & 0.113                    & 0.065                   & 0.104                    & 0.064                   & \textbf{0.104}           & \textbf{0.060}          & 0.110                    & 0.063                   & 0.116                    & 0.054                   & \textbf{0.116}           & \textbf{0.068}          & 0.114                    & 0.073                   & 0.121                    & 0.069                  \\
                       & ctree.gl                 & 0.110                    & 0.066                   & 0.116                    & 0.079                   & 0.099                    & 0.059                   & 0.119                    & 0.068                   & 0.140                    & 0.083                   & 0.111                    & 0.062                   & 0.129                    & 0.070                   & 0.139                    & 0.084                   & 0.122                    & 0.071                  \\
                       & ctree.agl                & \textbf{0.093}           & \textbf{0.056}          & 0.122                    & 0.086                   & 0.101                    & 0.059                   & \textbf{0.103}           & \textbf{0.057}          & 0.141                    & 0.083                   & 0.114                    & 0.059                   & 0.120                    & 0.063                   & 0.138                    & 0.081                   & 0.124                    & 0.069                  \\ \hline
                       & sbart                    & 0.106                    & 0.063                   & \textbf{0.086}           & \textbf{0.045}          & 0.093                    & 0.046                   & 0.107                    & 0.059                   & \textbf{0.084}           & \textbf{0.041}          & \textbf{0.091}           & \textbf{0.049}          & 0.105                    & 0.058                   & \textbf{0.082}           & \textbf{0.044}          & \textbf{0.089}           & \textbf{0.044}         \\
                       & tbart                    & 0.106                    & 0.056                   & 0.090                    & 0.049                   & 0.090                    & 0.049                   & 0.101                    & 0.052                   & 0.088                    & 0.047                   & \textbf{0.090}           & \textbf{0.051}          & 0.111                    & 0.053                   & 0.091                    & 0.046                   & 0.095                    & 0.046                  \\
                       & xbart                    & 0.095                    & 0.050                   & \textbf{0.083}           & \textbf{0.045}          & \textbf{0.086}           & \textbf{0.046}          & \textbf{0.096}           & \textbf{0.049}          & \textbf{0.083}           & \textbf{0.045}          & \textbf{0.087}           & \textbf{0.050}          & 0.101                    & 0.048                   & \textbf{0.084}           & \textbf{0.042}          & \textbf{0.090}           & \textbf{0.044}         \\
                       & mbart                    & 0.204                    & 0.112                   & 0.210                    & 0.117                   & 0.183                    & 0.093                   & 0.192                    & 0.111                   & 0.195                    & 0.107                   & 0.167                    & 0.098                   & 0.214                    & 0.117                   & 0.206                    & 0.116                   & 0.194                    & 0.101                  \\
                       & drbart                   & \textbf{0.089}           & \textbf{0.047}          & \textbf{0.082}           & \textbf{0.047}          & \textbf{0.085}           & \textbf{0.046}          & \textbf{0.092}           & \textbf{0.046}          & 0.085                    & 0.047                   & 0.089                    & 0.052                   & \textbf{0.096}           & \textbf{0.050}          & \textbf{0.086}           & \textbf{0.045}          & \textbf{0.093}           & \textbf{0.043}         \\
4                      & rfbart                   & 0.503                    & 0.105                   & 0.421                    & 0.130                   & 0.478                    & 0.107                   & 0.512                    & 0.107                   & 0.430                    & 0.135                   & 0.480                    & 0.108                   & 0.500                    & 0.103                   & 0.419                    & 0.128                   & 0.470                    & 0.107                  \\
                       & rbart                    & 0.104                    & 0.046                   & 0.132                    & 0.059                   & 0.106                    & 0.043                   & 0.101                    & 0.045                   & 0.131                    & 0.058                   & 0.103                    & 0.043                   & 0.107                    & 0.045                   & 0.130                    & 0.056                   & 0.109                    & 0.045                  \\
                       & gbm.gl                   & 0.188                    & 0.072                   & 0.197                    & 0.074                   & 0.163                    & 0.068                   & 0.185                    & 0.077                   & 0.094                    & 0.048                   & 0.135                    & 0.062                   & 0.159                    & 0.072                   & 0.105                    & 0.056                   & 0.135                    & 0.066                  \\
                       & gbm.agl                  & \textbf{0.091}           & \textbf{0.054}          & 0.089                    & 0.049                   & 0.093                    & 0.048                   & \textbf{0.095}           & \textbf{0.046}          & \textbf{0.081}           & \textbf{0.044}          & 0.093                    & 0.051                   & \textbf{0.097}           & \textbf{0.052}          & 0.089                    & 0.048                   & 0.094                    & 0.053                  \\
                       & ctree.gl                 & 0.108                    & 0.060                   & 0.091                    & 0.048                   & 0.088                    & 0.045                   & 0.107                    & 0.058                   & 0.096                    & 0.052                   & 0.096                    & 0.054                   & 0.107                    & 0.058                   & 0.099                    & 0.046                   & 0.099                    & 0.045                  \\
                       & ctree.agl                & \textbf{0.087}           & \textbf{0.047}          & 0.089                    & 0.049                   & \textbf{0.080}           & \textbf{0.048}          & 0.091                    & 0.048                   & 0.099                    & 0.053                   & 0.095                    & 0.055                   & \textbf{0.096}           & \textbf{0.049}          & 0.104                    & 0.046                   & 0.096                    & 0.047                  \\ \hline
                       & sbart                    & 0.106                    & 0.050                   & \textbf{0.082}           & \textbf{0.037}          & 0.091                    & 0.042                   & 0.108                    & 0.051                   & \textbf{0.080}           & \textbf{0.037}          & \textbf{0.089}           & \textbf{0.043}          & 0.103                    & 0.046                   & \textbf{0.083}           & \textbf{0.033}          & \textbf{0.090}           & \textbf{0.042}         \\
                       & tbart                    & 0.107                    & 0.046                   & 0.090                    & 0.041                   & 0.090                    & 0.040                   & 0.102                    & 0.043                   & 0.089                    & 0.038                   & 0.094                    & 0.042                   & 0.105                    & 0.047                   & 0.090                    & 0.041                   & 0.091                    & 0.039                  \\
                       & xbart                    & 0.098                    & 0.042                   & \textbf{0.084}           & \textbf{0.037}          & \textbf{0.086}           & \textbf{0.039}          & 0.097                    & 0.041                   & \textbf{0.085}           & \textbf{0.036}          & 0.091                    & 0.039                   & \textbf{0.096}           & \textbf{0.042}          & \textbf{0.084}           & \textbf{0.037}          & \textbf{0.086}           & \textbf{0.037}         \\
                       & mbart                    & 0.208                    & 0.091                   & 0.209                    & 0.088                   & 0.184                    & 0.086                   & 0.177                    & 0.086                   & 0.189                    & 0.081                   & 0.173                    & 0.088                   & 0.209                    & 0.091                   & 0.193                    & 0.099                   & 0.188                    & 0.090                  \\
                       & drbart                   & \textbf{0.093}           & \textbf{0.039}          & \textbf{0.082}           & \textbf{0.039}          & \textbf{0.085}           & \textbf{0.038}          & \textbf{0.095}           & \textbf{0.042}          & 0.087                    & 0.039                   & \textbf{0.092}           & \textbf{0.041}          & \textbf{0.096}           & \textbf{0.042}          & 0.088                    & 0.038                   & \textbf{0.088}           & \textbf{0.040}         \\
5                      & rfbart                   & 0.456                    & 0.112                   & 0.368                    & 0.093                   & 0.431                    & 0.106                   & 0.461                    & 0.113                   & 0.375                    & 0.096                   & 0.434                    & 0.110                   & 0.450                    & 0.108                   & 0.368                    & 0.093                   & 0.425                    & 0.106                  \\
                       & rbart                    & 0.102                    & 0.038                   & 0.117                    & 0.045                   & 0.104                    & 0.037                   & 0.103                    & 0.037                   & 0.118                    & 0.045                   & 0.105                    & 0.039                   & 0.107                    & 0.039                   & 0.119                    & 0.047                   & 0.107                    & 0.039                  \\
                       & gbm.gl                   & 0.162                    & 0.058                   & 0.181                    & 0.061                   & 0.142                    & 0.056                   & 0.152                    & 0.058                   & 0.089                    & 0.039                   & 0.116                    & 0.053                   & 0.150                    & 0.062                   & 0.099                    & 0.045                   & 0.129                    & 0.056                  \\
                       & gbm.agl                  & \textbf{0.082}           & \textbf{0.038}          & 0.084                    & 0.036                   & \textbf{0.084}           & \textbf{0.037}          & \textbf{0.091}           & \textbf{0.037}          & \textbf{0.084}           & \textbf{0.040}          & \textbf{0.090}           & \textbf{0.039}          & 0.098                    & 0.041                   & \textbf{0.083}           & \textbf{0.039}          & 0.096                    & 0.041                  \\
                       & ctree.gl                 & 0.111                    & 0.051                   & 0.098                    & 0.046                   & 0.090                    & 0.043                   & 0.108                    & 0.047                   & 0.104                    & 0.045                   & 0.096                    & 0.044                   & 0.111                    & 0.046                   & 0.098                    & 0.047                   & 0.097                    & 0.044                  \\
                       & ctree.agl                & \textbf{0.086}           & \textbf{0.036}          & 0.097                    & 0.043                   & 0.088                    & 0.043                   & \textbf{0.097}           & \textbf{0.044}          & 0.102                    & 0.045                   & 0.099                    & 0.043                   & \textbf{0.097}           & \textbf{0.042}          & 0.106                    & 0.050                   & 0.094                    & 0.047                  \\ \hline
\end{tabular}
\vspace{0.1cm}
\\\textbf{note}: Bolded text means the best three results for each scenario. The first column indicates the number of groups, the second column indicates the evaluation method, and the third to last columns are the results of the different simulation scenarios. Each scenario is named “main effect function - treatment effect function” based on the corresponding data generating function, where “L” is a linear function, “S” is a step-wise function and “N” is a nonlinear function.
\end{table}
\end{landscape}

\begin{landscape}
\begin{table}[]
\caption{Results of mean Cohen’s kappa  across twelve scenarios for meta-learners using BART and proposed approaches in RCT setting}\label{kappa.rct.bart}
\vspace{1cm}
\small
\begin{tabular}{c|c|ll|ll|ll|ll|ll|ll|ll|ll|ll}
\multirow{2}{*}{Group} & \multirow{2}{*}{Methods} & \multicolumn{2}{c|}{L-L}                           & \multicolumn{2}{c|}{L-S}                           & \multicolumn{2}{c|}{L-N}                           & \multicolumn{2}{c|}{S-L}                           & \multicolumn{2}{c|}{S-S}                           & \multicolumn{2}{c|}{S-N}                           & \multicolumn{2}{c|}{N-L}                           & \multicolumn{2}{c|}{N-S}                           & \multicolumn{2}{c}{N-N}                           \\ \cline{3-20} 
                       &                          & \multicolumn{1}{c}{mean} & \multicolumn{1}{c|}{sd} & \multicolumn{1}{c}{mean} & \multicolumn{1}{c|}{sd} & \multicolumn{1}{c}{mean} & \multicolumn{1}{c|}{sd} & \multicolumn{1}{c}{mean} & \multicolumn{1}{c|}{sd} & \multicolumn{1}{c}{mean} & \multicolumn{1}{c|}{sd} & \multicolumn{1}{c}{mean} & \multicolumn{1}{c|}{sd} & \multicolumn{1}{c}{mean} & \multicolumn{1}{c|}{sd} & \multicolumn{1}{c}{mean} & \multicolumn{1}{c|}{sd} & \multicolumn{1}{c}{mean} & \multicolumn{1}{c}{sd} \\ \hline
                       & sbart                    & 0.850                    & 0.016                   & \textbf{0.890}           & \textbf{0.036}          & 0.855                    & 0.015                   & 0.850                    & 0.015                   & \textbf{0.890}           & \textbf{0.033}          & 0.855                    & 0.015                   & 0.846                    & 0.018                   & \textbf{0.893}           & \textbf{0.035}          & 0.851                    & 0.016                  \\
                       & tbart                    & 0.816                    & 0.016                   & 0.869                    & 0.018                   & 0.822                    & 0.014                   & 0.823                    & 0.016                   & 0.871                    & 0.019                   & 0.824                    & 0.015                   & 0.802                    & 0.018                   & 0.857                    & 0.019                   & 0.806                    & 0.015                  \\
                       & xbart                    & 0.867                    & 0.014                   & 0.883                    & 0.020                   & \textbf{0.869}           & \textbf{0.014}          & 0.866                    & 0.014                   & 0.883                    & 0.019                   & \textbf{0.866}           & \textbf{0.014}          & 0.860                    & 0.014                   & 0.880                    & 0.020                   & \textbf{0.863}           & \textbf{0.015}         \\
                       & mbart                    & 0.546                    & 0.044                   & 0.623                    & 0.049                   & 0.562                    & 0.039                   & 0.566                    & 0.038                   & 0.667                    & 0.044                   & 0.573                    & 0.035                   & 0.546                    & 0.040                   & 0.637                    & 0.050                   & 0.558                    & 0.037                  \\
                       & drbart                   & 0.788                    & 0.016                   & 0.847                    & 0.020                   & 0.783                    & 0.016                   & 0.777                    & 0.018                   & 0.862                    & 0.021                   & 0.768                    & 0.018                   & 0.743                    & 0.016                   & 0.850                    & 0.023                   & 0.732                    & 0.019                  \\
3                      & rfbart                   & 0.923                    & 0.015                   & 0.614                    & 0.020                   & 0.814                    & 0.021                   & 0.924                    & 0.015                   & 0.614                    & 0.021                   & 0.814                    & 0.021                   & \textbf{0.923}           & \textbf{0.014}          & 0.613                    & 0.021                   & 0.813                    & 0.021                  \\
                       & rbart                    & 0.941                    & 0.013                   & 0.617                    & 0.020                   & 0.826                    & 0.020                   & \textbf{0.942}           & \textbf{0.014}          & 0.617                    & 0.021                   & 0.826                    & 0.020                   & \textbf{0.941}           & \textbf{0.015}          & 0.617                    & 0.021                   & 0.824                    & 0.021                  \\
                       & gbm.gl                   & \textbf{0.942}           & \textbf{0.012}          & 0.845                    & 0.027                   & \textbf{0.881}           & \textbf{0.016}          & \textbf{0.939}           & \textbf{0.013}          & 0.874                    & 0.034                   & \textbf{0.867}           & \textbf{0.017}          & \textbf{0.920}           & \textbf{0.014}          & 0.877                    & 0.045                   & \textbf{0.869}           & \textbf{0.016}         \\
                       & gbm.agl                  & \textbf{0.952}           & \textbf{0.012}          & 0.841                    & 0.028                   & 0.866                    & 0.016                   & \textbf{0.942}           & \textbf{0.014}          & 0.873                    & 0.037                   & 0.856                    & 0.016                   & 0.917                    & 0.014                   & 0.875                    & 0.047                   & 0.856                    & 0.017                  \\
                       & ctree.gl                 & \textbf{0.942}           & \textbf{0.012}          & \textbf{0.897}           & \textbf{0.029}          & \textbf{0.880}           & \textbf{0.014}          & 0.932                    & 0.014                   & \textbf{0.902}           & \textbf{0.029}          & \textbf{0.870}           & \textbf{0.016}          & 0.901                    & 0.015                   & \textbf{0.899}           & \textbf{0.031}          & \textbf{0.864}           & \textbf{0.020}         \\
                       & ctree.agl                & 0.941                    & 0.013                   & \textbf{0.899}           & \textbf{0.030}          & 0.864                    & 0.015                   & 0.927                    & 0.017                   & \textbf{0.903}           & \textbf{0.028}          & 0.854                    & 0.018                   & 0.892                    & 0.016                   & \textbf{0.894}           & \textbf{0.030}          & 0.844                    & 0.021                  \\ \hline
                       & sbart                    & 0.827                    & 0.018                   & 0.862                    & 0.036                   & 0.835                    & 0.016                   & 0.830                    & 0.020                   & 0.862                    & 0.036                   & 0.837                    & 0.019                   & 0.823                    & 0.019                   & 0.864                    & 0.035                   & 0.831                    & 0.015                  \\
                       & tbart                    & 0.797                    & 0.017                   & 0.861                    & 0.017                   & 0.807                    & 0.014                   & 0.805                    & 0.014                   & 0.864                    & 0.016                   & 0.807                    & 0.014                   & 0.781                    & 0.015                   & 0.849                    & 0.017                   & 0.792                    & 0.015                  \\
                       & xbart                    & 0.847                    & 0.014                   & 0.874                    & 0.017                   & 0.852                    & 0.014                   & 0.846                    & 0.014                   & 0.874                    & 0.017                   & 0.848                    & 0.013                   & 0.840                    & 0.014                   & \textbf{0.870}           & \textbf{0.019}          & 0.846                    & 0.014                  \\
                       & mbart                    & 0.507                    & 0.037                   & 0.559                    & 0.048                   & 0.528                    & 0.033                   & 0.530                    & 0.035                   & 0.607                    & 0.040                   & 0.545                    & 0.033                   & 0.499                    & 0.034                   & 0.572                    & 0.045                   & 0.521                    & 0.035                  \\
                       & drbart                   & 0.761                    & 0.018                   & 0.840                    & 0.020                   & 0.767                    & 0.016                   & 0.761                    & 0.018                   & 0.858                    & 0.020                   & 0.761                    & 0.017                   & 0.723                    & 0.020                   & 0.826                    & 0.025                   & 0.725                    & 0.019                  \\
4                      & rfbart                   & 0.863                    & 0.018                   & 0.569                    & 0.025                   & 0.778                    & 0.022                   & 0.863                    & 0.019                   & 0.569                    & 0.025                   & 0.777                    & 0.023                   & 0.861                    & 0.018                   & 0.568                    & 0.025                   & 0.776                    & 0.024                  \\
                       & rbart                    & \textbf{0.926}           & \textbf{0.016}          & 0.592                    & 0.023                   & 0.821                    & 0.020                   & \textbf{0.927}           & \textbf{0.015}          & 0.593                    & 0.023                   & 0.821                    & 0.022                   & 0.922                    & 0.017                   & 0.592                    & 0.023                   & 0.820                    & 0.024                  \\
                       & gbm.gl                   & \textbf{0.932}           & \textbf{0.015}          & 0.863                    & 0.029                   & \textbf{0.872}           & \textbf{0.016}          & \textbf{0.923}           & \textbf{0.016}          & \textbf{0.880}           & \textbf{0.036}          & \textbf{0.862}           & \textbf{0.018}          & \textbf{0.902}           & \textbf{0.019}          & 0.865                    & 0.044                   & \textbf{0.858}           & \textbf{0.019}         \\
                       & gbm.agl                  & \textbf{0.940}           & \textbf{0.015}          & \textbf{0.873}           & \textbf{0.037}          & \textbf{0.864}           & \textbf{0.015}          & \textbf{0.925}           & \textbf{0.014}          & \textbf{0.880}           & \textbf{0.037}          & \textbf{0.849}           & \textbf{0.015}          & \textbf{0.898}           & \textbf{0.018}          & 0.861                    & 0.043                   & \textbf{0.849}           & \textbf{0.017}         \\
                       & ctree.gl                 & 0.928                    & 0.014                   & \textbf{0.884}           & \textbf{0.026}          & \textbf{0.866}           & \textbf{0.015}          & 0.914                    & 0.017                   & \textbf{0.879}           & \textbf{0.027}          & \textbf{0.854}           & \textbf{0.017}          & \textbf{0.885}           & \textbf{0.019}          & \textbf{0.878}           & \textbf{0.026}          & \textbf{0.850}           & \textbf{0.018}         \\
                       & ctree.agl                & 0.924                    & 0.018                   & \textbf{0.883}           & \textbf{0.027}          & 0.839                    & 0.017                   & 0.901                    & 0.018                   & 0.875                    & 0.026                   & 0.825                    & 0.019                   & 0.858                    & 0.022                   & \textbf{0.871}           & \textbf{0.025}          & 0.820                    & 0.022                  \\ \hline
                       & sbart                    & 0.784                    & 0.020                   & 0.892                    & 0.030                   & 0.800                    & 0.021                   & 0.784                    & 0.019                   & 0.891                    & 0.026                   & 0.800                    & 0.016                   & 0.776                    & 0.019                   & 0.886                    & 0.028                   & 0.791                    & 0.022                  \\
                       & tbart                    & 0.752                    & 0.018                   & 0.871                    & 0.018                   & 0.759                    & 0.017                   & 0.762                    & 0.019                   & 0.877                    & 0.018                   & 0.760                    & 0.017                   & 0.729                    & 0.020                   & 0.850                    & 0.019                   & 0.736                    & 0.020                  \\
                       & xbart                    & 0.806                    & 0.018                   & 0.889                    & 0.017                   & 0.807                    & 0.017                   & 0.806                    & 0.019                   & 0.890                    & 0.017                   & 0.803                    & 0.018                   & 0.792                    & 0.018                   & 0.879                    & 0.017                   & 0.796                    & 0.018                  \\
                       & mbart                    & 0.418                    & 0.033                   & 0.475                    & 0.045                   & 0.430                    & 0.032                   & 0.437                    & 0.033                   & 0.527                    & 0.041                   & 0.440                    & 0.030                   & 0.405                    & 0.034                   & 0.487                    & 0.042                   & 0.416                    & 0.033                  \\
                       & drbart                   & 0.709                    & 0.020                   & 0.844                    & 0.023                   & 0.712                    & 0.019                   & 0.707                    & 0.019                   & 0.863                    & 0.023                   & 0.706                    & 0.019                   & 0.682                    & 0.019                   & 0.833                    & 0.027                   & 0.681                    & 0.021                  \\
5                      & rfbart                   & 0.811                    & 0.022                   & 0.553                    & 0.022                   & 0.711                    & 0.026                   & 0.811                    & 0.022                   & 0.553                    & 0.023                   & 0.710                    & 0.027                   & 0.808                    & 0.023                   & 0.553                    & 0.023                   & 0.709                    & 0.027                  \\
                       & rbart                    & 0.902                    & 0.018                   & 0.578                    & 0.022                   & 0.771                    & 0.023                   & \textbf{0.902}           & \textbf{0.018}          & 0.577                    & 0.023                   & 0.770                    & 0.024                   & \textbf{0.897}           & \textbf{0.019}          & 0.577                    & 0.022                   & 0.769                    & 0.024                  \\
                       & gbm.gl                   & \textbf{0.915}           & \textbf{0.016}          & 0.899                    & 0.030                   & \textbf{0.845}           & \textbf{0.019}          & \textbf{0.903}           & \textbf{0.016}          & \textbf{0.925}           & \textbf{0.028}          & \textbf{0.831}           & \textbf{0.019}          & \textbf{0.879}           & \textbf{0.020}          & \textbf{0.909}           & \textbf{0.037}          & \textbf{0.826}           & \textbf{0.018}         \\
                       & gbm.agl                  & \textbf{0.918}           & \textbf{0.016}          & \textbf{0.912}           & \textbf{0.028}          & \textbf{0.830}           & \textbf{0.018}          & \textbf{0.897}           & \textbf{0.017}          & \textbf{0.914}           & \textbf{0.028}          & \textbf{0.814}           & \textbf{0.018}          & \textbf{0.865}           & \textbf{0.020}          & \textbf{0.897}           & \textbf{0.034}          & \textbf{0.809}           & \textbf{0.020}         \\
                       & ctree.gl                 & \textbf{0.909}           & \textbf{0.016}          & \textbf{0.921}           & \textbf{0.027}          & \textbf{0.830}           & \textbf{0.018}          & 0.890                    & 0.018                   & \textbf{0.922}           & \textbf{0.026}          & \textbf{0.822}           & \textbf{0.021}          & 0.858                    & 0.021                   & \textbf{0.914}           & \textbf{0.024}          & \textbf{0.812}           & \textbf{0.020}         \\
                       & ctree.agl                & 0.903                    & 0.024                   & \textbf{0.912}           & \textbf{0.031}          & 0.790                    & 0.021                   & 0.864                    & 0.028                   & 0.905                    & 0.029                   & 0.777                    & 0.025                   & 0.819                    & 0.034                   & 0.890                    & 0.027                   & 0.761                    & 0.026                  \\ \hline
\end{tabular}
\\\textbf{note}: Bolded text means the best three results for each scenario. The first column indicates the number of groups, the second column indicates the evaluation method, and the third to last columns are the results of the different simulation scenarios. Each scenario is named “main effect function - treatment effect function” based on the corresponding data generating function, where “L” is a linear function, “S” is a step-wise function and “N” is a nonlinear function.
\end{table}
\end{landscape}

\begin{landscape}
\begin{table}[]
\caption{Results of mean Cohen’s kappa  across twelve scenarios for meta-learners using BART and proposed approaches in observational setting}\label{kappa.obv.bart}
\vspace{1cm}
\small
\begin{tabular}{c|c|ll|ll|ll|ll|ll|ll|ll|ll|ll}
\multirow{2}{*}{Group} & \multirow{2}{*}{Methods} & \multicolumn{2}{c|}{L-L}                           & \multicolumn{2}{c|}{L-S}                           & \multicolumn{2}{c|}{L-N}                           & \multicolumn{2}{c|}{S-L}                           & \multicolumn{2}{c|}{S-S}                           & \multicolumn{2}{c|}{S-N}                           & \multicolumn{2}{c|}{N-L}                           & \multicolumn{2}{c|}{N-S}                           & \multicolumn{2}{c}{N-N}                           \\ \cline{3-20} 
                       &                          & \multicolumn{1}{c}{mean} & \multicolumn{1}{c|}{sd} & \multicolumn{1}{c}{mean} & \multicolumn{1}{c|}{sd} & \multicolumn{1}{c}{mean} & \multicolumn{1}{c|}{sd} & \multicolumn{1}{c}{mean} & \multicolumn{1}{c|}{sd} & \multicolumn{1}{c}{mean} & \multicolumn{1}{c|}{sd} & \multicolumn{1}{c}{mean} & \multicolumn{1}{c|}{sd} & \multicolumn{1}{c}{mean} & \multicolumn{1}{c|}{sd} & \multicolumn{1}{c}{mean} & \multicolumn{1}{c|}{sd} & \multicolumn{1}{c}{mean} & \multicolumn{1}{c}{sd} \\ \hline
                       & sbart                    & 0.840                    & 0.017                   & \textbf{0.894}           & \textbf{0.035}          & 0.850                    & 0.016                   & 0.835                    & 0.018                   & \textbf{0.890}           & \textbf{0.033}          & 0.850                    & 0.016                   & 0.834                    & 0.016                   & \textbf{0.889}           & \textbf{0.033}          & 0.847                    & 0.016                  \\
                       & tbart                    & 0.808                    & 0.016                   & 0.869                    & 0.015                   & 0.822                    & 0.015                   & 0.813                    & 0.015                   & 0.866                    & 0.017                   & 0.818                    & 0.015                   & 0.791                    & 0.015                   & 0.852                    & 0.018                   & 0.801                    & 0.017                  \\
                       & xbart                    & 0.859                    & 0.016                   & 0.879                    & 0.017                   & \textbf{0.864}           & \textbf{0.013}          & 0.856                    & 0.015                   & 0.877                    & 0.018                   & \textbf{0.859}           & \textbf{0.014}          & 0.848                    & 0.016                   & \textbf{0.874}           & \textbf{0.018}          & \textbf{0.853}           & \textbf{0.014}         \\
                       & mbart                    & 0.422                    & 0.034                   & 0.481                    & 0.049                   & 0.438                    & 0.040                   & 0.443                    & 0.037                   & 0.529                    & 0.053                   & 0.460                    & 0.040                   & 0.430                    & 0.035                   & 0.505                    & 0.050                   & 0.444                    & 0.039                  \\
                       & drbart                   & 0.778                    & 0.018                   & 0.843                    & 0.019                   & 0.777                    & 0.017                   & 0.770                    & 0.019                   & 0.856                    & 0.019                   & 0.764                    & 0.019                   & 0.735                    & 0.020                   & 0.842                    & 0.024                   & 0.724                    & 0.019                  \\
3                      & rfbart                   & 0.768                    & 0.038                   & 0.546                    & 0.032                   & 0.686                    & 0.038                   & 0.768                    & 0.036                   & 0.545                    & 0.031                   & 0.685                    & 0.038                   & 0.768                    & 0.038                   & 0.545                    & 0.033                   & 0.686                    & 0.038                  \\
                       & rbart                    & 0.936                    & 0.016                   & 0.619                    & 0.023                   & 0.822                    & 0.022                   & \textbf{0.936}           & \textbf{0.016}          & 0.620                    & 0.023                   & 0.820                    & 0.023                   & \textbf{0.934}           & \textbf{0.017}          & 0.619                    & 0.024                   & 0.821                    & 0.023                  \\
                       & gbm.gl                   & 0.922                    & 0.017                   & 0.844                    & 0.033                   & 0.855                    & 0.017                   & 0.909                    & 0.018                   & 0.866                    & 0.030                   & \textbf{0.844}           & \textbf{0.019}          & \textbf{0.900}           & \textbf{0.021}          & 0.862                    & 0.041                   & \textbf{0.844}           & \textbf{0.020}         \\
                       & gbm.agl                  & \textbf{0.944}           & \textbf{0.016}          & 0.832                    & 0.038                   & 0.856                    & 0.019                   & \textbf{0.932}           & \textbf{0.015}          & 0.856                    & 0.040                   & 0.844                    & 0.019                   & \textbf{0.904}           & \textbf{0.020}          & 0.845                    & 0.043                   & 0.843                    & 0.019                  \\
                       & ctree.gl                 & \textbf{0.937}           & \textbf{0.014}          & \textbf{0.890}           & \textbf{0.021}          & \textbf{0.873}           & \textbf{0.016}          & \textbf{0.925}           & \textbf{0.017}          & \textbf{0.879}           & \textbf{0.024}          & \textbf{0.856}           & \textbf{0.019}          & 0.894                    & 0.018                   & 0.871                    & 0.021                   & \textbf{0.853}           & \textbf{0.019}         \\
                       & ctree.agl                & \textbf{0.934}           & \textbf{0.016}          & \textbf{0.889}           & \textbf{0.025}          & \textbf{0.857}           & \textbf{0.018}          & 0.919                    & 0.019                   & \textbf{0.879}           & \textbf{0.028}          & 0.838                    & 0.021                   & 0.883                    & 0.020                   & \textbf{0.865}           & \textbf{0.023}          & 0.836                    & 0.020                  \\ \hline
                       & sbart                    & 0.809                    & 0.023                   & 0.858                    & 0.034                   & 0.829                    & 0.020                   & 0.812                    & 0.020                   & 0.860                    & 0.034                   & 0.828                    & 0.019                   & 0.807                    & 0.021                   & 0.857                    & 0.032                   & 0.825                    & 0.020                  \\
                       & tbart                    & 0.788                    & 0.017                   & 0.861                    & 0.019                   & 0.805                    & 0.014                   & 0.798                    & 0.017                   & 0.860                    & 0.020                   & 0.805                    & 0.015                   & 0.765                    & 0.018                   & 0.845                    & 0.021                   & 0.779                    & 0.018                  \\
                       & xbart                    & 0.835                    & 0.014                   & \textbf{0.870}           & \textbf{0.019}          & 0.844                    & 0.014                   & 0.835                    & 0.016                   & 0.868                    & 0.019                   & \textbf{0.841}           & \textbf{0.014}          & 0.819                    & 0.016                   & \textbf{0.864}           & \textbf{0.019}          & \textbf{0.830}           & \textbf{0.015}         \\
                       & mbart                    & 0.420                    & 0.036                   & 0.449                    & 0.043                   & 0.437                    & 0.037                   & 0.443                    & 0.037                   & 0.497                    & 0.046                   & 0.454                    & 0.039                   & 0.419                    & 0.038                   & 0.469                    & 0.040                   & 0.438                    & 0.038                  \\
                       & drbart                   & 0.751                    & 0.017                   & 0.836                    & 0.024                   & 0.759                    & 0.017                   & 0.752                    & 0.017                   & 0.850                    & 0.023                   & 0.753                    & 0.018                   & 0.709                    & 0.020                   & 0.820                    & 0.025                   & 0.715                    & 0.020                  \\
4                      & rfbart                   & 0.681                    & 0.028                   & 0.489                    & 0.028                   & 0.625                    & 0.029                   & 0.682                    & 0.029                   & 0.489                    & 0.028                   & 0.624                    & 0.029                   & 0.681                    & 0.028                   & 0.487                    & 0.028                   & 0.623                    & 0.029                  \\
                       & rbart                    & \textbf{0.923}           & \textbf{0.016}          & 0.590                    & 0.021                   & 0.813                    & 0.021                   & \textbf{0.924}           & \textbf{0.017}          & 0.590                    & 0.021                   & 0.811                    & 0.022                   & \textbf{0.922}           & \textbf{0.019}          & 0.590                    & 0.022                   & 0.809                    & 0.023                  \\
                       & gbm.gl                   & 0.909                    & 0.018                   & 0.864                    & 0.031                   & \textbf{0.846}           & \textbf{0.020}          & 0.901                    & 0.019                   & \textbf{0.887}           & \textbf{0.036}          & \textbf{0.839}           & \textbf{0.020}          & \textbf{0.883}           & \textbf{0.023}          & \textbf{0.871}           & \textbf{0.036}          & \textbf{0.836}           & \textbf{0.020}         \\
                       & gbm.agl                  & \textbf{0.932}           & \textbf{0.018}          & \textbf{0.874}           & \textbf{0.037}          & \textbf{0.848}           & \textbf{0.018}          & \textbf{0.917}           & \textbf{0.017}          & \textbf{0.879}           & \textbf{0.036}          & 0.837                    & 0.018                   & \textbf{0.880}           & \textbf{0.019}          & 0.850                    & 0.035                   & 0.827                    & 0.020                  \\
                       & ctree.gl                 & \textbf{0.919}           & \textbf{0.016}          & \textbf{0.871}           & \textbf{0.027}          & \textbf{0.857}           & \textbf{0.018}          & \textbf{0.906}           & \textbf{0.019}          & \textbf{0.871}           & \textbf{0.027}          & \textbf{0.846}           & \textbf{0.019}          & 0.876                    & 0.020                   & \textbf{0.864}           & \textbf{0.023}          & \textbf{0.839}           & \textbf{0.020}         \\
                       & ctree.agl                & 0.906                    & 0.022                   & 0.867                    & 0.030                   & 0.832                    & 0.020                   & 0.890                    & 0.024                   & 0.864                    & 0.027                   & 0.818                    & 0.021                   & 0.850                    & 0.025                   & 0.852                    & 0.025                   & 0.812                    & 0.023                  \\ \hline
                       & sbart                    & 0.761                    & 0.025                   & \textbf{0.886}           & \textbf{0.029}          & 0.784                    & 0.023                   & 0.762                    & 0.026                   & 0.876                    & 0.034                   & 0.782                    & 0.026                   & 0.752                    & 0.027                   & \textbf{0.875}           & \textbf{0.031}          & \textbf{0.779}           & \textbf{0.023}         \\
                       & tbart                    & 0.736                    & 0.022                   & 0.860                    & 0.020                   & 0.756                    & 0.018                   & 0.748                    & 0.021                   & 0.858                    & 0.019                   & 0.755                    & 0.019                   & 0.707                    & 0.024                   & 0.839                    & 0.023                   & 0.725                    & 0.021                  \\
                       & xbart                    & 0.785                    & 0.022                   & 0.875                    & 0.019                   & 0.800                    & 0.019                   & 0.786                    & 0.020                   & 0.869                    & 0.019                   & 0.791                    & 0.019                   & 0.762                    & 0.020                   & 0.862                    & 0.020                   & 0.777                    & 0.020                  \\
                       & mbart                    & 0.342                    & 0.031                   & 0.379                    & 0.033                   & 0.352                    & 0.028                   & 0.359                    & 0.030                   & 0.431                    & 0.036                   & 0.369                    & 0.030                   & 0.338                    & 0.029                   & 0.399                    & 0.038                   & 0.349                    & 0.031                  \\
                       & drbart                   & 0.697                    & 0.020                   & 0.835                    & 0.024                   & 0.706                    & 0.022                   & 0.690                    & 0.020                   & 0.844                    & 0.024                   & 0.698                    & 0.022                   & 0.666                    & 0.020                   & 0.820                    & 0.030                   & 0.670                    & 0.021                  \\
5                      & rfbart                   & 0.604                    & 0.031                   & 0.471                    & 0.025                   & 0.536                    & 0.031                   & 0.604                    & 0.031                   & 0.470                    & 0.024                   & 0.534                    & 0.032                   & 0.601                    & 0.031                   & 0.468                    & 0.024                   & 0.532                    & 0.032                  \\
                       & rbart                    & 0.893                    & 0.020                   & 0.577                    & 0.024                   & 0.762                    & 0.025                   & \textbf{0.893}           & \textbf{0.019}          & 0.576                    & 0.024                   & 0.760                    & 0.026                   & \textbf{0.888}           & \textbf{0.021}          & 0.576                    & 0.024                   & 0.758                    & 0.027                  \\
                       & gbm.gl                   & \textbf{0.893}           & \textbf{0.019}          & \textbf{0.908}           & \textbf{0.034}          & \textbf{0.818}           & \textbf{0.023}          & \textbf{0.883}           & \textbf{0.021}          & \textbf{0.920}           & \textbf{0.033}          & \textbf{0.814}           & \textbf{0.022}          & \textbf{0.857}           & \textbf{0.024}          & \textbf{0.904}           & \textbf{0.032}          & \textbf{0.802}           & \textbf{0.025}         \\
                       & gbm.agl                  & \textbf{0.909}           & \textbf{0.019}          & \textbf{0.907}           & \textbf{0.032}          & \textbf{0.818}           & \textbf{0.022}          & 0.877                    & 0.022                   & \textbf{0.897}           & \textbf{0.035}          & \textbf{0.801}           & \textbf{0.020}          & \textbf{0.835}           & \textbf{0.025}          & 0.871                    & 0.036                   & 0.779                    & 0.024                  \\
                       & ctree.gl                 & \textbf{0.897}           & \textbf{0.020}          & 0.882                    & 0.032                   & \textbf{0.820}           & \textbf{0.019}          & \textbf{0.880}           & \textbf{0.024}          & \textbf{0.879}           & \textbf{0.039}          & \textbf{0.806}           & \textbf{0.021}          & 0.852                    & 0.025                   & \textbf{0.871}           & \textbf{0.034}          & \textbf{0.797}           & \textbf{0.023}         \\
                       & ctree.agl                & 0.874                    & 0.025                   & 0.855                    & 0.039                   & 0.778                    & 0.021                   & 0.840                    & 0.030                   & 0.852                    & 0.046                   & 0.758                    & 0.025                   & 0.799                    & 0.029                   & 0.835                    & 0.038                   & 0.751                    & 0.026                  \\ \hline
\end{tabular}
\\\textbf{note}: Bolded text means the best three results for each scenario. The first column indicates the number of groups, the second column indicates the evaluation method, and the third to last columns are the results of the different simulation scenarios. Each scenario is named “main effect function - treatment effect function” based on the corresponding data generating function, where “L” is a linear function, “S” is a step-wise function and “N” is a nonlinear function.
\end{table}
\end{landscape}

\begin{landscape}
\begin{table}[]
\caption{Results of mean Spearman’s rank correlation across twelve scenarios for meta-learners using BART and proposed approaches in RCT setting}\label{mord.rct.bart}
\vspace{1cm}
\small
\begin{tabular}{c|c|ll|ll|ll|ll|ll|ll|ll|ll|ll}
\multirow{2}{*}{Group} & \multirow{2}{*}{Methods} & \multicolumn{2}{c|}{L-L}                           & \multicolumn{2}{c|}{L-S}                           & \multicolumn{2}{c|}{L-N}                           & \multicolumn{2}{c|}{S-L}                           & \multicolumn{2}{c|}{S-S}                           & \multicolumn{2}{c|}{S-N}                           & \multicolumn{2}{c|}{N-L}                           & \multicolumn{2}{c|}{N-S}                           & \multicolumn{2}{c}{N-N}                           \\ \cline{3-20} 
                       &                          & \multicolumn{1}{c}{mean} & \multicolumn{1}{c|}{sd} & \multicolumn{1}{c}{mean} & \multicolumn{1}{c|}{sd} & \multicolumn{1}{c}{mean} & \multicolumn{1}{c|}{sd} & \multicolumn{1}{c}{mean} & \multicolumn{1}{c|}{sd} & \multicolumn{1}{c}{mean} & \multicolumn{1}{c|}{sd} & \multicolumn{1}{c}{mean} & \multicolumn{1}{c|}{sd} & \multicolumn{1}{c}{mean} & \multicolumn{1}{c|}{sd} & \multicolumn{1}{c}{mean} & \multicolumn{1}{c|}{sd} & \multicolumn{1}{c}{mean} & \multicolumn{1}{c}{sd} \\ \hline
                       & sbart                    & 0.821                    & 0.017                   & \textbf{0.879}           & \textbf{0.030}          & 0.831                    & 0.016                   & 0.821                    & 0.016                   & \textbf{0.878}           & \textbf{0.030}          & 0.833                    & 0.015                   & 0.818                    & 0.017                   & \textbf{0.879}           & \textbf{0.032}          & 0.828                    & 0.016                  \\
                       & tbart                    & 0.781                    & 0.016                   & 0.840                    & 0.018                   & 0.790                    & 0.017                   & 0.789                    & 0.016                   & 0.843                    & 0.018                   & 0.793                    & 0.017                   & 0.766                    & 0.019                   & 0.826                    & 0.019                   & 0.773                    & 0.018                  \\
                       & xbart                    & 0.840                    & 0.015                   & 0.860                    & 0.019                   & 0.844                    & 0.015                   & 0.838                    & 0.014                   & 0.860                    & 0.019                   & \textbf{0.841}           & \textbf{0.015}          & 0.833                    & 0.016                   & 0.856                    & 0.018                   & \textbf{0.838}           & \textbf{0.017}         \\
                       & mbart                    & 0.483                    & 0.046                   & 0.598                    & 0.053                   & 0.509                    & 0.043                   & 0.508                    & 0.041                   & 0.649                    & 0.044                   & 0.522                    & 0.040                   & 0.487                    & 0.043                   & 0.615                    & 0.051                   & 0.506                    & 0.043                  \\
                       & drbart                   & 0.748                    & 0.018                   & 0.801                    & 0.017                   & 0.745                    & 0.021                   & 0.736                    & 0.019                   & 0.822                    & 0.017                   & 0.729                    & 0.023                   & 0.714                    & 0.018                   & 0.816                    & 0.019                   & 0.706                    & 0.023                  \\
3                      & rfbart                   & 0.910                    & 0.018                   & 0.603                    & 0.020                   & 0.792                    & 0.023                   & 0.910                    & 0.017                   & 0.603                    & 0.021                   & 0.792                    & 0.023                   & \textbf{0.910}           & \textbf{0.018}          & 0.602                    & 0.019                   & 0.789                    & 0.025                  \\
                       & rbart                    & \textbf{0.927}           & \textbf{0.017}          & 0.608                    & 0.019                   & 0.805                    & 0.024                   & \textbf{0.927}           & \textbf{0.016}          & 0.608                    & 0.020                   & 0.804                    & 0.023                   & \textbf{0.926}           & \textbf{0.018}          & 0.608                    & 0.019                   & 0.802                    & 0.026                  \\
                       & gbm.gl                   & 0.921                    & 0.018                   & 0.802                    & 0.019                   & \textbf{0.865}           & \textbf{0.016}          & \textbf{0.923}           & \textbf{0.018}          & 0.848                    & 0.028                   & \textbf{0.855}           & \textbf{0.017}          & \textbf{0.904}           & \textbf{0.018}          & 0.852                    & 0.032                   & \textbf{0.851}           & \textbf{0.017}         \\
                       & gbm.agl                  & \textbf{0.939}           & \textbf{0.016}          & 0.799                    & 0.019                   & \textbf{0.851}           & \textbf{0.015}          & \textbf{0.930}           & \textbf{0.018}          & 0.851                    & 0.037                   & 0.838                    & 0.016                   & 0.897                    & 0.019                   & 0.858                    & 0.037                   & 0.835                    & 0.018                  \\
                       & ctree.gl                 & \textbf{0.926}           & \textbf{0.015}          & \textbf{0.863}           & \textbf{0.028}          & \textbf{0.861}           & \textbf{0.014}          & 0.916                    & 0.017                   & \textbf{0.873}           & \textbf{0.029}          & \textbf{0.848}           & \textbf{0.016}          & 0.881                    & 0.017                   & \textbf{0.869}           & \textbf{0.031}          & \textbf{0.839}           & \textbf{0.020}         \\
                       & ctree.agl                & 0.927                    & 0.016                   & \textbf{0.868}           & \textbf{0.030}          & 0.843                    & 0.017                   & 0.910                    & 0.021                   & \textbf{0.877}           & \textbf{0.029}          & 0.830                    & 0.018                   & 0.867                    & 0.020                   & \textbf{0.865}           & \textbf{0.031}          & 0.815                    & 0.023                  \\ \hline
                       & sbart                    & 0.825                    & 0.013                   & \textbf{0.818}           & \textbf{0.041}          & 0.852                    & 0.012                   & 0.827                    & 0.015                   & \textbf{0.819}           & \textbf{0.038}          & 0.853                    & 0.013                   & 0.819                    & 0.015                   & \textbf{0.818}           & \textbf{0.041}          & 0.848                    & 0.013                  \\
                       & tbart                    & 0.803                    & 0.014                   & 0.778                    & 0.024                   & 0.828                    & 0.012                   & 0.810                    & 0.014                   & 0.783                    & 0.025                   & 0.828                    & 0.011                   & 0.788                    & 0.014                   & 0.765                    & 0.024                   & 0.811                    & 0.014                  \\
                       & xbart                    & 0.844                    & 0.012                   & 0.805                    & 0.024                   & 0.863                    & 0.012                   & 0.843                    & 0.013                   & \textbf{0.805}           & \textbf{0.026}          & \textbf{0.861}           & \textbf{0.011}          & 0.836                    & 0.013                   & \textbf{0.799}           & \textbf{0.025}          & 0.856                    & 0.013                  \\
                       & mbart                    & 0.549                    & 0.031                   & 0.535                    & 0.037                   & 0.575                    & 0.031                   & 0.570                    & 0.030                   & 0.575                    & 0.034                   & 0.590                    & 0.029                   & 0.541                    & 0.030                   & 0.543                    & 0.041                   & 0.562                    & 0.030                  \\
                       & drbart                   & 0.763                    & 0.016                   & 0.715                    & 0.027                   & 0.788                    & 0.013                   & 0.766                    & 0.014                   & 0.749                    & 0.028                   & 0.786                    & 0.013                   & 0.739                    & 0.014                   & 0.723                    & 0.030                   & 0.761                    & 0.014                  \\
4                      & rfbart                   & 0.882                    & 0.017                   & 0.488                    & 0.018                   & 0.807                    & 0.018                   & 0.882                    & 0.017                   & 0.488                    & 0.017                   & 0.807                    & 0.019                   & 0.880                    & 0.017                   & 0.488                    & 0.018                   & 0.805                    & 0.019                  \\
                       & rbart                    & \textbf{0.925}           & \textbf{0.014}          & 0.494                    & 0.018                   & 0.833                    & 0.017                   & \textbf{0.925}           & \textbf{0.014}          & 0.493                    & 0.018                   & 0.834                    & 0.017                   & \textbf{0.921}           & \textbf{0.014}          & 0.494                    & 0.018                   & 0.833                    & 0.018                  \\
                       & gbm.gl                   & 0.914                    & 0.017                   & 0.738                    & 0.034                   & \textbf{0.877}           & \textbf{0.012}          & 0.906                    & 0.016                   & 0.780                    & 0.046                   & \textbf{0.868}           & \textbf{0.014}          & \textbf{0.890}           & \textbf{0.018}          & 0.754                    & 0.052                   & \textbf{0.862}           & \textbf{0.015}         \\
                       & gbm.agl                  & \textbf{0.937}           & \textbf{0.015}          & \textbf{0.810}           & \textbf{0.058}          & \textbf{0.869}           & \textbf{0.014}          & \textbf{0.921}           & \textbf{0.013}          & \textbf{0.824}           & \textbf{0.051}          & 0.859                    & 0.012                   & \textbf{0.896}           & \textbf{0.017}          & \textbf{0.797}           & \textbf{0.054}          & \textbf{0.857}           & \textbf{0.015}         \\
                       & ctree.gl                 & 0.915                    & 0.016                   & 0.803                    & 0.042                   & \textbf{0.874}           & \textbf{0.011}          & \textbf{0.903}           & \textbf{0.015}          & 0.795                    & 0.038                   & \textbf{0.863}           & \textbf{0.014}          & 0.879                    & 0.016                   & 0.794                    & 0.043                   & \textbf{0.859}           & \textbf{0.012}         \\
                       & ctree.agl                & \textbf{0.917}           & \textbf{0.018}          & \textbf{0.815}           & \textbf{0.042}          & 0.853                    & 0.015                   & 0.895                    & 0.017                   & 0.799                    & 0.035                   & 0.838                    & 0.016                   & 0.856                    & 0.019                   & 0.792                    & 0.040                   & 0.831                    & 0.017                  \\ \hline
                       & sbart                    & 0.761                    & 0.013                   & \textbf{0.808}           & \textbf{0.032}          & 0.785                    & 0.014                   & 0.761                    & 0.015                   & 0.807                    & 0.030                   & 0.788                    & 0.016                   & 0.755                    & 0.015                   & \textbf{0.803}           & \textbf{0.035}          & 0.780                    & 0.015                  \\
                       & tbart                    & 0.732                    & 0.013                   & 0.738                    & 0.021                   & 0.747                    & 0.014                   & 0.744                    & 0.014                   & 0.746                    & 0.022                   & 0.749                    & 0.014                   & 0.712                    & 0.014                   & 0.717                    & 0.024                   & 0.724                    & 0.014                  \\
                       & xbart                    & 0.783                    & 0.013                   & 0.768                    & 0.021                   & 0.793                    & 0.013                   & 0.783                    & 0.014                   & 0.768                    & 0.022                   & 0.789                    & 0.014                   & 0.771                    & 0.014                   & 0.757                    & 0.023                   & 0.781                    & 0.014                  \\
                       & mbart                    & 0.415                    & 0.026                   & 0.381                    & 0.031                   & 0.427                    & 0.028                   & 0.434                    & 0.025                   & 0.435                    & 0.027                   & 0.438                    & 0.028                   & 0.407                    & 0.029                   & 0.398                    & 0.028                   & 0.414                    & 0.029                  \\
                       & drbart                   & 0.701                    & 0.014                   & 0.695                    & 0.023                   & 0.711                    & 0.014                   & 0.694                    & 0.016                   & 0.726                    & 0.026                   & 0.701                    & 0.014                   & 0.674                    & 0.015                   & 0.704                    & 0.023                   & 0.680                    & 0.016                  \\
5                      & rfbart                   & 0.842                    & 0.019                   & 0.418                    & 0.015                   & 0.726                    & 0.017                   & 0.842                    & 0.019                   & 0.418                    & 0.016                   & 0.726                    & 0.017                   & 0.839                    & 0.021                   & 0.418                    & 0.016                   & 0.724                    & 0.018                  \\
                       & rbart                    & \textbf{0.888}           & \textbf{0.017}          & 0.424                    & 0.015                   & 0.754                    & 0.016                   & \textbf{0.887}           & \textbf{0.016}          & 0.424                    & 0.015                   & 0.753                    & 0.016                   & \textbf{0.883}           & \textbf{0.019}          & 0.423                    & 0.015                   & 0.751                    & 0.017                  \\
                       & gbm.gl                   & 0.880                    & 0.020                   & 0.765                    & 0.025                   & \textbf{0.824}           & \textbf{0.016}          & \textbf{0.873}           & \textbf{0.019}          & \textbf{0.830}           & \textbf{0.027}          & \textbf{0.807}           & \textbf{0.017}          & \textbf{0.847}           & \textbf{0.021}          & 0.791                    & 0.038                   & \textbf{0.804}           & \textbf{0.016}         \\
                       & gbm.agl                  & \textbf{0.902}           & \textbf{0.016}          & \textbf{0.816}           & \textbf{0.037}          & \textbf{0.812}           & \textbf{0.016}          & \textbf{0.879}           & \textbf{0.016}          & \textbf{0.829}           & \textbf{0.035}          & \textbf{0.792}           & \textbf{0.015}          & \textbf{0.843}           & \textbf{0.018}          & \textbf{0.799}           & \textbf{0.040}          & \textbf{0.788}           & \textbf{0.017}         \\
                       & ctree.gl                 & 0.879                    & 0.017                   & \textbf{0.800}           & \textbf{0.041}          & \textbf{0.811}           & \textbf{0.014}          & 0.861                    & 0.018                   & \textbf{0.817}           & \textbf{0.043}          & \textbf{0.800}           & \textbf{0.014}          & 0.833                    & 0.018                   & \textbf{0.795}           & \textbf{0.035}          & \textbf{0.789}           & \textbf{0.017}         \\
                       & ctree.agl                & \textbf{0.881}           & \textbf{0.027}          & 0.792                    & 0.046                   & 0.773                    & 0.018                   & 0.839                    & 0.029                   & 0.790                    & 0.041                   & 0.758                    & 0.018                   & 0.796                    & 0.034                   & 0.759                    & 0.037                   & 0.738                    & 0.022                  \\ \hline
\end{tabular}
\\\textbf{note}: Bolded text means the best three results for each scenario. The first column indicates the number of groups, the second column indicates the evaluation method, and the third to last columns are the results of the different simulation scenarios. Each scenario is named “main effect function - treatment effect function” based on the corresponding data generating function, where “L” is a linear function, “S” is a step-wise function and “N” is a nonlinear function.
\end{table}
\end{landscape}

\begin{landscape}
\begin{table}[]
\caption{Results of mean Spearman’s rank correlation across twelve scenarios for meta-learners using BART and proposed approaches in observational setting}\label{mord.obv.bart}
\vspace{1cm}
\small
\begin{tabular}{c|c|ll|ll|ll|ll|ll|ll|ll|ll|ll}
\multirow{2}{*}{Group} & \multirow{2}{*}{Methods} & \multicolumn{2}{c|}{L-L}                           & \multicolumn{2}{c|}{L-S}                           & \multicolumn{2}{c|}{L-N}                           & \multicolumn{2}{c|}{S-L}                           & \multicolumn{2}{c|}{S-S}                           & \multicolumn{2}{c|}{S-N}                           & \multicolumn{2}{c|}{N-L}                           & \multicolumn{2}{c|}{N-S}                           & \multicolumn{2}{c}{N-N}                           \\ \cline{3-20} 
                       &                          & \multicolumn{1}{c}{mean} & \multicolumn{1}{c|}{sd} & \multicolumn{1}{c}{mean} & \multicolumn{1}{c|}{sd} & \multicolumn{1}{c}{mean} & \multicolumn{1}{c|}{sd} & \multicolumn{1}{c}{mean} & \multicolumn{1}{c|}{sd} & \multicolumn{1}{c}{mean} & \multicolumn{1}{c|}{sd} & \multicolumn{1}{c}{mean} & \multicolumn{1}{c|}{sd} & \multicolumn{1}{c}{mean} & \multicolumn{1}{c|}{sd} & \multicolumn{1}{c}{mean} & \multicolumn{1}{c|}{sd} & \multicolumn{1}{c}{mean} & \multicolumn{1}{c}{sd} \\ \hline
                       & sbart                    & 0.810                    & 0.019                   & \textbf{0.881}           & \textbf{0.035}          & 0.825                    & 0.019                   & 0.808                    & 0.020                   & \textbf{0.877}           & \textbf{0.031}          & 0.826                    & 0.016                   & 0.807                    & 0.017                   & \textbf{0.871}           & \textbf{0.030}          & 0.823                    & 0.017                  \\
                       & tbart                    & 0.773                    & 0.016                   & 0.838                    & 0.015                   & 0.788                    & 0.017                   & 0.780                    & 0.019                   & 0.838                    & 0.017                   & 0.786                    & 0.018                   & 0.757                    & 0.017                   & 0.819                    & 0.017                   & 0.767                    & 0.019                  \\
                       & xbart                    & 0.830                    & 0.017                   & 0.855                    & 0.016                   & 0.839                    & 0.016                   & 0.827                    & 0.017                   & \textbf{0.853}           & \textbf{0.016}          & \textbf{0.832}           & \textbf{0.017}          & 0.820                    & 0.018                   & \textbf{0.845}           & \textbf{0.016}          & \textbf{0.826}           & \textbf{0.017}         \\
                       & mbart                    & 0.376                    & 0.035                   & 0.463                    & 0.055                   & 0.400                    & 0.044                   & 0.395                    & 0.039                   & 0.514                    & 0.058                   & 0.419                    & 0.042                   & 0.390                    & 0.033                   & 0.487                    & 0.053                   & 0.406                    & 0.038                  \\
                       & drbart                   & 0.739                    & 0.019                   & 0.793                    & 0.018                   & 0.737                    & 0.018                   & 0.729                    & 0.021                   & 0.811                    & 0.018                   & 0.725                    & 0.022                   & 0.703                    & 0.019                   & 0.801                    & 0.022                   & 0.698                    & 0.023                  \\
3                      & rfbart                   & 0.759                    & 0.038                   & 0.511                    & 0.034                   & 0.686                    & 0.035                   & 0.759                    & 0.036                   & 0.511                    & 0.034                   & 0.684                    & 0.035                   & 0.759                    & 0.038                   & 0.511                    & 0.035                   & 0.685                    & 0.035                  \\
                       & rbart                    & 0.920                    & 0.021                   & 0.608                    & 0.021                   & 0.804                    & 0.027                   & \textbf{0.920}           & \textbf{0.021}          & 0.608                    & 0.021                   & 0.803                    & 0.027                   & \textbf{0.918}           & \textbf{0.021}          & 0.608                    & 0.022                   & 0.803                    & 0.027                  \\
                       & gbm.gl                   & 0.892                    & 0.021                   & 0.792                    & 0.026                   & \textbf{0.840}           & \textbf{0.019}          & 0.887                    & 0.024                   & 0.825                    & 0.025                   & \textbf{0.834}           & \textbf{0.020}          & \textbf{0.886}           & \textbf{0.021}          & \textbf{0.826}           & \textbf{0.041}          & \textbf{0.829}           & \textbf{0.023}         \\
                       & gbm.agl                  & \textbf{0.928}           & \textbf{0.021}          & 0.789                    & 0.027                   & \textbf{0.842}           & \textbf{0.018}          & \textbf{0.915}           & \textbf{0.019}          & 0.824                    & 0.037                   & 0.829                    & 0.020                   & \textbf{0.882}           & \textbf{0.022}          & 0.826                    & 0.044                   & 0.822                    & 0.023                  \\
                       & ctree.gl                 & \textbf{0.918}           & \textbf{0.016}          & \textbf{0.847}           & \textbf{0.023}          & \textbf{0.853}           & \textbf{0.019}          & \textbf{0.908}           & \textbf{0.020}          & 0.843                    & 0.024                   & \textbf{0.837}           & \textbf{0.023}          & 0.871                    & 0.022                   & 0.823                    & 0.021                   & \textbf{0.826}           & \textbf{0.023}         \\
                       & ctree.agl                & \textbf{0.917}           & \textbf{0.018}          & \textbf{0.848}           & \textbf{0.027}          & 0.836                    & 0.020                   & 0.902                    & 0.022                   & \textbf{0.845}           & \textbf{0.026}          & 0.818                    & 0.025                   & 0.854                    & 0.024                   & \textbf{0.817}           & \textbf{0.025}          & 0.806                    & 0.025                  \\ \hline
                       & sbart                    & 0.814                    & 0.016                   & \textbf{0.815}           & \textbf{0.041}          & 0.847                    & 0.013                   & 0.817                    & 0.014                   & \textbf{0.816}           & \textbf{0.044}          & 0.845                    & 0.013                   & 0.810                    & 0.015                   & \textbf{0.808}           & \textbf{0.044}          & \textbf{0.842}           & \textbf{0.013}         \\
                       & tbart                    & 0.796                    & 0.015                   & 0.775                    & 0.024                   & 0.824                    & 0.011                   & 0.803                    & 0.013                   & 0.780                    & 0.025                   & 0.824                    & 0.013                   & 0.776                    & 0.016                   & 0.750                    & 0.024                   & 0.801                    & 0.014                  \\
                       & xbart                    & 0.832                    & 0.013                   & \textbf{0.797}           & \textbf{0.026}          & 0.854                    & 0.012                   & 0.832                    & 0.014                   & \textbf{0.799}           & \textbf{0.025}          & \textbf{0.852}           & \textbf{0.012}          & 0.818                    & 0.015                   & \textbf{0.780}           & \textbf{0.026}          & 0.841                    & 0.012                  \\
                       & mbart                    & 0.486                    & 0.024                   & 0.441                    & 0.035                   & 0.509                    & 0.027                   & 0.505                    & 0.028                   & 0.489                    & 0.038                   & 0.523                    & 0.032                   & 0.489                    & 0.028                   & 0.465                    & 0.032                   & 0.507                    & 0.030                  \\
                       & drbart                   & 0.755                    & 0.015                   & 0.715                    & 0.027                   & 0.780                    & 0.015                   & 0.758                    & 0.014                   & 0.748                    & 0.026                   & 0.780                    & 0.014                   & 0.730                    & 0.018                   & 0.714                    & 0.028                   & 0.751                    & 0.017                  \\
4                      & rfbart                   & 0.727                    & 0.030                   & 0.455                    & 0.023                   & 0.684                    & 0.029                   & 0.727                    & 0.030                   & 0.454                    & 0.023                   & 0.684                    & 0.030                   & 0.726                    & 0.030                   & 0.453                    & 0.023                   & 0.682                    & 0.031                  \\
                       & rbart                    & \textbf{0.918}           & \textbf{0.015}          & 0.491                    & 0.020                   & 0.828                    & 0.018                   & \textbf{0.918}           & \textbf{0.016}          & 0.491                    & 0.020                   & 0.827                    & 0.018                   & \textbf{0.915}           & \textbf{0.017}          & 0.491                    & 0.020                   & 0.826                    & 0.019                  \\
                       & gbm.gl                   & 0.893                    & 0.019                   & 0.712                    & 0.035                   & \textbf{0.861}           & \textbf{0.015}          & 0.889                    & 0.021                   & 0.764                    & 0.044                   & \textbf{0.855}           & \textbf{0.015}          & \textbf{0.887}           & \textbf{0.017}          & 0.737                    & 0.048                   & \textbf{0.853}           & \textbf{0.015}         \\
                       & gbm.agl                  & \textbf{0.920}           & \textbf{0.021}          & \textbf{0.802}           & \textbf{0.055}          & \textbf{0.855}           & \textbf{0.015}          & \textbf{0.910}           & \textbf{0.018}          & \textbf{0.818}           & \textbf{0.047}          & 0.848                    & 0.016                   & \textbf{0.878}           & \textbf{0.015}          & \textbf{0.755}           & \textbf{0.051}          & 0.839                    & 0.015                  \\
                       & ctree.gl                 & \textbf{0.908}           & \textbf{0.016}          & 0.784                    & 0.041                   & \textbf{0.867}           & \textbf{0.014}          & \textbf{0.898}           & \textbf{0.017}          & 0.782                    & 0.044                   & \textbf{0.856}           & \textbf{0.016}          & 0.873                    & 0.017                   & 0.755                    & 0.039                   & \textbf{0.850}           & \textbf{0.015}         \\
                       & ctree.agl                & 0.899                    & 0.022                   & 0.792                    & 0.040                   & 0.842                    & 0.016                   & 0.884                    & 0.022                   & 0.784                    & 0.040                   & 0.830                    & 0.017                   & 0.841                    & 0.023                   & 0.743                    & 0.036                   & 0.822                    & 0.020                  \\ \hline
                       & sbart                    & 0.743                    & 0.018                   & \textbf{0.800}           & \textbf{0.035}          & 0.771                    & 0.017                   & 0.743                    & 0.018                   & 0.795                    & 0.035                   & 0.770                    & 0.018                   & 0.735                    & 0.018                   & \textbf{0.791}           & \textbf{0.033}          & \textbf{0.765}           & \textbf{0.017}         \\
                       & tbart                    & 0.716                    & 0.020                   & 0.727                    & 0.023                   & 0.738                    & 0.017                   & 0.727                    & 0.019                   & 0.720                    & 0.026                   & 0.737                    & 0.019                   & 0.684                    & 0.019                   & 0.702                    & 0.025                   & 0.700                    & 0.018                  \\
                       & xbart                    & 0.756                    & 0.021                   & 0.751                    & 0.025                   & 0.773                    & 0.018                   & 0.756                    & 0.019                   & 0.738                    & 0.027                   & 0.767                    & 0.018                   & 0.729                    & 0.019                   & 0.733                    & 0.026                   & 0.744                    & 0.019                  \\
                       & mbart                    & 0.359                    & 0.025                   & 0.311                    & 0.029                   & 0.363                    & 0.026                   & 0.371                    & 0.025                   & 0.358                    & 0.028                   & 0.377                    & 0.028                   & 0.351                    & 0.025                   & 0.330                    & 0.028                   & 0.354                    & 0.028                  \\
                       & drbart                   & 0.686                    & 0.017                   & 0.681                    & 0.026                   & 0.701                    & 0.017                   & 0.680                    & 0.016                   & 0.703                    & 0.027                   & 0.690                    & 0.017                   & 0.655                    & 0.014                   & 0.689                    & 0.027                   & 0.662                    & 0.015                  \\
5                      & rfbart                   & 0.662                    & 0.031                   & 0.382                    & 0.019                   & 0.607                    & 0.029                   & 0.661                    & 0.032                   & 0.382                    & 0.019                   & 0.606                    & 0.029                   & 0.660                    & 0.032                   & 0.381                    & 0.019                   & 0.603                    & 0.030                  \\
                       & rbart                    & \textbf{0.866}           & \textbf{0.022}          & 0.421                    & 0.018                   & 0.741                    & 0.021                   & \textbf{0.865}           & \textbf{0.022}          & 0.420                    & 0.019                   & 0.739                    & 0.022                   & \textbf{0.861}           & \textbf{0.023}          & 0.420                    & 0.019                   & 0.736                    & 0.022                  \\
                       & gbm.gl                   & 0.852                    & 0.024                   & \textbf{0.754}           & \textbf{0.031}          & \textbf{0.794}           & \textbf{0.020}          & \textbf{0.849}           & \textbf{0.022}          & \textbf{0.813}           & \textbf{0.036}          & \textbf{0.788}           & \textbf{0.019}          & \textbf{0.832}           & \textbf{0.021}          & \textbf{0.785}           & \textbf{0.035}          & \textbf{0.785}           & \textbf{0.017}         \\
                       & gbm.agl                  & \textbf{0.882}           & \textbf{0.024}          & \textbf{0.812}           & \textbf{0.044}          & \textbf{0.792}           & \textbf{0.018}          & 0.848                    & 0.024                   & \textbf{0.805}           & \textbf{0.039}          & \textbf{0.769}           & \textbf{0.020}          & 0.796                    & 0.024                   & \textbf{0.764}           & \textbf{0.041}          & 0.745                    & 0.018                  \\
                       & ctree.gl                 & \textbf{0.864}           & \textbf{0.021}          & 0.734                    & 0.049                   & \textbf{0.795}           & \textbf{0.018}          & \textbf{0.849}           & \textbf{0.023}          & \textbf{0.745}           & \textbf{0.064}          & \textbf{0.777}           & \textbf{0.018}          & \textbf{0.821}           & \textbf{0.025}          & 0.733                    & 0.052                   & \textbf{0.772}           & \textbf{0.019}         \\
                       & ctree.agl                & 0.843                    & 0.027                   & 0.711                    & 0.052                   & 0.748                    & 0.022                   & 0.806                    & 0.029                   & 0.709                    & 0.068                   & 0.723                    & 0.025                   & 0.757                    & 0.032                   & 0.686                    & 0.053                   & 0.714                    & 0.025                  \\ \hline
\end{tabular}
\\\textbf{note}: Bolded text means the best three results for each scenario. The first column indicates the number of groups, the second column indicates the evaluation method, and the third to last columns are the results of the different simulation scenarios. Each scenario is named “main effect function - treatment effect function” based on the corresponding data generating function, where “L” is a linear function, “S” is a step-wise function and “N” is a nonlinear function.
\end{table}
\end{landscape}

\clearpage
\section{Appendix 2: All Simulation Results}\label{apd2}
\begin{landscape}
\begin{table}[]
\caption{Results of mean mPEHE across twelve scenarios for meta-learners using xgboost and random forest in RCT setting}\label{mPEHE.rct.brf}
\vspace{0.5cm}
\small
\begin{tabular}{c|l|ll|ll|ll|ll|ll|ll|ll|ll|ll}
\multirow{2}{*}{Group} & \multicolumn{1}{c|}{\multirow{2}{*}{Methods}} & \multicolumn{2}{c|}{L-L}                           & \multicolumn{2}{c|}{L-S}                           & \multicolumn{2}{c|}{L-N}                           & \multicolumn{2}{c|}{S-L}                           & \multicolumn{2}{c|}{S-S}                           & \multicolumn{2}{c|}{S-N}                           & \multicolumn{2}{c|}{N-L}                           & \multicolumn{2}{c|}{N-S}                           & \multicolumn{2}{c}{N-N}                           \\ \cline{3-20} 
                       & \multicolumn{1}{c|}{}                         & \multicolumn{1}{c}{mean} & \multicolumn{1}{c|}{sd} & \multicolumn{1}{c}{mean} & \multicolumn{1}{c|}{sd} & \multicolumn{1}{c}{mean} & \multicolumn{1}{c|}{sd} & \multicolumn{1}{c}{mean} & \multicolumn{1}{c|}{sd} & \multicolumn{1}{c}{mean} & \multicolumn{1}{c|}{sd} & \multicolumn{1}{c}{mean} & \multicolumn{1}{c|}{sd} & \multicolumn{1}{c}{mean} & \multicolumn{1}{c|}{sd} & \multicolumn{1}{c}{mean} & \multicolumn{1}{c|}{sd} & \multicolumn{1}{c}{mean} & \multicolumn{1}{c}{sd} \\ \hline
                       & sboost                                        & 0.89                     & 0.04                    & 0.69                     & 0.05                    & 0.79                     & 0.03                    & 0.84                     & 0.03                    & 0.67                     & 0.05                    & 0.77                     & 0.03                    & 0.89                     & 0.03                    & 0.73                     & 0.04                    & 0.83                     & 0.03                   \\
\hline
\end{tabular}
\\\textbf{note}: The first column indicates the number of groups, the second column indicates the evaluation method, and the third to last columns are the results of the different simulation scenarios. Each scenario is named “main effect function - treatment effect function” based on the corresponding data generating function, where “L” is a linear function, “S” is a step-wise function and “N” is a nonlinear function.
\end{table}
\end{landscape}

\begin{landscape}
\begin{table}[]
\caption{Results of mean mPEHE across twelve scenarios for meta-learners using xgboost and random forest in observational study setting}\label{mPEHE.obv.brf}
\vspace{0.5cm}
\small
\begin{tabular}{c|l|ll|ll|ll|ll|ll|ll|ll|ll|ll}
\multirow{2}{*}{Group} & \multicolumn{1}{c|}{\multirow{2}{*}{Methods}} & \multicolumn{2}{c|}{L-L}                           & \multicolumn{2}{c|}{L-S}                           & \multicolumn{2}{c|}{L-N}                           & \multicolumn{2}{c|}{S-L}                           & \multicolumn{2}{c|}{S-S}                           & \multicolumn{2}{c|}{S-N}                           & \multicolumn{2}{c|}{N-L}                           & \multicolumn{2}{c|}{N-S}                           & \multicolumn{2}{c}{N-N}                           \\ \cline{3-20} 
                       & \multicolumn{1}{c|}{}                         & \multicolumn{1}{c}{mean} & \multicolumn{1}{c|}{sd} & \multicolumn{1}{c}{mean} & \multicolumn{1}{c|}{sd} & \multicolumn{1}{c}{mean} & \multicolumn{1}{c|}{sd} & \multicolumn{1}{c}{mean} & \multicolumn{1}{c|}{sd} & \multicolumn{1}{c}{mean} & \multicolumn{1}{c|}{sd} & \multicolumn{1}{c}{mean} & \multicolumn{1}{c|}{sd} & \multicolumn{1}{c}{mean} & \multicolumn{1}{c|}{sd} & \multicolumn{1}{c}{mean} & \multicolumn{1}{c|}{sd} & \multicolumn{1}{c}{mean} & \multicolumn{1}{c}{sd} \\ \hline
                       & sboost                                        & 0.98                     & 0.05                    & 0.68                     & 0.04                    & 0.84                     & 0.04                    & 0.91                     & 0.05                    & 0.65                     & 0.03                    & 0.82                     & 0.04                    & 0.98                     & 0.04                    & 0.72                     & 0.04                    & 0.88                     & 0.04                   \\
\hline
\end{tabular}
\\\textbf{note}: The first column indicates the number of groups, the second column indicates the evaluation method, and the third to last columns are the results of the different simulation scenarios. Each scenario is named “main effect function - treatment effect function” based on the corresponding data generating function, where “L” is a linear function, “S” is a step-wise function and “N” is a nonlinear function.
\end{table}
\end{landscape}

\begin{landscape}
\begin{table}[]
\caption{Results of mean absolute relative bias across twelve scenarios for meta-learners using xgboost and random forest in RCT setting}\label{mbias.rct.brf}
\vspace{0.5cm}
\small
\begin{tabular}{c|l|ll|ll|ll|ll|ll|ll|ll|ll|ll}
\multirow{2}{*}{Group} & \multicolumn{1}{c|}{\multirow{2}{*}{Methods}} & \multicolumn{2}{c|}{L-L}                           & \multicolumn{2}{c|}{L-S}                           & \multicolumn{2}{c|}{L-N}                           & \multicolumn{2}{c|}{S-L}                           & \multicolumn{2}{c|}{S-S}                           & \multicolumn{2}{c|}{S-N}                           & \multicolumn{2}{c|}{N-L}                           & \multicolumn{2}{c|}{N-S}                           & \multicolumn{2}{c}{N-N}                           \\ \cline{3-20} 
                       & \multicolumn{1}{c|}{}                         & \multicolumn{1}{c}{mean} & \multicolumn{1}{c|}{sd} & \multicolumn{1}{c}{mean} & \multicolumn{1}{c|}{sd} & \multicolumn{1}{c}{mean} & \multicolumn{1}{c|}{sd} & \multicolumn{1}{c}{mean} & \multicolumn{1}{c|}{sd} & \multicolumn{1}{c}{mean} & \multicolumn{1}{c|}{sd} & \multicolumn{1}{c}{mean} & \multicolumn{1}{c|}{sd} & \multicolumn{1}{c}{mean} & \multicolumn{1}{c|}{sd} & \multicolumn{1}{c}{mean} & \multicolumn{1}{c|}{sd} & \multicolumn{1}{c}{mean} & \multicolumn{1}{c}{sd} \\ \hline
                       & sboost                                        & 0.141                    & 0.076                   & 0.085                    & 0.050                   & 0.128                    & 0.076                   & 0.098                    & 0.059                   & 0.090                    & 0.052                   & 0.096                    & 0.058                   & 0.156                    & 0.070                   & 0.088                    & 0.058                   & 0.143                    & 0.076                  \\
\hline
\end{tabular}
\\\textbf{note}: The first column indicates the number of groups, the second column indicates the evaluation method, and the third to last columns are the results of the different simulation scenarios. Each scenario is named “main effect function - treatment effect function” based on the corresponding data generating function, where “L” is a linear function, “S” is a step-wise function and “N” is a nonlinear function.
\end{table}
\end{landscape}

\begin{landscape}
\begin{table}[]
\caption{Results of mean absolute relative bias across twelve scenarios for meta-learners using xgboost and random forest in observational study setting}\label{mbias.obv.brf}
\vspace{0.5cm}
\small
\begin{tabular}{c|l|ll|ll|ll|ll|ll|ll|ll|ll|ll}
\multirow{2}{*}{Group} & \multicolumn{1}{c|}{\multirow{2}{*}{Methods}} & \multicolumn{2}{c|}{L-L}                           & \multicolumn{2}{c|}{L-S}                           & \multicolumn{2}{c|}{L-N}                           & \multicolumn{2}{c|}{S-L}                           & \multicolumn{2}{c|}{S-S}                           & \multicolumn{2}{c|}{S-N}                           & \multicolumn{2}{c|}{N-L}                           & \multicolumn{2}{c|}{N-S}                           & \multicolumn{2}{c}{N-N}                           \\ \cline{3-20} 
                       & \multicolumn{1}{c|}{}                         & \multicolumn{1}{c}{mean} & \multicolumn{1}{c|}{sd} & \multicolumn{1}{c}{mean} & \multicolumn{1}{c|}{sd} & \multicolumn{1}{c}{mean} & \multicolumn{1}{c|}{sd} & \multicolumn{1}{c}{mean} & \multicolumn{1}{c|}{sd} & \multicolumn{1}{c}{mean} & \multicolumn{1}{c|}{sd} & \multicolumn{1}{c}{mean} & \multicolumn{1}{c|}{sd} & \multicolumn{1}{c}{mean} & \multicolumn{1}{c|}{sd} & \multicolumn{1}{c}{mean} & \multicolumn{1}{c|}{sd} & \multicolumn{1}{c}{mean} & \multicolumn{1}{c}{sd} \\ \hline
                       & sboost                                        & 0.182                    & 0.100                   & 0.123                    & 0.071                   & 0.121                    & 0.067                   & 0.241                    & 0.102                   & 0.105                    & 0.062                   & 0.182                    & 0.091                   & 0.190                    & 0.093                   & 0.108                    & 0.062                   & 0.150                    & 0.084                  \\
\hline
\end{tabular}
\\\textbf{note}: The first column indicates the number of groups, the second column indicates the evaluation method, and the third to last columns are the results of the different simulation scenarios. Each scenario is named “main effect function - treatment effect function” based on the corresponding data generating function, where “L” is a linear function, “S” is a step-wise function and “N” is a nonlinear function.
\end{table}
\end{landscape}

\begin{landscape}
\begin{table}[]
\caption{Results of mean Cohen’s kappa across twelve scenarios for meta-learners using xgboost and random forest in RCT setting}\label{mkappa.rct.brf}
\vspace{0.5cm}
\small
\begin{tabular}{c|l|ll|ll|ll|ll|ll|ll|ll|ll|ll}
\multirow{2}{*}{Group} & \multicolumn{1}{c|}{\multirow{2}{*}{Methods}} & \multicolumn{2}{c|}{L-L}                           & \multicolumn{2}{c|}{L-S}                           & \multicolumn{2}{c|}{L-N}                           & \multicolumn{2}{c|}{S-L}                           & \multicolumn{2}{c|}{S-S}                           & \multicolumn{2}{c|}{S-N}                           & \multicolumn{2}{c|}{N-L}                           & \multicolumn{2}{c|}{N-S}                           & \multicolumn{2}{c}{N-N}                           \\ \cline{3-20} 
                       & \multicolumn{1}{c|}{}                         & \multicolumn{1}{c}{mean} & \multicolumn{1}{c|}{sd} & \multicolumn{1}{c}{mean} & \multicolumn{1}{c|}{sd} & \multicolumn{1}{c}{mean} & \multicolumn{1}{c|}{sd} & \multicolumn{1}{c}{mean} & \multicolumn{1}{c|}{sd} & \multicolumn{1}{c}{mean} & \multicolumn{1}{c|}{sd} & \multicolumn{1}{c}{mean} & \multicolumn{1}{c|}{sd} & \multicolumn{1}{c}{mean} & \multicolumn{1}{c|}{sd} & \multicolumn{1}{c}{mean} & \multicolumn{1}{c|}{sd} & \multicolumn{1}{c}{mean} & \multicolumn{1}{c}{sd} \\ \hline
                       & sboost                                        & 0.746                    & 0.022                   & 0.857                    & 0.022                   & 0.754                    & 0.023                   & 0.763                    & 0.022                   & 0.860                    & 0.025                   & 0.763                    & 0.021                   & 0.746                    & 0.023                   & 0.853                    & 0.027                   & 0.744                    & 0.023                  \\
\hline
\end{tabular}
\\\textbf{note}: The first column indicates the number of groups, the second column indicates the evaluation method, and the third to last columns are the results of the different simulation scenarios. Each scenario is named “main effect function - treatment effect function” based on the corresponding data generating function, where “L” is a linear function, “S” is a step-wise function and “N” is a nonlinear function.
\end{table}
\end{landscape}

\begin{landscape}
\begin{table}[]
\caption{Results of mean Cohen’s kappa across twelve scenarios for meta-learners using xgboost and random forest in observational study setting}\label{mkappa.obv.brf}
\vspace{0.5cm}
\small
\begin{tabular}{c|l|ll|ll|ll|ll|ll|ll|ll|ll|ll}
\multirow{2}{*}{Group} & \multicolumn{1}{c|}{\multirow{2}{*}{Methods}} & \multicolumn{2}{c|}{L-L}                           & \multicolumn{2}{c|}{L-S}                           & \multicolumn{2}{c|}{L-N}                           & \multicolumn{2}{c|}{S-L}                           & \multicolumn{2}{c|}{S-S}                           & \multicolumn{2}{c|}{S-N}                           & \multicolumn{2}{c|}{N-L}                           & \multicolumn{2}{c|}{N-S}                           & \multicolumn{2}{c}{N-N}                           \\ \cline{3-20} 
                       & \multicolumn{1}{c|}{}                         & \multicolumn{1}{c}{mean} & \multicolumn{1}{c|}{sd} & \multicolumn{1}{c}{mean} & \multicolumn{1}{c|}{sd} & \multicolumn{1}{c}{mean} & \multicolumn{1}{c|}{sd} & \multicolumn{1}{c}{mean} & \multicolumn{1}{c|}{sd} & \multicolumn{1}{c}{mean} & \multicolumn{1}{c|}{sd} & \multicolumn{1}{c}{mean} & \multicolumn{1}{c|}{sd} & \multicolumn{1}{c}{mean} & \multicolumn{1}{c|}{sd} & \multicolumn{1}{c}{mean} & \multicolumn{1}{c|}{sd} & \multicolumn{1}{c}{mean} & \multicolumn{1}{c}{sd} \\ \hline
                       & sboost                                        & 0.683                    & 0.037                   & 0.868                    & 0.018                   & 0.723                    & 0.025                   & 0.709                    & 0.029                   & 0.869                    & 0.023                   & 0.727                    & 0.024                   & 0.688                    & 0.025                   & 0.865                    & 0.029                   & 0.702                    & 0.029                  \\
\hline
\end{tabular}
\\\textbf{note}: The first column indicates the number of groups, the second column indicates the evaluation method, and the third to last columns are the results of the different simulation scenarios. Each scenario is named “main effect function - treatment effect function” based on the corresponding data generating function, where “L” is a linear function, “S” is a step-wise function and “N” is a nonlinear function.
\end{table}
\end{landscape}

\begin{landscape}
\begin{table}[]
\caption{Results of mean Spearman's rank correlation across twelve scenarios for meta-learners using xgboost and random forest in RCT setting}\label{mcord.rct.brf}
\vspace{0.5cm}
\small
\begin{tabular}{c|l|ll|ll|ll|ll|ll|ll|ll|ll|ll}
\multirow{2}{*}{Group} & \multicolumn{1}{c|}{\multirow{2}{*}{Methods}} & \multicolumn{2}{c|}{L-L}                           & \multicolumn{2}{c|}{L-S}                           & \multicolumn{2}{c|}{L-N}                           & \multicolumn{2}{c|}{S-L}                           & \multicolumn{2}{c|}{S-S}                           & \multicolumn{2}{c|}{S-N}                           & \multicolumn{2}{c|}{N-L}                           & \multicolumn{2}{c|}{N-S}                           & \multicolumn{2}{c}{N-N}                           \\ \cline{3-20} 
                       & \multicolumn{1}{c|}{}                         & \multicolumn{1}{c}{mean} & \multicolumn{1}{c|}{sd} & \multicolumn{1}{c}{mean} & \multicolumn{1}{c|}{sd} & \multicolumn{1}{c}{mean} & \multicolumn{1}{c|}{sd} & \multicolumn{1}{c}{mean} & \multicolumn{1}{c|}{sd} & \multicolumn{1}{c}{mean} & \multicolumn{1}{c|}{sd} & \multicolumn{1}{c}{mean} & \multicolumn{1}{c|}{sd} & \multicolumn{1}{c}{mean} & \multicolumn{1}{c|}{sd} & \multicolumn{1}{c}{mean} & \multicolumn{1}{c|}{sd} & \multicolumn{1}{c}{mean} & \multicolumn{1}{c}{sd} \\ \hline
                       & sboost                                        & 0.692                    & 0.025                   & 0.817                    & 0.017                   & 0.710                    & 0.026                   & 0.718                    & 0.023                   & 0.822                    & 0.019                   & 0.722                    & 0.024                   & 0.712                    & 0.023                   & 0.814                    & 0.022                   & 0.710                    & 0.021                  \\
\hline
\end{tabular}
\\\textbf{note}: The first column indicates the number of groups, the second column indicates the evaluation method, and the third to last columns are the results of the different simulation scenarios. Each scenario is named “main effect function - treatment effect function” based on the corresponding data generating function, where “L” is a linear function, “S” is a step-wise function and “N” is a nonlinear function.
\end{table}
\end{landscape}

\begin{landscape}
\begin{table}[]
\caption{Results of mean Spearman's rank correlation across twelve scenarios for meta-learners using xgboost and random forest in observational study setting}\label{mcord.obv.brf}
\vspace{0.5cm}
\small
\begin{tabular}{c|l|ll|ll|ll|ll|ll|ll|ll|ll|ll}
\multirow{2}{*}{Group} & \multicolumn{1}{c|}{\multirow{2}{*}{Methods}} & \multicolumn{2}{c|}{L-L}                           & \multicolumn{2}{c|}{L-S}                           & \multicolumn{2}{c|}{L-N}                           & \multicolumn{2}{c|}{S-L}                           & \multicolumn{2}{c|}{S-S}                           & \multicolumn{2}{c|}{S-N}                           & \multicolumn{2}{c|}{N-L}                           & \multicolumn{2}{c|}{N-S}                           & \multicolumn{2}{c}{N-N}                           \\ \cline{3-20} 
                       & \multicolumn{1}{c|}{}                         & \multicolumn{1}{c}{mean} & \multicolumn{1}{c|}{sd} & \multicolumn{1}{c}{mean} & \multicolumn{1}{c|}{sd} & \multicolumn{1}{c}{mean} & \multicolumn{1}{c|}{sd} & \multicolumn{1}{c}{mean} & \multicolumn{1}{c|}{sd} & \multicolumn{1}{c}{mean} & \multicolumn{1}{c|}{sd} & \multicolumn{1}{c}{mean} & \multicolumn{1}{c|}{sd} & \multicolumn{1}{c}{mean} & \multicolumn{1}{c|}{sd} & \multicolumn{1}{c}{mean} & \multicolumn{1}{c|}{sd} & \multicolumn{1}{c}{mean} & \multicolumn{1}{c}{sd} \\ \hline
                       & sboost                                        & 0.635                    & 0.037                   & 0.818                    & 0.019                   & 0.669                    & 0.027                   & 0.668                    & 0.028                   & 0.818                    & 0.021                   & 0.683                    & 0.027                   & 0.661                    & 0.026                   & 0.814                    & 0.027                   & 0.668                    & 0.029                  \\
\hline
\end{tabular}
\\\textbf{note}: The first column indicates the number of groups, the second column indicates the evaluation method, and the third to last columns are the results of the different simulation scenarios. Each scenario is named “main effect function - treatment effect function” based on the corresponding data generating function, where “L” is a linear function, “S” is a step-wise function and “N” is a nonlinear function.
\end{table}
\end{landscape}

\clearpage
\section{Appendix 3: Grid Search results in real data application}\label{apd3}

In table\ref{grid.res}, we present the average metric values Eq.(\ref{tune_metric}) for each combination of rule generation and rule ensemble in the proposed method, for each combination of number of trees, tree depth, and shrinkage rate, applied 10 times to the data. According to these results, for "gbm" rule generation with group lasso, the number of trees = 333, depth of tree = 4 and shrinkage rate = 0.1 is the best hyperparameter combination; for "gbm" rule generation with adaptive group lasso, the number of trees = 1000, depth of tree = 3 and shrinkage rate = 0.01 is the best hyperparameter combination; for "ctree" rule generation with group lasso , the number of trees = 333, depth of tree = 3 and shrinkage rate = 0.001 is the best hyperparameter combination; for "ctree" rule generation with adaptive group lasso , the number of trees = 1000, depth of tree = 3 and shrinkage rate = 0.01 is the best hyperparameter combination. Comparing the different data generation processes, the "gbm" rule generation with adaptive group lasso performs better than other model generation processes. In addition, the "gbm" rule generation with group lasso tends to perform worse than other generation process, while the performance of "ctree" rule generation with group lasso or adaptive group lasso tends to be comparable to that of "gbm" rule generation with adaptive group lasso.
\vspace{-0.5cm}
\begin{table}[h]
\centering
\caption{Grid search results Eq.(\ref{tune_metric}), The number of trees, depth of tree and shirinkage rate are the hyper-parameters of proposed approach. The values of "gbm.gl" means the results of "gbm" rule generation with group lasso, the values of "gbm.agl" means the results of "gbm" rule generation with adaptive group lasso, the values of "ctree.gl" means the results of "ctree" rule generation with group lasso, and the values of "ctree.agl" means the results of the "ctree" rule generation with adaptive group lasso. The bold means the best results}\label{grid.res}

\begin{tabular}{ccccccc}
\hline
Number of trees & Depth of tree & Shrinkage rate & gbm.gl & gbm.agl & ctree.gl & ctree.agl \\ \hline
333             & 2             & 0.1            & 38.52  & 10.28   & 14.93    & 13.46     \\
333             & 2             & 0.01           & 41.79  & 12.02   & 15.37    & 14.92     \\
333             & 2             & 0.001          & 45.06  & 11.65   & 14.67    & 13.86     \\
333             & 3             & 0.1            & 35.88  & 10.7    & 14.87    & 13.12     \\
333             & 3             & 0.01           & 43.47  & 10.5    & 14.91    & 14.91     \\
333             & 3             & 0.001          & 47.25  & 10.34   & \textbf{14.21}    & 13.86     \\
333             & 4             & 0.1            & \textbf{30.23}  & 10.26   & 14.54    & 12.89     \\
333             & 4             & 0.01           & 47.71  & 9.51    & 15.43    & 14.18     \\
333             & 4             & 0.001          & 44.57  & 10.87   & 14.54    & 13.55     \\
666             & 2             & 0.1            & 32.89  & 10.41   & 15.29    & 12.64     \\
666             & 2             & 0.01           & 42.56  & 10.46   & 21.61    & 13.68     \\
666             & 2             & 0.001          & 46.05  & 9.53    & 22.2     & 13.62     \\
666             & 3             & 0.1            & 37.83  & 10.59   & 15.06    & 12.35     \\
666             & 3             & 0.01           & 43.68  & 10.24   & 16.65    & 13.68     \\
666             & 3             & 0.001          & 49.32  & 9.66    & 16.9     & 14.66     \\
666             & 4             & 0.1            & 33.17  & 12.01   & 15.47    & 12.65     \\
666             & 4             & 0.01           & 45.74  & 9.32    & 23.01    & 13.83     \\
666             & 4             & 0.001          & 44.64  & 10.04   & 21.64    & 14.86     \\
1000            & 2             & 0.1            & 34.56  & 9.81    & 14.71    & 11.48     \\
1000            & 2             & 0.01           & 41.56  & 10.73   & 15.77    & 13.98     \\
1000            & 2             & 0.001          & 45.19  & 9.44    & 16       & 15.15     \\
1000            & 3             & 0.1            & 39.2   & 10.46   & 15.96    & \textbf{11.19}     \\
1000            & 3             & 0.01           & 46.28  & \textbf{9.18}    & 15.4     & 14.61     \\
1000            & 3             & 0.001          & 46.05  & 10.72   & 15.32    & 15.78     \\
1000            & 4             & 0.1            & 34.67  & 9.24    & 15.96    & 11.19     \\
1000            & 4             & 0.01           & 38.43  & 10.48   & 15.99    & 15.17     \\
1000            & 4             & 0.001          & 43.43  & 10.71   & 15.73    & 15.43     \\ \hline
\end{tabular}
\end{table}

\begin{figure*}[tb]
\centering
\subfloat["gbm" rule generation with group lasso]{\includegraphics[clip, width=0.9\linewidth]{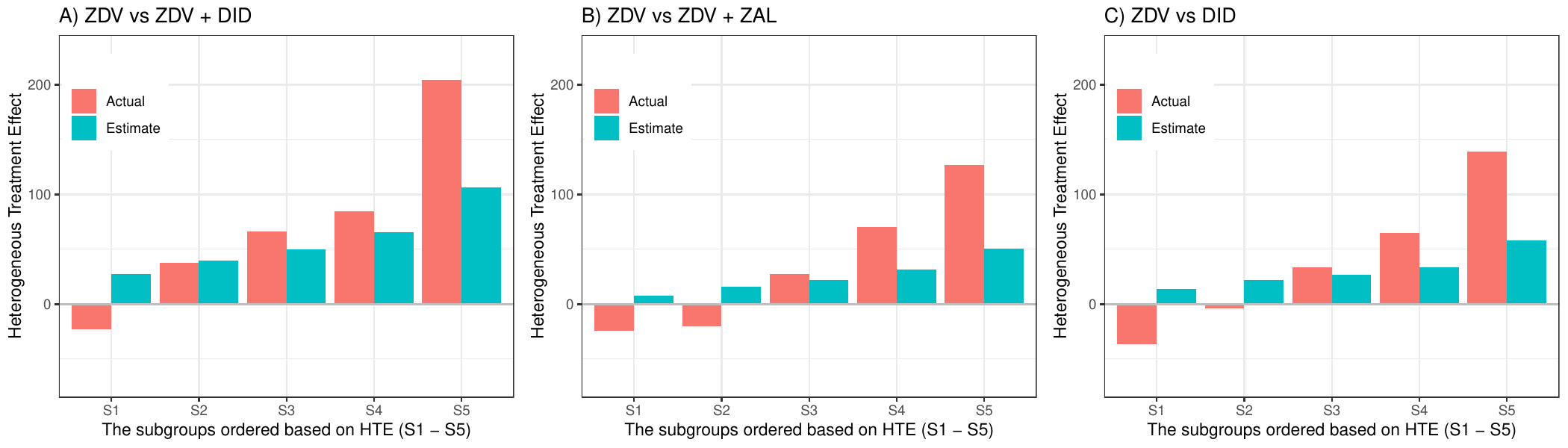}
\label{fig:label-A}}
\\
\subfloat["gbm" rule generation with adaptive group lasso]{\includegraphics[clip, width=0.9\linewidth]{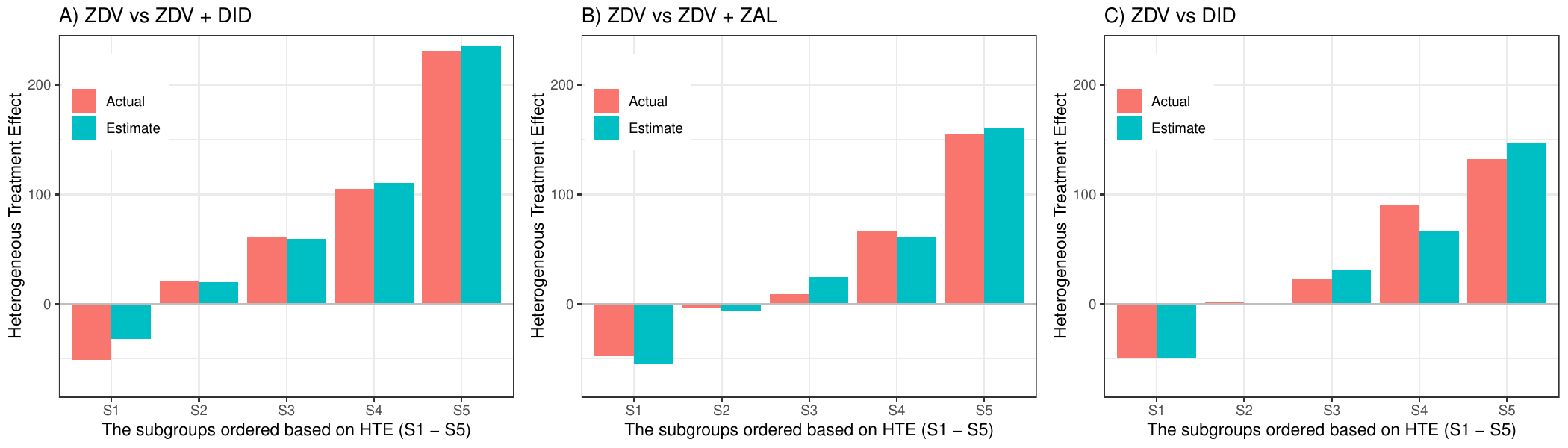}
\label{fig:label-B}}
\\
\subfloat["ctree" rule generation with group lasso]{\includegraphics[clip, width=0.9\linewidth]{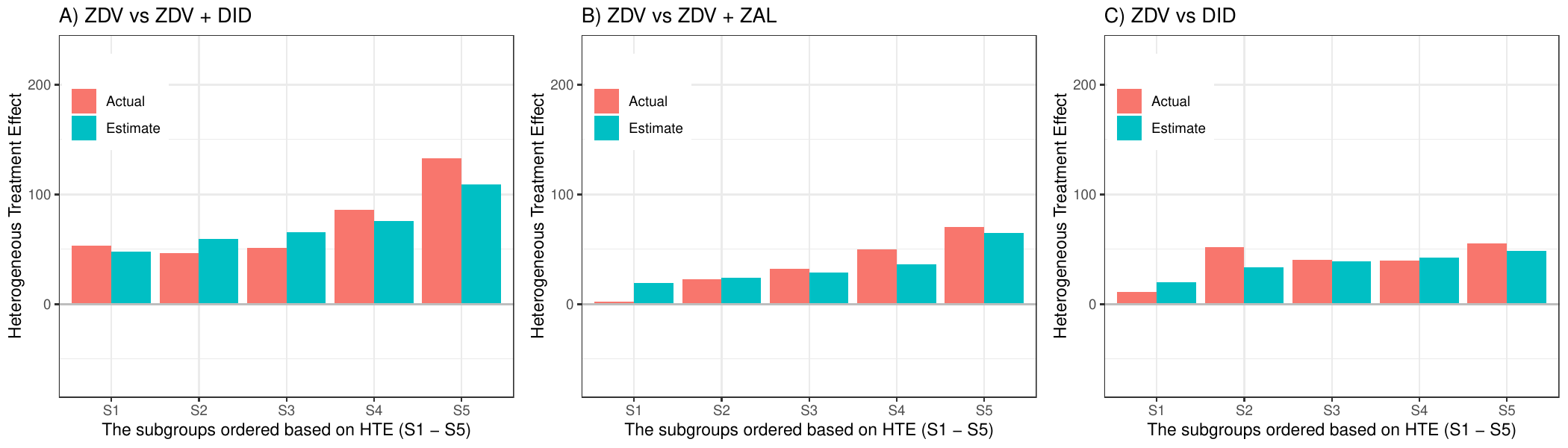}
\label{fig:label-C}}
\\
\subfloat["ctree" rule generation with adaptive group lasso]{\includegraphics[clip, width=0.9\linewidth]{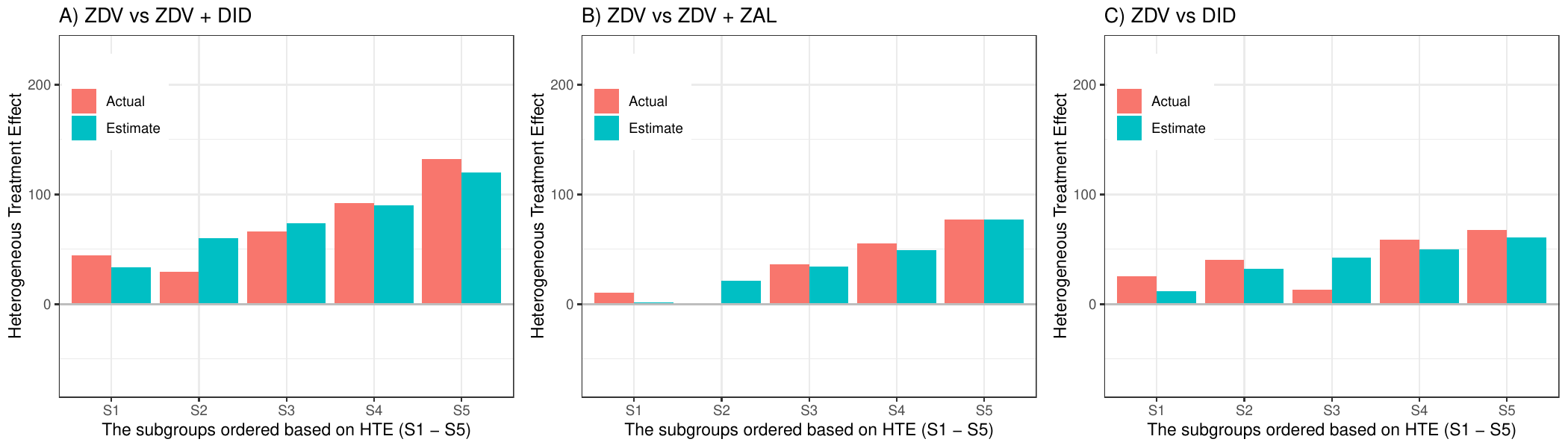}
\label{fig:label-D}}
\caption{The y-axis is the HTE of each subgroups. The x-axis is the name of the subgroups.The red bar is the actual HTE calculated as the mean difference between treatment and control within subgroups; The blue bar is the estimated HTE calculated as the mean of the estimated HTE within subgroups.}
\label{para.evl}
\end{figure*}

We also provide the graphic evaluation for each model generation process with best hyperparameters for proposed approach as in Figure\ref{para.evl}. The detail of graphical evaluation is demonstrated in section 5.2, second paragraph. First row is the graphical evaluation of "gbm" rule generation with group lasso, for these results, although both the estimated and actual HTE seems to have same trend,  there are obvious difference between the estimated and actual HTE for sub groups. The second row is the graphical evaluation of "gbm" with adaptive group lasso which shows the best results as introduced in section\ref{sec5}. Furthermore, the third and fourth rows, are the results of "ctree" with group lasso and "ctree" with adaptive group lasso. Both of them, have comparable metrics values to the metric value of "gbm" rule generation with adaptive group lasso, therefore they seems to have well graphical evaluation results. However, both results show different trends between estimated and actual HTE, and therefore these results are not as good as the best performance one. 

\end{document}